%% file: main.tex
\documentclass[lettersize,journal]{IEEEtran}

\ifCLASSOPTIONcompsoc
  \usepackage[nocompress]{cite}
\else
  \usepackage{cite}
\fi

\usepackage{makecell}
\usepackage{wrapfig}
\usepackage{graphicx}
\usepackage[table]{xcolor}
\definecolor{LightCyan}{rgb}{0.88,1,1}
\definecolor{LightYellow}{rgb}{1,1,0.7}
\usepackage{verbatim}
\usepackage{amsmath}
\usepackage[colorlinks]{hyperref}
\usepackage[capitalize]{cleveref}

\usepackage[percent]{overpic}
\usepackage{pifont}
\usepackage{algorithm}
\usepackage{algpseudocode}
\usepackage{multirow}
\usepackage{float}
\usepackage{array}
\usepackage{tikz}
\usepackage{bm}
\usepackage{bbding}
\usepackage{pifont}
\usepackage{wasysym}
\usepackage{amssymb}
\usepackage{pifont}
\usepackage{tabularx}
\usepackage{booktabs}
\usepackage[export]{adjustbox}
\usepackage{enumitem}

\definecolor{lightgray}{gray}{0.9}
\definecolor{handcrafted}{gray}{0.1}
\definecolor{deep_learning}{RGB}{28, 0, 63}
\usepackage{xspace} 
\def\eg{\emph{e.g.}\xspace}

\def\ie{\emph{i.e.}\xspace}

\Crefname{equation}{Eq.}{Eqs.}
\Crefname{figure}{Fig.}{Figs.}
\Crefname{tabular}{Tab.}{Tabs.}
\Crefname{section}{Sec.}{Secs.}

\usepackage{caption}
\newcommand{\withcode}[1]{\color{black} #1 \color{black}}

\definecolor{gold}{rgb}{1.0, 0.874, 0}
\definecolor{silver}{rgb}{0.77,0.77,0.77}
\definecolor{brown}{rgb}{0.95, 0.678, 0.4}

\definecolor{mylightgray}{RGB}{238,238,238} 
\colorlet{bgcolor}{mylightgray}

\definecolor{maincategories}{HTML}{CDE1DB} 
\definecolor{salmon}{HTML}{EDF4F2} 

\usepackage{tabulary}
\newcolumntype{I}{!{\vrule width 1pt}}
\newcolumntype{x}[1]{>{\centering\arraybackslash}p{#1pt}}
\newcolumntype{y}[1]{>{\raggedright\arraybackslash}p{#1pt}}
\newcolumntype{z}[1]{>{\raggedleft\arraybackslash}p{#1pt}}

\definecolor{higher}{HTML}{9FC4D9} 
\definecolor{lower}{HTML}{f9b0c7}  
\makeatletter
\newcommand{\thickhline}{%
    \noalign {\ifnum 0=`}\fi \hrule height 1pt
    \futurelet \reserved@a \@xhline
}
\makeatother

\title{How NeRFs and 3D Gaussian Splatting are Reshaping SLAM: a Survey}
\date{July 2022}

\author{Fabio~Tosi\textsuperscript{1} \hspace{1.5cm} 
    Youmin Zhang\textsuperscript{1,2} \hspace{1.5cm}  
    Ziren Gong\textsuperscript{1} \hspace{1.5cm}  
    Erik Sandström\textsuperscript{3} \\
    Stefano~Mattoccia\textsuperscript{1} \hspace{1.5cm}
    Martin R. Oswald\textsuperscript{3,4} \hspace{1.5cm}
    Matteo~Poggi\textsuperscript{1} \\    
    \vspace{0.3cm}
    \small \textsuperscript{1}University of Bologna, Italy \hspace{0.25cm} \textsuperscript{2}\href{https://twitter.com/hi_rockuniverse}{\color{black}{Rock Universe, China}} \hspace{0.25cm}
    \textsuperscript{3}ETH Zurich, Switzerland \hspace{0.25cm} \textsuperscript{4}University of Amsterdam, Netherlands \\
\IEEEcompsocitemizethanks{\IEEEcompsocthanksitem F. Tosi, Y. Zhang, Z. Gong, S. Mattoccia, and M. Poggi are with the University of Bologna, Italy.%
\IEEEcompsocthanksitem E. Sandström and M. R. Oswald are with ETH Zürich, Switzerland and the University of Amsterdam, Netherlands}%
}

\begin{document}

\twocolumn[{
\renewcommand\twocolumn[1][]{#1}
\maketitle
\begin{center}\vspace{-0.75cm}
\renewcommand{\tabcolsep}{1pt}
\begin{overpic}[width=1.\textwidth]{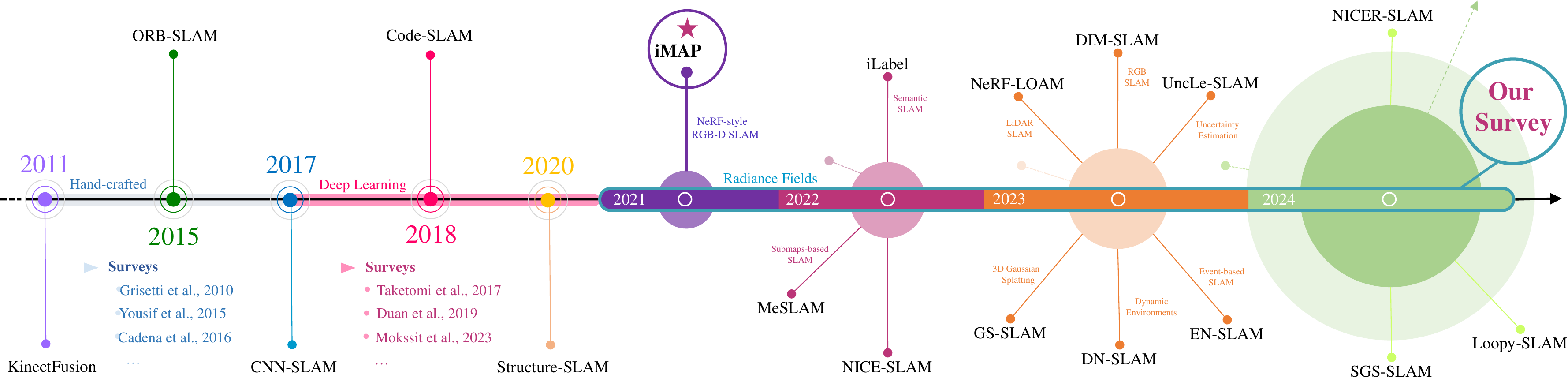}
\put (44.6,20.55) {\tiny{\cite{sucar2021imap}}} 
\put (6.05,0.3) {\tiny{\cite{kinectfusion}}} 
\put (21.35,0.3) {\tiny{\cite{tateno2017cnn}}} 
\put (38.7,0.3) {\tiny{\cite{li2020structure}}} 
\put (59.3,0.3) {\tiny{\cite{zhu2022nice}}} 
\put (52.5,4.1) {\tiny{\cite{kruzhkov2022meslam}}} 
\put (57.8,19.77) {\tiny{\cite{zhi2022ilabel}}}
\put (91.2,0.17) {\tiny{\cite{li2024sgsslam}}}
\put (99.8,1.9) {\tiny{\cite{liso2024loopy}}} 
\put (13.75,21.4) {\tiny{\cite{mur2015orb}}} 
\put (30,21.5) {\tiny{\cite{bloesch2018codeslam}}}
\put (66.6,2.58) {\tiny{\cite{yan2023gs}}} 
\put (73.6,0.91) {\tiny{\cite{ruan2023dn}}} 
\put (80.5,2.4) {\tiny{\cite{qu2023implicit}}} 
\put (67.7,18.41) {\tiny{\cite{deng2023nerf}}} 
\put (73.8,21.22) {\tiny{\cite{li2023dense}}} 
\put (80.1,18.5) {\tiny{\cite{uncleslam2023}}} 
\put (91.28,22.72) {\tiny{\cite{Zhu2023NICER}}} 
\put (14.85,5.19) {\tiny{\cite{grisetti2010tutorial}}}
\put (14.5,3.8) {\tiny{\cite{yousif2015overview}}}
\put (14.8,2.2) {\tiny{\cite{cadena2016past}}}
\put (31.9,5.3) {\tiny{\cite{taketomi2017visual}}}
\put (30.4,3.65) {\tiny{\cite{duan2019deep}}}
\put (31.4,2.2) {\tiny{\cite{mokssit2023deep}}}
\end{overpic}
\end{center}
\label{fig:SLAM_timeline}
\small \hypertarget{fig:SLAM_timeline}{Fig. 1:} {\textbf{Timeline SLAM Evolution.} This timeline shows the evolution from hand-crafted to deep learning SLAM, with key surveys marking both periods. A significant shift occurs in 2021 with iMap \cite{sucar2021imap}, introducing radiance-field-based approaches. Circle sizes on the right indicate yearly publication volumes, with 2024's outer circle projecting increased interest in NeRF and 3DGS-based SLAM.
}
\vspace{0.5cm}
}]

\begin{abstract}
Over the past two decades, research in the field of Simultaneous Localization and Mapping (SLAM) has undergone a significant evolution, highlighting its critical role in enabling autonomous exploration of unknown environments. This evolution ranges from hand-crafted methods, through the era of deep learning, to more recent developments focused on Neural Radiance Fields (NeRFs) and 3D Gaussian Splatting (3DGS) representations. Recognizing the growing body of research and the absence of a comprehensive survey on the topic, this paper aims to provide the first comprehensive overview of SLAM progress through the lens of the latest advancements in radiance fields.
It sheds light on the background, evolutionary path, inherent strengths and limitations, and serves as a fundamental reference to highlight the dynamic progress and specific challenges.

\end{abstract}


\input{chapters/introduction.tex}
\input{chapters/background.tex}

\input{chapters/taxonomy/rgbd/intro}

\input{chapters/taxonomy/rgbd/vanilla}
\input{chapters/taxonomy/rgbd/3DGS}
\input{chapters/taxonomy/rgbd/submaps}
\input{chapters/taxonomy/rgbd/segmentation}
\input{chapters/taxonomy/rgbd/dynamic}
\input{chapters/taxonomy/rgbd/uncertainty}
\input{chapters/taxonomy/rgbd/event}

\input{chapters/taxonomy/rgb/intro}
\input{chapters/taxonomy/rgb/vanilla}
\input{chapters/taxonomy/rgb/priors}
\input{chapters/taxonomy/rgb/segmentation}
\input{chapters/taxonomy/rgb/uncertainty}

\input{chapters/taxonomy/lidar/vanilla}
\input{chapters/taxonomy/lidar/3DGS}

\input{chapters/experiments}

\input{chapters/discussion.tex}
\input{chapters/conclusion.tex}

\bibliographystyle{IEEEtran}
\bibliography{egbib}

\end{document}

%% file: chapters/introduction.tex
\section{Introduction}
Simultaneous Localization and Mapping (SLAM) is a fundamental concept in the fields of computer vision and robotics. It addresses the challenge of enabling machines to autonomously  navigate and incrementally build a map of unknown environments  (\textit{mapping}) while simultaneously determining their own position and orientation (\textit{tracking}).

Originally conceived for robotics and automated systems, the demand for SLAM has expanded into a variety of domains, including augmented reality (AR), visual surveillance, medical applications, and beyond. To meet these needs, researchers have focused on developing methods for machines to autonomously construct increasingly highly accurate scene representations, influenced by the convergence of robotics, computer vision, sensor technology, and the recent progress in artificial intelligence (AI).

Typically, SLAM techniques rely on the integration of diverse sensing technologies, including cameras, laser range instruments, inertial devices, and GPS, to effectively accomplish the task at hand. Initially, sonar and LiDAR sensors were prevalent choices due to their high precision, despite being cumbersome and costly. Subsequently, the focus shifted towards visual sensors such as monocular/stereo or RGB-D cameras, which offer advantages in terms of portability, cost-effectiveness, and deployment ease. These visual sensors enable Visual Simultaneous Localization and Mapping (VSLAM) systems to capture more detailed environmental information, improve precise positioning in complex scenarios, and deliver versatile and accessible solutions.

As we outline the ideal SLAM criteria, several key aspects emerge. These include global consistency,  robust camera tracking, accurate surface modeling, real-time performance, accurate prediction in unobserved regions, scalability to large scenes, and robustness to noisy data. 

Over the years, SLAM methodologies have evolved significantly to meet these specific requirements.
At the outset, hand-crafted algorithms \cite{kinectfusion, newcombe2011dtam, salas2013slam++, elasticfusion, mur2015orb} demonstrated remarkable real-time performance and scalability. However, they face challenges in strong illumination, radiometric changes, and dynamic/poorly textured environments, resulting in unsatisfactory performance. The incorporation of advanced techniques, employing deep learning methodologies \cite{bloesch2018codeslam, tateno2017cnn, li2020structure, teed2021droid}, became crucial in improving the precision and reliability of localization and mapping. This integration takes advantage of the robust feature extraction capabilities of deep neural networks, which are particularly effective in challenging conditions. 
Nonetheless, their dependence on extensive training data and accurate ground truth annotations limits their ability to generalize to unseen scenarios. Furthermore, both hand-crafted and deep learning-based methods encounter limitations related to using discrete surface representations (point/surfel clouds \cite{zhou2013elastic, schops2019bad}, voxel hashing \cite{niessner2013real}, voxel-grids \cite{kinectfusion}, octrees \cite{steinbrucker2013large}), which lead to challenges such as sparse 3D modeling, limited spatial resolution and distortion during the reconstruction process. Additionally, accurately estimating geometries in unobserved areas remains an ongoing hurdle.

Driven by the need to overcome existing obstacles and influenced by the success of recent Neural Radiance Fields (NeRF) \cite{mildenhall2021nerf} and 3D Gaussian Splatting (3DGS) \cite{kerbl20233d} representations in novel view synthesis, along with the introduction of learned representations for modeling geometric fields  -- extensively discussed in \cite{xie2022neural} -- a revolution is reshaping SLAM. Leveraging insights from contemporary research, these approaches offer several advantages over previous methods, including continuous surface modeling, reduced memory requirements, improved noise/outlier handling, and enhanced hole filling and scene inpainting capabilities for occluded or sparse observations. In addition, they have the potential to produce denser and more compact maps that can be reconstructed as 3D meshes at arbitrary resolutions.  However, at this early stage, each technique presents both strengths and specific limitations. As such, the field is constantly evolving and requires ongoing research and innovation to make further progress. 

In response to the lack of SLAM surveys focusing on the latest developments and the growing interest in research exploring this paradigm
\footnote{\url{https://github.com/DoongLi/awesome-Implicit-NeRF-SLAM}}, this paper conducts a thorough review of contemporary radiance field-inspired SLAM techniques. 
Specifically, we undertake an in-depth investigation of 80 SLAM systems that have been published in the past three years\footnote{with a cut-off date on ECCV/IROS 2024}, reflecting the rapid pace of progress in the field.
This evolution is illustrated in Figure \textcolor{red}{\hyperref[fig:SLAM_timeline]{1}}, which provides a visual timeline of the current state of SLAM advancements. Our aim is to fill the existing gap in the survey literature by closely examining and analyzing these cutting-edge techniques, and by highlighting the rapid emergence of innovative solutions aimed at improving their inherent weaknesses. Through a detailed exploration, we intend to categorize these methods, trace their progression, and offer insights that are tailored to the specific requirements of SLAM. By serving as a valuable resource for both novices and experts, we believe that this survey represents a significant cornerstone for the future of this paradigm.

The upcoming sections are organized as follows:

\begin{itemize}
    \item Section \ref{sec:background} reviews existing SLAM surveys (\ref{sec:survey}), covers basics of radiance-field rendering theory (\ref{sec:theory}), introduces key datasets and benchmarks (\ref{sec:datasets}), and describes main evaluation metrics used in this context (\ref{sec:metrics}).
    \item Section \ref{sec:methods} is the core of our paper, focusing on key NeRF and 3DGS-inspired SLAM techniques and our structured taxonomy for organizing these advancements.
    \item Section \ref{sec:experiments}  presents quantitative results evaluating SLAM frameworks in tracking, mapping, rendering, and performance analysis across diverse scenarios.
    \item Sections \ref{sec:discussion} and \ref{sec:conclusion} focus on limitations, future research directions, and summarize the survey comprehensively.
\end{itemize}

%% file: chapters/background.tex
\section{Background}
\label{sec:background}
\subsection{Existing SLAM Surveys}
\label{sec:survey}

\begin{figure*}[]
    \centering
    \renewcommand{\tabcolsep}{1pt}
    \begin{tabular}{ccc}
        \hspace{-0.7em} \includegraphics[width=0.28\textwidth]{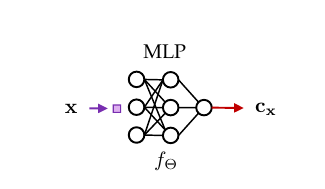} & \hspace{1.8em}
        \includegraphics[width=0.27\textwidth]{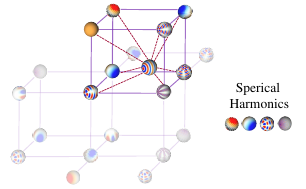} & \hspace{1em}
        \includegraphics[width=0.37\textwidth]{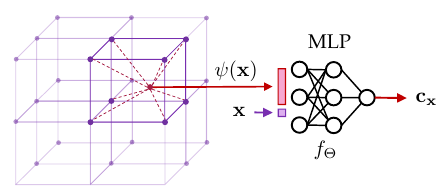} \\ 
    \end{tabular}
    \vspace{-0.5em}
    \caption{\textbf{Comparison of Scene Representations: \textit{Implicit}, \textit{Explicit}, and \textit{Hybrid}.} From left to right: \textit{Implicit} uses a neural network to approximate a radiance field, \textit{explicit} conducts volume rendering directly on learned spatial feature (voxels, hash grids, etc.), excluding neural components, and \textit{hybrid} incorporates learned spatial features $\psi$ with neural networks. Both \textit{hybrid} and \textit{explicit} approaches enable accelerated training and rendering but require additional memory resources.
    }\vspace{-0.3cm}
    \label{fig:output_representation_ihe}
\end{figure*}

SLAM has seen significant growth, resulting in a variety range of comprehensive survey papers. In the early stages, Durrant-Whyte and Bailey introduced the probabilistic nature of the SLAM problem and highlighted key methods, alongside implementations \cite{durrant2006simultaneous, bailey2006simultaneous}. Grisetti et al. \cite{grisetti2010tutorial} further delved into the graph-based SLAM problem, emphasizing its role in navigating in unknown environments. In the field of visual SLAM, Yousif \cite{yousif2015overview} provided an overview of localization and mapping techniques, incorporating basic methods and advances in visual odometry and SLAM. The advent of multiple-robot systems led to Saeedi and Clark \cite{saeedi2016multiple} reviewing state-of-the-art approaches, with a focus on multiple-robot SLAM challenges and solutions. Cadena et al. \cite{cadena2016past} presented a comprehensive reflection on the history, robustness, and new frontiers of SLAM, addressing its evolving significance across real-world applications. Taketomi et al. \cite{taketomi2017visual} categorized and summarized VSLAM algorithms from 2010 to 2016, classifying them based on feature-based, direct, and RGB-D camera approaches. Saputra et al. \cite{saputra2018visual} addressed the challenge of dynamic environments in VSLAM and Structure from Motion (SfM), presenting a taxonomy of techniques for reconstruction, segmentation and tracking of dynamic objects. The integration of deep learning with SLAM was meticulously examined by Duan et al. \cite{duan2019deep}, highlighting the progression of deep learning methods in visual SLAM. In sensor-specific contexts, Zaffar et al. \cite{zaffar2018sensors} discussed sensors employed in SLAM, while Yang et al. \cite{yang2019survey} and Zhao et al. \cite{zhao2019review} explored the applications of LiDAR and underwater SLAM, respectively. In recent years, deep learning-based VSLAM has gained considerable attention, extensively covered in \cite{chen2020survey, chen2022overview, kazerouni2022survey, tang2022perception}. Notably, \cite{zollhofer2018state} delves into recent advancements in RGB-D scene reconstruction. Ongoing developments in SLAM are explored in surveys like \cite{placed2023survey}, focusing on active SLAM strategies for precise mapping through motion planning. 

Despite the extensive body of work describing SLAM systems covering traditional and deep learning-based approaches, there is no comprehensive exploration of the advancing frontiers in SLAM techniques rooted in the latest progress in radiance fields.
Nonetheless, within the existing literature of our interest, notably in influential works like \cite{bundlefusion}, two principal SLAM strategies emerge as the \textit{frame-to-frame} and \textit{frame-to-model} tracking approaches, which are influencing the development of new methodologies based on radiance fields.
Typically, the former strategy is used in real-time systems, often involving further optimization of the estimated poses through \textit{loop-closure} (LC) or global Bundle Adjustment (BA), whereas the latter estimates camera poses from the reconstructed 3D model, often avoiding further optimizations, yet resulting less scalable to large scenes.
These strategies, often associated with the concepts of decoupled and coupled methods in recent SLAM research, serve as the foundation for the methodologies we will explore. Decoupled methods employ separate frameworks for tracking and mapping, treating them as independent tasks, while coupled methods utilize a unified representation for both tasks, allowing for a more integrated approach.

\subsection{Progress in Radiance Field Theory}
\label{sec:theory}

The term radiance field refers to a representation that describes the behavior and distribution of light within a three-dimensional space. It encapsulates how light interacts with surfaces, materials, and the surrounding environment. It can be represented implicitly, by encoding it entirely within the weights of a neural network or explicitly, by mapping light within a discrete spatial structure such as voxel grids. Explicit representations typically offer faster access but require more memory and have resolution constraints, while implicit representations provide a compact scene encoding with potentially higher rendering computational needs.  Hybrid approaches take advantage of both by using a combination of explicitly stored local latent features and shallow neural networks, using various structures such as sparse voxel hashing grids \cite{mueller2022instant, li2023neuralangelo}, multi-resolution dense voxel grids \cite{peng2020convolutional},  unordered point sets \cite{xu2022point}, and more. Figure \ref{fig:output_representation_ihe} visually illustrates these representations, which have recently had a significant impact on SLAM methodologies, primarily through the incorporation of models derived from NeRF and more recent explicit methods such as 3DGS.  Below, we briefly describe NeRF -- for image rendering and surface reconstruction -- and 3DGS, essential for understanding the upcoming SLAM approaches.

\subsubsection{Neural Radiance Field (NeRF)}
In 2020, Mildenhall et al. \cite{mildenhall2021nerf} introduced NeRF, an implicit, continuous volumetric representation, setting a new standard for novel view synthesis. In contrast to conventional explicit volumetric models, this method employs a sparse set of input views to optimize a continuous volumetric scene function, representing three-dimensional scenes via a radiance field. 
To achieve this, the original NeRF implementation requires knowledge of the camera poses and intrinsic parameters corresponding to each input view, which are estimated using the COLMAP structure-from-motion package \cite{schoenberger2016sfm,schoenberger2016mvs}. This approach has become the common practice in subsequent research building upon the NeRF framework.
Formally expressed as $f(\mathbf{x}, \mathbf{d}) \rightarrow (\mathbf{c}, \sigma)$, the model leverages an MLP (Multi-Layer Perceptron) with weights $\mathit{\Theta}$, denoted as $f_\mathit{\Theta}$, approximating a 5D function of viewing direction $\mathbf{d} = (\theta,\phi)$ and in-scene 3D coordinates $\mathbf{x} = (x,y,z)$. Notably, the representation ensures multi-view consistency by predicting the volume density $\sigma$ independently of viewing direction, while color $\mathbf{c} = (r,g,b)$ depends on both viewing direction and 3D coordinates.

The NeRF workflow for novel view synthesis involves casting camera rays through the scene to generate sampling points per pixel, computing local color and density using the NeRF MLP(s) for each sampling point, and employing volume rendering to synthesize the 2D image. Specifically, the computation of the color $C(\mathbf{r})$ resulting from a camera ray $\mathbf{r}(t) = \mathbf{o} + t\mathbf{d}$ involves an integral formulation:

\begin{equation}\label{eq:rendering}
C(\mathbf{r}) = \int_{t_1}^{t_2} T(t) \sigma(\mathbf{r}(t)) \mathbf{c}(\mathbf{r}(t),\mathbf{d}) \text{d}t
\end{equation}

Here, $\text{d}t$ denotes the differential distance traveled by the ray at each integration step. The terms $\sigma(\mathbf{r}(t))$ and $\mathbf{c}(\mathbf{r}(t), \mathbf{d})$ represent the volume density and color at point $\mathbf{r}(t)$ along the camera ray with viewing direction $\mathbf{d}$, respectively. Additionally,  $T(t) = \exp \left( -\int_{t_1}^{t} \sigma (\mathbf{r}(s)) , \text{d}s\right)$ is the accumulated transmittance from $t_1$ to $t$.

The integral computation uses quadrature by dividing the ray into $N$ evenly-spaced bins:

\begin{equation}\label{eq:quadrature}
    C(\mathbf{r}) = \sum_{i=1}^{N} \alpha_iT_i\mathbf{c}_i, \hspace{0.4cm} T_i = \text{exp}\Big(-{\sum_{j=1}^{i-1}\sigma_j\delta_j}\Big)  
\end{equation}
where, $\delta_i$ denotes the interval between consecutive samples $t_i$ and $t_{i+1}$, while $\sigma_i$ and $\mathbf{c}_i$ indicate the density and color evaluated at sample point $i$ along the ray, respectively. Additionally, $\alpha_i = (1-\text{exp}(-\sigma_i\delta_i))$ characterizes the opacity resulting from alpha compositing at sample point $i$.

The expected depth along a ray, instead, can be calculated using the accumulated transmittance:
\begin{equation}
d(\mathbf{r}) = \int_{t_1}^{t_2} T(t) \cdot \sigma(\mathbf{r}(t)) \cdot t \, dt,
\end{equation}
Similarly to Eq. \ref{eq:quadrature}, this can be approximated as:

\begin{equation}\label{eq:depth_nerf}
\hat{D}(\mathbf{r}) = \sum_{i=1}^{N} \alpha_i t_i T_i.
\end{equation}

In this context, some methods propose using expected depth to either impose depth supervision from external priors \cite{deng2022depth, roessle2022dense}  or apply regularization techniques \cite{niemeyer2022regnerf}, enhancing scene geometry and enforcing depth smoothness.

For optimization, a square error photometric loss is employed, represented as \(L = \sum_{r \in R} \| \hat{C}(\mathbf{r}) - C_{gt}(\mathbf{r}) \|^2_2\). Here, \(C_{gt}(r)\) denotes the ground truth color for the pixel associated with \(\mathbf{r}\) in the training image, and \(R\) denotes the batch of rays for synthesizing the target image.

While NeRF achieved success, challenges like slow training/rendering speeds persist. Follow-up methods, comprehensively surveyed in \cite{gao2022nerf,shahariar2023beyondpixels,tewari2022advances}, seek to enhance quality or faster training/rendering using techniques such as hashing \cite{mueller2022instant} or sparse 3D grids \cite{fridovich2022plenoxels}. 
However, these methods still struggle to accurately represent empty spaces and face image quality limitations due to structured grids, which significantly impede rendering speeds.
Other works \cite{yeninerf,wang2021nerf,lin2021barf}, instead, aims to reduce the reliance on external tools like COLMAP for camera pose estimation. While these approaches share the goal of joint pose estimation and scene reconstruction with SLAM, they differ in their processing paradigm. SLAM typically processes images sequentially as they are captured, enabling real-time operation. In contrast, these NeRF-based pose estimation methods often require a set of images to be processed simultaneously, limiting their applicability in real-time scenarios. Moreover, they either need a pre-trained neural implicit network or cannot optimize poses and the network concurrently, further constraining their use in SLAM applications.

\subsubsection{Surface Reconstruction from Neural Fields}

Despite the potential of NeRF and its variants to capture the 3D geometry of a scene, these models are implicitly defined in the weights of the neural network. Obtaining an explicit representation of the scene through 3D meshes is desirable for 3D reconstruction applications. Starting with NeRF, a basic approach to achieving coarse scene geometry is to threshold the density predicted by the MLP. More advanced solutions explore three main representations.

\textbf{Occupancy.} This representation models free versus occupied space by replacing alpha values $\alpha_i$ along the ray with a learned discrete function $o(x) \in \{0, 1\}$. Specifically, an occupancy probability $\in [0, 1]$ is estimated and surfaces are obtained by running the marching cubes algorithm \cite{lorensen1998marching}. 

\textbf{Signed Distance Function (SDF).} An alternative method for scene geometry is the signed distance from any point to the nearest surface, yielding negative values inside objects and positive values outside. NeuS \cite{wang2021neus} was the first to revisit the NeRF volumetric rendering engine, predicting the SDF with an MLP as $f(\mathbf{r}(t))$ and replacing $\alpha$ with $\rho(t)$, derived from the SDF as follows:

\begin{equation}
    \rho(t) = \max \big( \frac{-\frac{d\Phi}{dt}(f(\mathbf{r}(t)))}{\Phi(f(\mathbf{r}(t)))}, 0 \big)
\end{equation}
with $\Phi$ being the sigmoid function and $\frac{d\Phi}{dt}$ its derivative. 

\textbf{Truncated Signed Distance Function (TSDF).}
Finally, predicting a truncated SDF with the MLP allows for removing the contribution by any SDF value too far from individual surfaces during rendering. In \cite{azinovic2022neural}, pixel color is obtained as a weighted sum of colors sampled along the ray: 

\begin{equation}
    C(\mathbf{r}) = \frac{\sum_{i=1}^N w_i \textbf{c}_i}{\sum_{i=1}^N w_i}
\end{equation}
with $w_i$ defined, according to truncation distance $t_r$, as

\begin{equation}
   w_i = \Phi \big( \frac{f(\mathbf{r}(t))}{t_r} \big) \cdot \Phi \big( -\frac{f(\mathbf{r}(t))}{t_r} \big)
\end{equation}

\subsubsection{3D Gaussian Splatting (3DGS)}
\begin{figure}[t]

    \centering
    \renewcommand{\tabcolsep}{1pt}
    \begin{tabular}{cc}
        \includegraphics[width=0.20\textwidth]{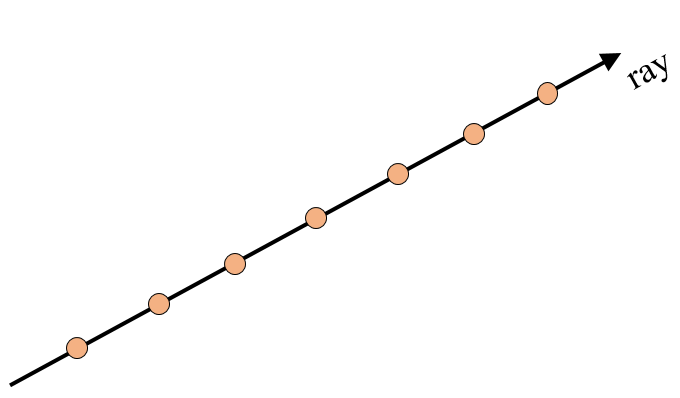} & 
        \includegraphics[width=0.20\textwidth]{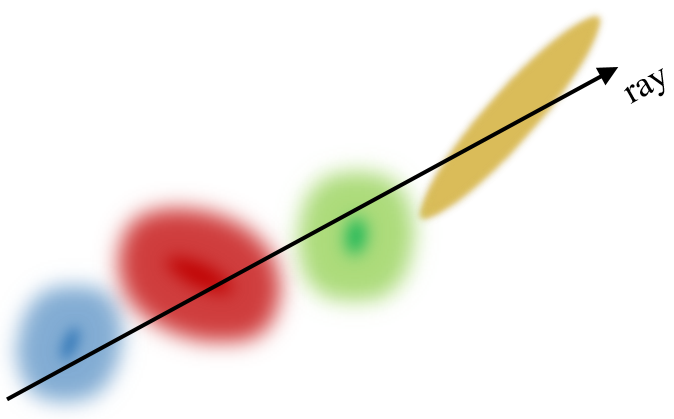} \\
        NeRF \cite{mildenhall2021nerf} & Gaussian Splatting \cite{kerbl20233d}
    \end{tabular}
    \caption{\textbf{NeRF and 3DGS differ conceptually.} (left) NeRF  queries an MLP along the ray, while (right) 3DGS blends Gaussians for the given ray.}\vspace{-0.3cm}
    \label{fig:nerf_vs_3dgs}
\end{figure}

Introduced by Kerbl et al. \cite{kerbl20233d} in 2023, 3DGS is an explicit radiance field technique for efficient and high-quality rendering of 3D scenes. Unlike conventional explicit volumetric representations, such as voxel grids, it provides a continuous and flexible representation for modeling 3D scenes in terms of differentiable 3D Gaussian-shaped primitives.  These primitives are used to parameterize the radiance field and can be rendered to produce novel views. In addition, in contrast to NeRF, which relies on computationally expensive volumetric ray sampling, 3DGS achieves real-time rendering through a tile-based rasterizer. 
This conceptual difference is highlighted in Figure \ref{fig:nerf_vs_3dgs}.
This approach offers improved visual quality and faster training without relying on neural components, while also avoiding computation in empty space. More specifically, starting from multi-view images with known camera poses, 3DGS learns a set $\mathcal{G} = \{g_1, g_2, \dots, g_N\}$ of 3D Gaussians, where $N$ denotes the number of Gaussians in the scene. Each primitive $g_i$, with $1 < i < N$, is parameterized by a full 3D covariance matrix \(\mathbf{\Sigma}_i \in \mathbb{R}^{3 \times 3}\), the mean or center position \(\bm{\mu}_i \in \mathbb{R}^3\), the opacity \(o_i \in [0,1]\), and color $\mathbf{c}_i$ represented by spherical harmonics (SH) for view-dependent appearance, where all the properties are learnable and optimized through back-propagation. This allows for the compact expression of the spatial influence of an individual Gaussian primitive as: 

\begin{equation}
    g_i(\mathbf{x}) = e^{-\frac{1}{2} (\mathbf{x}-\bm{\mu}_i)^\top \mathbf{\Sigma}_i^{-1} (\mathbf{x}-\bm{\mu}_i)}
\end{equation}

Here, the spatial covariance $\mathbf{\Sigma}$ defines an ellipsoid and it is computed as $\mathbf{\Sigma} = \mathbf{RSS^{\top}R^{\top}}$, where $\mathbf{S} \in \mathbb{R}^3$ is the spatial scale and $\mathbf{R} \in \mathbb{R}^{3 \times 3}$ represents the rotation, parameterized by a quaternion. For rendering, 3DGS operates akin to NeRF but diverges significantly in the computation of blending coefficients. Specifically, the process involves first projecting 3D Gaussian points onto a 2D image plane, a process commonly referred to as ``splatting''. This is done expressing the projected 2D covariance matrix and center as $\mathbf{\Sigma}' = \mathbf{JW \Sigma W^T J^T}$ and $\bm{\mu}' = \mathbf{JW}\bm{\mu}$, where $\mathbf{W}$ represents the viewing transformation, and $\mathbf{J}$ is the Jacobian of the affine approximation of the projective transformation. Consequently, 3DGS  computes the final pixel color $C$ by blending 3D Gaussian splats that overlap at a given pixel, sorted by their depth:

\begin{equation}
    C = \sum_{i \in \mathcal{N}} \mathbf{c}_i \alpha_i \prod_{j=1}^{i-1} (1 - \alpha_j)
\end{equation}
where the final opacity $\alpha_i$ is the multiplication result of the learned opacity $o_i$ and the Gaussian:

\begin{equation}
    \alpha_i = o_i \exp\left(-\frac{1}{2} (\mathbf{x}' - \bm{\mu}'_i)^\top \mathbf{\Sigma}'^{-1}_i (\mathbf{x}' - \bm{\mu}'_i)\right)
\end{equation}
where $\mathbf{x}'$ and $\bm{\mu}'_i$ are coordinates in the projected space. 
Similarly, the depth $D$ is rendered as:

\begin{equation}
    D = \sum_{i \in \mathcal{N}} d_i \alpha_i \prod_{j=1}^{i-1} (1 - \alpha_j)
\end{equation}
Here, $d_i$ refers to the depth of the center of the $i$-th 3D Gaussian, obtained by projecting onto the z-axis in the camera coordinate system.
For optimization, instead, the process begins with parameter initialization from SfM point clouds or random values, followed by Stochastic Gradient Descent (SGD) using an L1 and D-SSIM loss function against ground truth and render views. Additionally, periodic adaptive densification handles under- and over-reconstruction by adjusting points with significant gradients and removing low-opacity points, refining scene representation and reducing rendering errors. For more details on 3DGS and related works, refer to \cite{chen2024survey,wu2024recent,fei20243d}.

\subsection{Datasets}
\label{sec:datasets}

\begin{figure}[]
    \centering
    \renewcommand{\tabcolsep}{1pt}
    \begin{tabular}{cc}
        \includegraphics[width=0.24\textwidth]{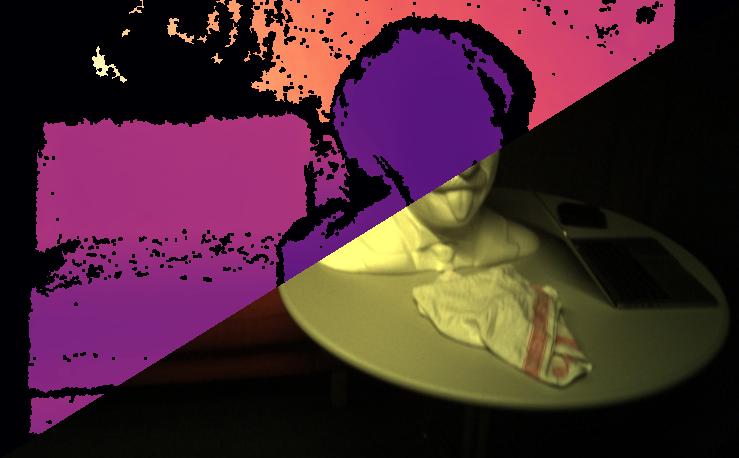} & 
        \includegraphics[width=0.24\textwidth]{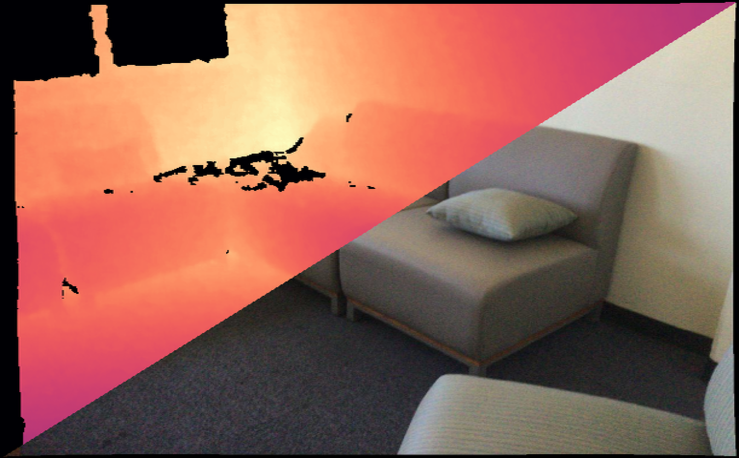} \\
        (a) ETH3D-SLAM \cite{schops2019bad} & (b) ScanNet \cite{dai2017scannet} \\
        \includegraphics[height=0.16\textwidth,width=0.24\textwidth]{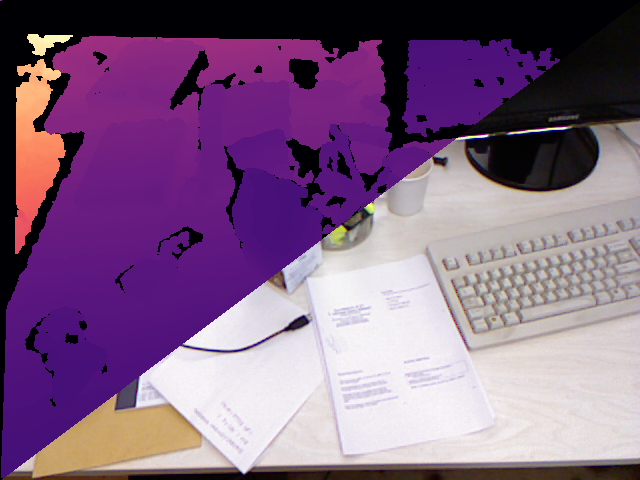} &
        \includegraphics[width=0.24\textwidth, height=0.16\textwidth]{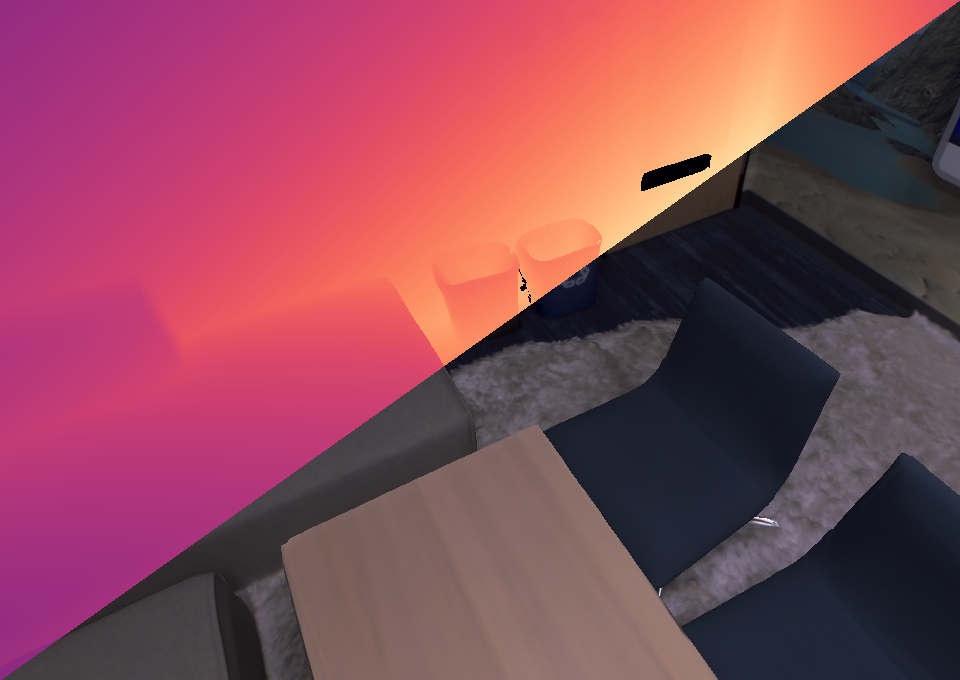}\\
        (c) TUM RGB-D \cite{tum} & (d) Replica  \cite{replica19arxiv} \\
    \end{tabular}
    \caption{\textbf{Qualitative Comparison of Key SLAM Datasets.} RGB-D images from: (a) ETH3D-SLAM \cite{schops2019bad}, (b) ScanNet \cite{dai2017scannet}, (c) TUM RGB-D \cite{tum}, and (d) Replica \cite{replica19arxiv}. 
    }\vspace{-0.3cm}
    \label{fig:SLAM_datasets}
\end{figure}

This section summarizes datasets commonly used in recent SLAM methodologies, covering various attributes such as sensors, ground truth accuracy, and other key factors, in both indoor and outdoor environments. Figure \ref{fig:SLAM_datasets} presents qualitative examples from diverse datasets, which will be introduced in the remainder.

The \textbf{TUM RGB-D\cite{tum}\footnote{\url{https://cvg.cit.tum.de/data/datasets/RGB-D-dataset}}} dataset comprises RGB-D sequences with annotated camera trajectories, recorded using two platforms: handheld and robot, providing a diverse range of motions. The dataset features 39 sequences, some with loop closures. Core elements include color and depth images from a Microsoft Kinect sensor, captured at 30 Hz and $640 \times 480$ resolution. Ground-truth trajectories are derived from a motion-capture system with eight high-speed cameras operating at 100 Hz. The versatility of the dataset is demonstrated through various trajectories in typical office environments and an industrial hall, encompassing diverse translational and angular velocities.

The \textbf{ScanNet \cite{dai2017scannet}\footnote{\url{http://www.scan-net.org/}}} dataset provides a collection of real-world indoor RGB-D acquisitions, featuring 2.5 million images from 1513 scans in 707 unique spaces. In particular, it includes estimated calibration parameters, camera poses, 3D surface reconstructions, textured meshes, detailed semantic segmentations at the object-level, and aligned CAD models.

The development process involved the creation of a user-friendly capture pipeline using a custom RGB-D capture setup with structure sensors attached to handheld devices such as iPads. The subsequent offline processing phase resulted in comprehensive 3D scene reconstructions, complete with available 6-DoF camera poses and semantic labels. Note that camera poses in ScanNet are derived from the BundleFusion system \cite{bundlefusion}, which may not be as accurate as alternatives such as TUM RGB-D. 

The \textbf{Replica \cite{replica19arxiv}\footnote{\url{https://github.com/facebookresearch/Replica-Dataset}}} dataset features 18 photorealistic 3D indoor scenes with dense meshes, HDR textures, semantic data, and reflective surfaces. It spans different scene categories, includes 88 semantic classes, and incorporates 6 scans of a single space capturing different furniture arrangements and temporal snapshots. Reconstruction involved a custom-built RGB-D capture rig with synchronized IMU, RGB, IR, and wide-angle grayscale sensors, accurately fusing raw depth data through 6 degrees of freedom (DoF) poses.
Although the original data was captured in the real world, the portion of the dataset used for SLAM evaluation is synthetically generated from the accurate meshes produced during reconstruction. Consequently, synthetic sequences lack real-world characteristics like specular highlights, autoexposure, blur, and more.

The \textbf{KITTI \cite{Geiger2012CVPR}\footnote{\url{https://www.cvlibs.net/datasets/kitti/}}} dataset serves as a popular benchmark for evaluating stereo, optical flow, visual odometry/SLAM algorithms, among others. Acquired from a vehicle equipped with stereo cameras, Velodyne LiDAR, GPS and inertial sensors, the dataset contains 42,000 stereo pairs and LiDAR pointclouds from 61 scenes representing autonomous driving scenarios.  The KITTI odometry dataset, with 22 LiDAR scan sequences, contributes to the evaluation of odometry methods using LiDAR data. 

The \textbf{Newer College \cite{ramezani2020newer}\footnote{\url{https://arxiv.org/pdf/ori.ox.ac.uk/datasets/newer-college-dataset}}} dataset comprises sensor data captured during a 2.2 km walk around New College, Oxford. It includes information from a stereoscopic-inertial camera, a multi-beam 3D LiDAR with inertial measurements, and a tripod-mounted survey-grade LiDAR scanner, generating a detailed 3D map with around 290 million points. The dataset provides a 6 DoF ground truth pose for each LiDAR scan, accurate to approximately 3 cm. The dataset encompasses diverse environments, including built spaces, open areas, and vegetated zones.

\subsubsection{Other Datasets}

Moreover, we draw attention to less-utilized alternative datasets in recent SLAM research.

The \textbf{ETH3D-SLAM}\cite{schops2019bad}\footnote{\url{https://www.eth3d.net/slam_overview}} dataset includes videos from a custom camera rig, suitable for assessing visual-inertial mono, stereo, and RGB-D SLAM. It features 56 training datasets, 35 test datasets, and 5 independently captured training sequences using SfM techniques for ground truth.

The \textbf{EuRoC MAV \cite{burri2016euroc}\footnote{\url{https://projects.asl.ethz.ch/datasets/doku.php?id=kmavvisualinertialdatasets}}} dataset offers synchronized stereo images, IMU, and accurate ground truth for a micro aerial vehicle. It supports visual-inertial algorithm design and evaluation in diverse conditions, including an industrial setting with millimeter-accurate ground truth and a room for 3D environment reconstruction.

The \textbf{7-scenes \cite{glocker2013real}\footnote{\url{http://research.microsoft.com/7-scenes/}}} dataset, created for relocalization performance evaluation, was recorded using a Kinect at $640 \times 480$ resolution. Ground truth poses were obtained through KinectFusion \cite{kinectfusion}. Sequences from different users were divided into two sets—one for simulating keyframe harvesting and the other for error calculation. The dataset presents challenges such as specularities, motion blur, lighting conditions, flat surfaces, and sensor noise.

The \textbf{ScanNet++ \cite{yeshwanth2023scannet++}\footnote{\url{https://cy94.github.io/scannetpp/}}} dataset comprises 460 high-resolution 3D indoor scene reconstructions, dense semantic annotations, DSLR images, and iPhone RGB-D sequences. Captured with a high-end laser scanner at sub-millimeter resolution, each scene includes annotations for over 1,000 semantic classes, addressing label ambiguities and introducing new benchmarks for 3D semantic scene understanding and novel view synthesis.

The \textbf{NeuralRGBD \cite{azinovic2022neural}\footnote{\url{https://github.com/dazinovic/neural-rgbd-surface-reconstruction}}} dataset consists of 10 synthetic scenes with varying complexity and materials. It provides color and depth images rendered using BlenderProc \cite{denninger2019blenderproc}, simulating real-world depth sensor artifacts. Camera trajectories, initially estimated with BundleFusion \cite{bundlefusion}, are designed to scan only portions of scenes, mimicking real-world scanning behavior.

The \textbf{Bonn Dataset\cite{palazzolo2019refusion}\footnote{\url{https://www.ipb.uni-bonn.de/data/rgbd-dynamic-dataset/}}} offers 24 highly dynamic and 2 static RGB-D sequences featuring people manipulating objects. Recorded with an ASUS Xtion Pro LIVE sensor and an Optitrack Prime 13 system for ground truth trajectories, it follows the TUM RGB-D format. A Leica BLK360 provides ground truth 3D pointclouds of the static environment, specifically designed for evaluating SLAM in dynamic scenes.

\textbf{Additional Datasets.} For an exhaustive survey of specialized SLAM-related datasets beyond those mentioned, readers can refer to the work by Liu et al. \cite{liu2021simultaneous}. This paper provides an in-depth exploration of a wide range of datasets designed to facilitate research and benchmarking.

\input{chapters/tables/overview}

\subsection{Evaluation Metrics}
\label{sec:metrics}
The evaluation of SLAM systems typically employs several metrics across domains like 3D reconstruction, 2D depth estimation, trajectory estimation, and view synthesis to assess the effectiveness of methods against ground truth data.

\textit{A. Mapping}. Metrics assessing the quality of 3D reconstruction and 2D depth estimation include:
\begin{itemize}
    \item \textbf{Accuracy (cm)$\downarrow$:} Computes the average distance between sampled points from the reconstructed mesh and the nearest ground-truth point.
    \item \textbf{Completion (cm)$\downarrow$:} Measures the average distance between sampled points from the ground-truth mesh and the nearest reconstructed.
    \item \textbf{Precision ($\%$)$\uparrow$:} Indicates the proportion of points within the reconstructed mesh with Accuracy under a distance threshold $d$.
    \item \textbf{Recall ($\%$)$\uparrow$:} Indicates the proportion of points within the reconstructed mesh with Completion under a distance $d$. It is often referred to as \textit{Completion Ratio}.
    \item \textbf{F-Score ($\%$)$\uparrow$:} An aggregate score defined as the harmonic mean between Precision and Recall.

    \item \textbf{L1-Depth (cm)$\downarrow$:} Following \cite{zhu2022nice}, it computes the absolute difference between depth maps obtained from randomly sampled viewpoints from the reconstructed and the corresponding ground truth meshes respectively.
\end{itemize}

\textit{B. Tracking}. Metrics for pose estimation, crucial for
tracking performance, primarily include:

\begin{itemize}
    \item \textbf{Absolute Trajectory Error (ATE)(cm) $\downarrow$:} Evaluates trajectory estimation accuracy by measuring the average Euclidean translation distance between corresponding poses in estimated and ground truth trajectories, often reported in terms of Root Mean Square Error (RMSE). As both trajectories can be specified in arbitrary coordinate frames, alignment is required. Importantly, this metric focuses solely on the translation component.
\end{itemize}

\textit{C. View Synthesis}. The evaluation of view synthesis relies mainly on three visual quality assessment metrics:
\begin{itemize}
    \item \textbf{Peak Signal to Noise Ratio (PSNR)$\uparrow$:} Measures image quality by evaluating the ratio between the maximum pixel value and the root mean squared error, usually expressed in terms of the logarithmic decibel scale.
    \item \textbf{Structural Similarity Index Measure (SSIM\cite{wang2004image})$\uparrow$:} Assesses image quality by examining the similarities in luminance, contrast, and structural information among patches of pixels.
    \item \textbf{Learned Perceptual Image Patch Similarity (LPIPS\cite{zhang2018unreasonable})$\downarrow$ :} Utilizes learned convolutional features to assess image quality based on feature map mean squared error across layers.
\end{itemize}

\textit{D. Semantic Segmentation}. For SLAM methods that additionally estimate semantic information of the scene, the following metric is included to evaluate the performance of the semantic segmentation:
\begin{itemize}
\item \textbf{Mean Intersection over Union (mIoU)$\uparrow$:} mIoU is a widely used metric for evaluating semantic segmentation performance. It is computed by calculating the IoU for each class and then taking the average across all classes. IoU is defined as the ratio of the intersection between the predicted and ground truth segmentation masks to their union. A higher mIoU indicates better semantic segmentation accuracy.
\end{itemize}

\begin{figure*}[t]
  \centering
  \includegraphics[width=1.\textwidth]{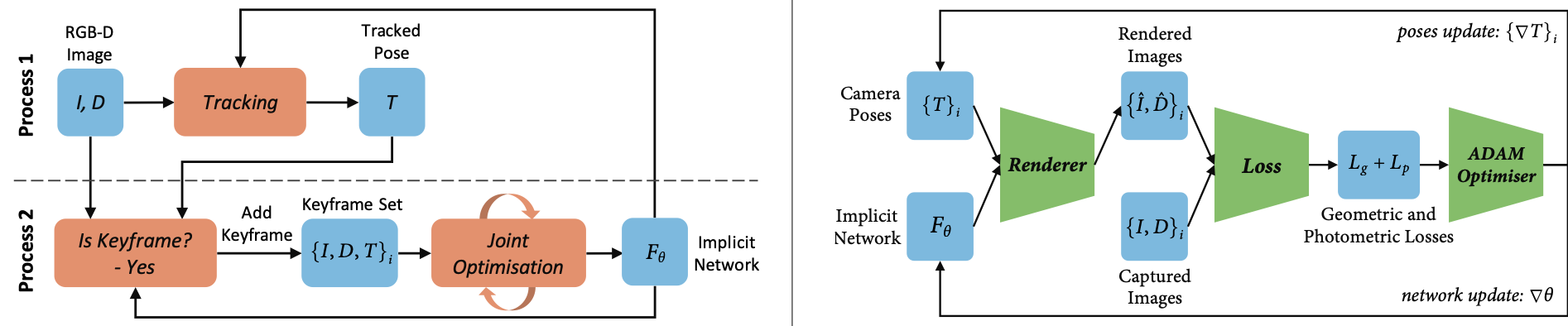}
  \caption{\textbf{Overview of iMap \cite{sucar2021imap}, the Pioneering Approach in Neural Implicit-based SLAM.} (Left) The illustration depicts two concurrent processes:  \textit{tracking}, optimizing the current frame's pose within the locked network;  \textit{mapping}, jointly refining the network and camera poses of selected keyframes. (Right) Jointly optimizing scene network parameters and camera poses for keyframes using differentiable rendering functions.  Figure from \cite{sucar2021imap}.}
  \label{fig:imap_overview}
\end{figure*}

%% file: chapters/tables/overview.tex
\begin{table*}[t]
\centering
\caption{\textbf{SLAM Systems Overview.} We categorize the different methods into main RGB-D, RGB, and LiDAR-based frameworks. In the leftmost column, we identify sub-categories of methods sharing specific properties, detailed in Sections \ref{sec:vanilla_rgb} to \ref{sec:multimodal_lidar}. Then, for each method, we report, from the second leftmost column to the second rightmost, the method name and publication venue, followed by (a) the input modalities they can process: RGB, RGB-D, D (\eg LiDAR, ToF, Kinect, etc.), stereo, IMU, or events; (b) mapping properties: scene encoding and geometry representations learned by the model; (c) additional outputs learned by the method, such as object/semantic segmentation, or uncertainty modeling (Uncert.); (d) tracking properties related to the adoption of a frame-to-frame or frame-to-model approach, the utilization of external trackers, Global Bundle Adjustment (BA), or Loop Closure; (e) advanced design strategies, such as modeling sub-maps or dealing with dynamic environments (Dyn. Env.); (f) the use of additional priors.  Finally, we report the link to the project page or source code in the rightmost column.
$\dagger${} indicates code or webpage not released yet.}
\setlength\tabcolsep{4pt}
\resizebox{1.\textwidth}{!}{
    \begin{tabular}{cc}
    \begin{tabular}{l}
        \\
        \Xhline{2pt}
         \multirow{2}{*}{Section} \\
         \\
        \Xhline{1pt}
         \cellcolor{maincategories} \\
        \Xhline{1pt}
         \multirow{19}{*}{\cref{sec:vanilla-RGB-D}} \\ \\ \\ \\ \\ \\ \\ \\ \\ \\ \\ \\ \\ \\ \\ \\ \\ \\ \\ \hline
         \multirow{9}{*}{\cref{sec:3DGS}} \\ \\ \\ \\ \\ \\ \\ \\ \\ \hline
         \multirow{10}{*}{\cref{sec:sub-maps}} \\ \\ \\ \\ \\ \\ \\ \\ \\ \\ \hline
         \multirow{9}{*}{\cref{sec:segmentation-priors}} \\ \\ \\ \\ \\ \\ \\ \\ \\ \hline
         \multirow{7}{*}{\cref{sec:dynamic-envs}} \\ \\ \\ \\ \\ \\ \\ \hline
         \multirow{4}{*}{\cref{sec:uncertainty}} \\ \\ \\ \\ \hline
         \cref{sec:event} \\ \Xhline{1pt}
         \cellcolor{maincategories} \\ \Xhline{1pt}
         \multirow{3}{*}{\cref{sec:vanilla_rgb} } \\ \\ \\ \hline
         \cref{sec:3dgs_rgb} \\ \hline
         \multirow{7}{*}{\cref{sec:priors_driver}} \\ \\ \\ \\ \\ \\ \\ \hline
         \multirow{2}{*}{\cref{sec:segmentation-priors_rgb}} \\ \\ \hline
         \cref{sec:uncertainty_rgb} \\ \Xhline{1pt}
         \cellcolor{maincategories} \\ \Xhline{1pt}
         \multirow{4}{*}{\cref{sec:vanilla_lidar} } \\ \\ \\ \\ \hline
         \multirow{2}{*}{\cref{sec:multimodal_lidar}} \\ \\
         \Xhline{2pt}
    \end{tabular}
    & \hspace{-0.5cm}
    \rowcolors{2}{salmon}{white}
    \begin{tabular}{|l|c|cccccc|cc|cc|ccccc|cc|c|c}
    \multicolumn{2}{c}{} & \multicolumn{6}{c}{(a)} & \multicolumn{2}{c}{(b)} & \multicolumn{2}{c}{(c)} & \multicolumn{5}{c}{(d)} & \multicolumn{2}{c}{(e)} & \multicolumn{1}{c}{(f)} \\
    \Xhline{2pt}
    \multirow{2}{*}{\rowcolors{white}{}{}Method} & \multirow{2}{*}{Venue} & \multirow{2}{*}{RGB-D} & \multirow{2}{*}{RGB} & \multirow{2}{*}{D} & \multirow{2}{*}{Stereo} & \multirow{2}{*}{IMU} & Event & Scene & Geometry & Obj/Sem. & \multirow{2}{*}{Uncert.} & Frame-to & Frame-to & External & Global & Loop & Sub- & Dyn. & \multirow{2}{*}{Extra Priors} & \multirow{2}{*}{Link} \\
    & & & & & & & Camera & Encoding & Representation & Segment. & & Model & Frame & Tracker & BA & Closure & Maps & Env. & & \\
    \Xhline{1pt}
    \rowcolor{maincategories}\multicolumn{21}{c}{\hyperlink{RGB-D}{\textcolor{magenta}{RGB-D (Sec. \ref{sec:RGB-D})}}} \\
    \Xhline{1pt}
    
   \hyperlink{iMAP}{\textcolor{magenta}{iMAP}}~\cite{sucar2021imap} & ICCV 2021 & \checkmark & &  &   &   &   & MLP & Density  &   &   &  \checkmark &  &   &   &   &  &  &   & \href{https://edgarsucar.github.io/iMAP/}{WebPage} \\   
   
   \hyperlink{NICE-SLAM}{\textcolor{magenta}{NICE-SLAM}}~\cite{zhu2022nice} & CVPR 2022 & \checkmark &  &   &   &   &   & Hier. Grid + MLP & Occupancy   &   &   & \checkmark &  &  &   &   &   &   &  & \href{https://github.com/cvg/nice-slam}{Code} \\
   
   \hyperlink{Vox-Fusion}{\textcolor{magenta}{Vox-Fusion}}~\cite{yang2022vox} & ISMAR 2022 & \checkmark &  &   &   &   &   &  Octree Grid + MLP & SDF   &   &  & \checkmark &   &  & &  &  &   &   & \href{https://github.com/zju3dv/Vox-Fusion}{Code} \\
   
   \hyperlink{ESLAM}{\textcolor{magenta}{ESLAM}}~\cite{ESLAM} & CVPR 2023 & \checkmark &   &  &   &   &   & Feature Planes + MLP & TSDF   &   &   &  \checkmark &  & &  &   &   &   &  & \href{https://github.com/idiap/ESLAM}{Code} \\
   
   \hyperlink{Co-SLAM}{\textcolor{magenta}{Co-SLAM}}~\cite{Wang_2023_CVPR} & CVPR 2023 & \checkmark &   &  &   &   &   &  Hash Grid + MLP & SDF  &   &   &  \checkmark &  &  & \checkmark & &  &   &   & \href{https://github.com/HengyiWang/Co-SLAM}{Code} \\
   
   \hyperlink{GO-SLAM}{\textcolor{magenta}{GO-SLAM}}~\cite{zhang2023go} & ICCV 2023 & \checkmark & \checkmark &  &  \checkmark &  &  & Hash Grid + MLP & SDF  &   &  & & \checkmark & DROID \cite{droidslam} & \checkmark & \checkmark &   &  &   & \href{https://youmi-zym.github.io/projects/GO-SLAM/}{Code} \\
   
   \hyperlink{Point-SLAM}{\textcolor{magenta}{Point-SLAM}}~\cite{Sandström2023ICCV} & ICCV 2023 & \checkmark &  &  &   &   &   & Neural Points + MLP & Occupancy   &   & & \checkmark &  &   &   &   &   &   &   & \href{https://github.com/eriksandstroem/Point-SLAM}{Code} \\
   
   \hyperlink{ToF-SLAM}{\textcolor{magenta}{ToF-SLAM}}~\cite{tofslam} & ICCV 2023 & \checkmark &  &  &  &  & &  Hier. Grid + MLP & SDF  &   &  & \checkmark &   & &  &   &   &   &   & \href{https://zju3dv.github.io/tof_slam/}{WebPage}  \\
   
   \hyperlink{ADFP}{\textcolor{magenta}{ADFP}}~\cite{Hu2023LNI-ADFP} & NeurIPS 2023 & \checkmark &  &   &   &   &   & Hier. Grid + MLP & Occupancy   &   &  & \checkmark &  &   &   &   &   &   &  & \href{https://machineperceptionlab.github.io/Attentive_DF_Prior/}{Code} \\
   
   \hyperlink{MLM-SLAM}{\textcolor{magenta}{MLM-SLAM}}~\cite{li2023end} & RAL 2023 & \checkmark &   &  &   &   &  & MLP & Occupancy  &  &  & \checkmark &  &  & &  &   &   &   & \\
   
   \hyperlink{Plenoxel-SLAM}{\textcolor{magenta}{Plenoxel-SLAM}}~\cite{teigen2024rgb} & WACV 2024 & \checkmark &  &   &   &   &   & Plenoxels  & Density &   &   & \checkmark &   &   &  &  &   &   &   & \href{https://github.com/ysus33/RGB-D_Plenoxel_Mapping_Tracking}{Code} \\
   
   \hyperlink{Structerf-SLAM}{\textcolor{magenta}{Structerf-SLAM}}~\cite{wang2024structerf} & C .\& G. 2024 & \checkmark &  &   &   &   &   & Hier. Grid  & Occupancy &   &   & & \checkmark  & ORB2 \cite{mur2017orb}  &  &  &   &   & Super-pixel Segmentation \cite{felzenszwalb2004efficient}  &  \\

   \hyperlink{KN-SLAM}{\textcolor{magenta}{KN-SLAM}}~\cite{wu2024kn} & TIM 2024 & \checkmark &  &   &   &   &   & Hier. Grid  & Occupancy &  &  &  & \checkmark &  EPNP \cite{lepetit2009ep} & & \checkmark &  & & HF-Net \cite{sarlin2019coarse}  & \\

    \hyperlink{SLAIM}{\textcolor{magenta}{SLAIM}}~\cite{cartillier2024slaim} & CVPRW 2024 & \checkmark & & & & & & Hier. Grid + MLP & Density & & & \checkmark & & & \checkmark & & & & & \href{vincentcartillier.github.io/slaim.html}{WebPage} \\
    \hyperlink{IBD-SLAM}{\textcolor{magenta}{IBD-SLAM}}~\cite{IBDSLAM} & CVPR 2024 & \checkmark & & & & & & Image Feat. + MLP & Density & & & & \checkmark & SuperPoint \cite{detone2018superpoint} \& SuperGlue \cite{sarlin20superglue} & & & & & GMFLow \cite{xu2022gmflow} &  \href{https://visual-ai.github.io/ibd-slam}{WebPage}$\dagger$ \\
    \hyperlink{VPE-SLAM}{\textcolor{magenta}{VPE-SLAM}}~\cite{vpeslam} & ICRA 2024 & \checkmark & & & & & & Octree Grid + MLP & TSDF & & & \checkmark & & & \checkmark & & & & & \href{https://github.com/NeuCV-IRMI/VPE-SLAM}{Code}$\dagger$ \\
    \hyperlink{HERO-SLAM}{\textcolor{magenta}{HERO-SLAM}}~\cite{heroslam} & ICRA 2024 & \checkmark & & & & & & Hash Grid + MLP & TSDF & & & \checkmark & & & & & & & & \href{https://hero-slam.github.io/}{WebPage} \\
    \hyperlink{LRSLAM}{\textcolor{magenta}{LRSLAM}}~\cite{park2024lrslam} & ECCV 2024 & \checkmark & & & && & Feature Planes + MLP & SDF & & & \checkmark & & & & & & & & \\
   \hyperlink{LCP-Fusion}{\textcolor{magenta}{LCP-Fusion}}~\cite{wang2024lcp} & IROS 2024 & \checkmark & & & & & & Octree Grid + MLP & SDF & & & \checkmark & & & \checkmark & & & & & \href{https://github.com/laliwang/LCP-Fusion}{Code}$\dagger$ \\

   \hline
    
   \hyperlink{MonoGS}{\textcolor{magenta}{MonoGS}}~\cite{matsuki2023gaussian} & CVPR 2024 & \checkmark & \checkmark &   &  &   &   &  3D Gaussians & Density &   &   & \checkmark &   &  & &   &   &   &   & \href{https://rmurai.co.uk/projects/GaussianSplattingSLAM/}{WebPage}\\

   \hyperlink{Photo-SLAM}{\textcolor{magenta}{Photo-SLAM}}~\cite{huang2023photo} & CVPR 2024 & \checkmark & \checkmark &  &  \checkmark &   &   &   3D Gaussians & Density &  &  & & \checkmark & ORB3 \cite{campos2021orb} &  & \checkmark &  &   &  & \\
   
   \hyperlink{SplaTAM}{\textcolor{magenta}{SplaTAM}}~\cite{keetha2023splatam} & CVPR 2024 & \checkmark &   &  &   &   &   &  3D Gaussians & Density &   &   & \checkmark &  &   &   &   & &  &   & \href{https://github.com/spla-tam/SplaTAM}{Code} \\
   
   \hyperlink{GS-SLAM}{\textcolor{magenta}{GS-SLAM}}~\cite{yan2023gs} & CVPR 2024 & \checkmark &  &  &  &  &  &  3D Gaussians &  Density &   &   & \checkmark &  &  &   &   &   &   &   & \\

   \hyperlink{GS-ICP SLAM}{\textcolor{magenta}{GS-ICP SLAM}}~\cite{ha2024rgbd} & ECCV 2024 & \checkmark &  &   &  &   &   &  3D Gaussians & Density &   &   & \checkmark &   & G-ICP \cite{segal2009generalized} &  &   &   &  &   & \href{https://github.com/Lab-of-AI-and-Robotics/GS_ICP_SLAM}{Code}\\ 

   \hyperlink{HF-GS SLAM}{\textcolor{magenta}{HF-GS SLAM}}~\cite{sun2024high} & IROS 2024 & \checkmark &  &  &  &   &   &  3D Gaussians & Density &   &   & \checkmark &   &  &  &   &   &  &   & \\ 
   
   \hyperlink{CG-SLAM}{\textcolor{magenta}{CG-SLAM}}~\cite{hu2024cg} & ECCV 2024 & \checkmark &  &  &  &   &   &  3D Gaussians & Density &   & \checkmark  & \checkmark &   &  &  &   &   &  &  NetVLAD \cite{arandjelovic2016netvlad} & \href{https://zju3dv.github.io/cg-slam}{Code}\\ 

   \hyperlink{MM3DGS-SLAM}{\textcolor{magenta}{MM3DGS-SLAM}}~\cite{sun2024multimodal3dgsslam} & IROS 2024 & \checkmark & \checkmark &  &  & \checkmark  &   &  3D Gaussians & Density &   &  & \checkmark &   &  &  &   &   &  & DPT \cite{Ranftl2021} & \href{https://vita-group.github.io/MM3DGS-SLAM/}{WebPage}\\ 

   \hyperlink{RTG-SLAM}{\textcolor{magenta}{RTG-SLAM}}~\cite{peng2024rtg} & SIGGRAPH 2024 & \checkmark & & & & & & 3D Gaussians & Density & & & \checkmark & & & & & & & & \href{https://github.com/MisEty/RTG-SLAM}{Code} \\
     
    \hline
    
   \hyperlink{MeSLAM}{\textcolor{magenta}{MeSLAM}}~\cite{kruzhkov2022meslam} & SMC 2022 & \checkmark &  &   &   &   &   & MLP & Density  &   &   & & \checkmark &  C-ICP \cite{park2017colored} &  &  & \checkmark &  &  &  \\
   
   \hyperlink{CP-SLAM}{\textcolor{magenta}{CP-SLAM}}~\cite{hu2023cp} & NeurIPS 2023 & \checkmark &  &  &   &  &   & Neural Points + MLP & Occupancy  &  &  & \checkmark &  & & \checkmark  & \checkmark & \checkmark &  &  &  \\
   
   \hyperlink{NISB-Map}{\textcolor{magenta}{NISB-Map}}~\cite{xiang2023nisb} & RAL 2023 & \checkmark &  &  &  &  &  & MLP & Density  &   &   & & \checkmark & Any &   &   & \checkmark &   &   &   \\
   
   \hyperlink{Multiple-SLAM}{\textcolor{magenta}{Multiple-SLAM}}~\cite{liu2023efficient} & TIV 2023 & \checkmark &  &  &  &  &  & Octree Grid + MLP & SDF  &   &   & \checkmark &  &  &   &   & \checkmark &   &   &   \\
   
   \hyperlink{MIPS-Fusion}{\textcolor{magenta}{MIPS-Fusion}}~\cite{tang2023mips} & TOG 2023 & \checkmark &  &  &  &  &  &  MLP &  TSDF  &   & \checkmark & \checkmark &  & &   &  \checkmark & \checkmark &   &   &   \\

   \hyperlink{NGEL-SLAM}{\textcolor{magenta}{NGEL-SLAM}}~\cite{mao2023ngel} & ICRA 2024 & \checkmark &  &  &  &  &  & Octree Grid + MLP & Occupancy  &  & \checkmark & & \checkmark & ORB3 \cite{campos2021orb} & \checkmark & \checkmark & \checkmark &  &  & \\
   
   \hyperlink{PLGSLAM}{\textcolor{magenta}{PLGSLAM}}~\cite{deng2023plgslam} & CVPR 2024 & \checkmark &  &  &   &   &   & Feature Planes + MLP & SDF  &   &   & \checkmark &  &  & & & \checkmark & & &  \\ 
   
   \hyperlink{Loopy-SLAM}{\textcolor{magenta}{Loopy-SLAM}}~\cite{liso2024loopy} & CVPR 2024 & \checkmark &  &  &  &  &  &  Neural Points + MLP &  Occupancy  &   &  & \checkmark &  & &  &  \checkmark & \checkmark &   &   & \href{https://notchla.github.io/Loopy-SLAM/}{Code} \\
   
   \hyperlink{NEWTON}{\textcolor{magenta}{NEWTON}}~\cite{matsuki2024newton}  & RAL 2024 & \checkmark & \checkmark &  &  &  &  & Hash Grid + MLP & Density &  &  & & \checkmark & ORB2 \cite{mur2017orb} &   & \checkmark & \checkmark &  &  &  \\

   \hyperlink{MAN-SLAM}{\textcolor{magenta}{MAN-SLAM}}~\cite{dengmulti}  & IROSw 2024 & \checkmark & & & & & & Features Planes + MLP & & TSDF & & & \checkmark & DROID \cite{droidslam} & \checkmark & \checkmark & \checkmark & & & \\

    \hline
    
   \hyperlink{iLabel}{\textcolor{magenta}{iLabel}}~\cite{zhi2022ilabel} & RAL 2023 & \checkmark &  &  &  &  &  & MLP & Density  &  \checkmark & \checkmark & \checkmark &  &  &   &   &  &   & User & \href{https://edgarsucar.github.io/ilabel/}{WebPage} \\
   
   \hyperlink{FR-Fusion}{\textcolor{magenta}{FR-Fusion}}~\cite{mazur2023feature} & ICRA 2023 & \checkmark &  &  &  &  &  & MLP & Density  &  \checkmark &  & \checkmark &  &  &   &   &  &   & User \& EfficientNet \cite{tan2019efficientnet}/DINO \cite{caron2021emerging} & \href{https://makezur.github.io/FeatureRealisticFusion/}{WebPage} \\
   
   \hyperlink{vMap}{\textcolor{magenta}{vMap}}~\cite{Kong_2023_CVPR} & CVPR 2023 & \checkmark &  &   &   &   &  & MLP & Occupancy  & \checkmark &   & \checkmark &   &   &   &   &  & & & \href{https://kxhit.github.io/vMAP}{Code} \\
   
   \hyperlink{SNI-SLAM}{\textcolor{magenta}{SNI-SLAM}}~\cite{zhu2023sni} & CVPR 2024 & \checkmark &  &  &  &  &  & Hier. Grid + MLP & TSDF   &  \checkmark &   & \checkmark &   &   &   &  & &  &  Dinov2 \cite{oquab2023dinov2} &  \\

   \hyperlink{DNS-SLAM}{\textcolor{magenta}{DNS-SLAM}}~\cite{li2023dns} & IROS 2024 & \checkmark &  &  &  &  &  & Hash Grid + MLP & Occupancy   &  \checkmark & & \checkmark &  &   & \checkmark &   &   &   & GT semantics  &  \\
   
   \hyperlink{SGS-SLAM}{\textcolor{magenta}{SGS-SLAM}}~\cite{li2024sgsslam} & ECCV 2024 & \checkmark &  &   &   &  &   &  3D Gaussians & Density &  \checkmark & & \checkmark &  &  &   &  \checkmark &   &   & GT semantics  & \\

   \hyperlink{NEDS-SLAM}{\textcolor{magenta}{NEDS-SLAM}}~\cite{ji2024neds} & RAL 2024 & \checkmark &  &   &   &  &   &  3D Gaussians & Density &  \checkmark & & \checkmark &  &  &   &  &   &   &  Depth Anything \cite{yang2024depth} + GT semantics & \\

   \hyperlink{GS3LAM}{\textcolor{magenta}{GS3LAM}}~\cite{GS3LAM} & MM 2024 & \checkmark & & & & & & 3D Gaussians & Density & \checkmark & & \checkmark & & & & & & & GT semantics & \href{https://github.com/lif314/GS3LAM}{Code} \\
   \hyperlink{NIS-SLAM}{\textcolor{magenta}{NIS-SLAM}}~\cite{nis_slam} & TVCG 2024 & \checkmark & & & & & & Tetra. Features + MLP & SDF & \checkmark & & \checkmark & & & \checkmark & & & & Mask2Former \cite{cheng2021maskformer} & \href{https://zju3dv.github.io/nis_slam/}{WebPage} \\

    \hline
    
   \hyperlink{DN-SLAM}{\textcolor{magenta}{DN-SLAM}}~\cite{ruan2023dn} & Sensors J. 2023 & \checkmark &  &  &  & & &  Hash Grid + MLP & Density  &  &   &  & \checkmark & ORB3 \cite{campos2021orb} &   &   &  &  \checkmark &  SAM \cite{kirillov2023segment} &  \\
   
   \hyperlink{DynaMoN}{\textcolor{magenta}{DynaMoN}}~\cite{karaoglu2023dynamon} & RAL 2024 & \checkmark & \checkmark &  &   &  &  & HexPlane + MLP  & Density   &   &   & & \checkmark & DROID \cite{droidslam} &   &   &   &  \checkmark & DeepLabV3 \cite{chen2017rethinking} & \href{https://github.com/HannahHaensen/DynaMoN}{Code} \\

   \hyperlink{NID-SLAM}{\textcolor{magenta}{NID-SLAM}}~\cite{xu2024nid} & ICME 2024 & \checkmark &  &   &   &   &   & Hier. Grid + MLP & Occupancy  &  &  & \checkmark &  & &  &  &  &  \checkmark &  & \\

    \hyperlink{TivNe-SLAM}{\textcolor{magenta}{TivNe-SLAM}}~\cite{duantivne} & IROS 2024 & \checkmark & & & & & & Octree Grid + MLP & SDF & \checkmark & & \checkmark & & & & & & \checkmark &  Mask R-CNN \cite{He_2017_ICCV} & \\
    \hyperlink{RoDyn-SLAM}{\textcolor{magenta}{RoDyn-SLAM}}~\cite{jiang2024rodynslam} & RAL 2024 & \checkmark & & & & & & Hash Grid + MLP & TSDF & \checkmark & & \checkmark & & & \checkmark & & & \checkmark & Oneformer\&RAFT-GMA \cite{jiang2021learning} & \href{https://github.com/fudan-zvg/Rodyn-SLAM}{Code} \\
    \hyperlink{DG-SLAM}{\textcolor{magenta}{DG-SLAM}}~\cite{xu2024dgslam} & NeurIPS 2024 & \checkmark & & & && & 3D Gaussians & Density & \checkmark & & & \checkmark & DROID \cite{droidslam} & & & & \checkmark &  Oneformer \cite{jain2023oneformer} & \href{https://github.com/fudan-zvg/DG-SLAM}{Code} \\
    \hyperlink{ONeK-SLAM}{\textcolor{magenta}{ONeK-SLAM}}~\cite{onekslam} & ICRA 2024 & \checkmark & & & & & & Hier. Grid + MLP &Occupancy & \checkmark & & & \checkmark & SIFT matching \cite{lowe2004distinctive} & & & & \checkmark & SAM-Track \cite{cheng2023segment} & \\
   
    \hline
    
   \hyperlink{OpenWorld-SLAM}{\textcolor{magenta}{OpenWorld-SLAM}}~\cite{lisus2023towards}  & CRV 2023 & \checkmark &  &  &  & \checkmark &  & Hier. Grid + MLP & Occupancy   &   &  \checkmark & \checkmark &  &  & & &  &  &  & \\
   
   \hyperlink{UncLe-SLAM}{\textcolor{magenta}{UncLe-SLAM}}~\cite{uncleslam2023} & ICCVW 2023 & \checkmark &  & \checkmark &  &  &  & Hier. Grid + MLP & Occupancy   &   &  \checkmark &  \checkmark &  &   &   &  & &   &   & \href{https://github.com/kev-in-ta/UncLe-SLAM}{Code} \\

   \hyperlink{NVINS}{\textcolor{magenta}{NVINS}}~\cite{han2024nvins} & IROS 2024 & \checkmark & & & & \checkmark & & Hash Grid + MLP & Density & & \checkmark & & \checkmark & Custom PoseNet & & & & & & \\

   \hyperlink{CDA-SLAM}{\textcolor{magenta}{CDA}}~\cite{cdaslam} & T-ITS 2024 & \checkmark & & & & & & Hash Grid + MLP & Density & & \checkmark & & \checkmark & ORB3 \cite{campos2021orb} &   & & & & &  \\
    \hline
    
   \hyperlink{EN-SLAM}{\textcolor{magenta}{EN-SLAM}}~\cite{qu2023implicit} & CVPR 2024 & \checkmark &   &  &   &   &  \checkmark & Hier. Grid + MLP & TSDF  &  &  & \checkmark &  & & \checkmark &  &  &  &  & \\

   \Xhline{1pt}
    
   \rowcolor{maincategories}\multicolumn{21}{c}{\hyperlink{RGB}{\textcolor{magenta}{RGB (Sec. \ref{sec:rgb})}}} \\
    
   \Xhline{1pt}
    
   \hyperlink{DIM-SLAM}{\textcolor{magenta}{DIM-SLAM}}~\cite{li2023dense} & ICLR 2023 &  & \checkmark &  &  &  &  & Hier. Grid + MLP & Density   &  &   & \checkmark &   & &  &  &  &   &  & \href{https://poptree.github.io/DIM-SLAM/}{Code} \\ 
   
   \hyperlink{Orbeez-SLAM}{\textcolor{magenta}{Orbeez-SLAM}}~\cite{chung2023orbeez} & ICRA 2023 &  & \checkmark &  &  &  &  & Hash Grid + MLP & Density   &  &  & & \checkmark & ORB2 \cite{mur2017orb} &  &  &  &  &  & \href{https://github.com/MarvinChung/Orbeez-SLAM}{Code} \\
   
   \hyperlink{TT-HO-SLAM}{\textcolor{magenta}{TT-HO-SLAM}}~\cite{lin2023ternary} & IROS 2024 &   & \checkmark &  &  &   &   & Hier. Grid + MLP  & Density & & & \checkmark &  & & & & & &  &  \\
   
   \hline

   \hyperlink{MonoGS++}{\textcolor{magenta}{MonoGS++}}~\cite{monogspp} & BMVC 2024 &   & \checkmark &  &  &   &   & 3D Gaussians  & Density & & & & \checkmark  & DPVO \cite{teed2022deep} & & & & &  &  \\
   
   \hline
   
   \hyperlink{iMode}{\textcolor{magenta}{iMode}}~\cite{matsuki2023imode} & ICRA 2023 &   & \checkmark &   &  &   &   & MLP & Density   &   &  & & \checkmark & ORB \cite{mur2015orb} &   &  &   & &  Sparse-to-dense \cite{Ma2017SparseToDense} &  \\
   
   \hyperlink{Hi-SLAM}{\textcolor{magenta}{Hi-SLAM}}~\cite{zhang2023hi} & RAL 2023 &  & \checkmark &  &  &  &  & Hash Grid + MLP & TSDF  &  &  & & \checkmark & DROID \cite{droidslam} & \checkmark & \checkmark &  &  &  Omnidata \cite{eftekhar2021omnidata} & \\
   
   \hyperlink{NICER-SLAM}{\textcolor{magenta}{NICER-SLAM}}~\cite{Zhu2023NICER} & 3DV 2024 &  & \checkmark &   &  &  &  & Hier. Grid + MLP & SDF  &  &  & \checkmark &  & & &  &  &  &  GMFlow \cite{xu2022gmflow} \& Omnidata~\cite{eftekhar2021omnidata} & \href{https://nicer-slam.github.io/}{WebPage} \\
   
   \hyperlink{NeRF-VO}{\textcolor{magenta}{NeRF-VO}}~\cite{naumann2023nerf} & RAL 2024 &   & \checkmark &  &   &  &  & Hash Grid + MLP & Density   &  & \checkmark  &  & \checkmark & DPVO \cite{teed2022deep} &  &  &   &   &  Omnidata \cite{eftekhar2021omnidata} &  \\
   
   \hyperlink{MoD-SLAM}{\textcolor{magenta}{MoD-SLAM}}~\cite{zhou2024modslam} & RAL 2024 & \checkmark & \checkmark &  &  &  &  & Hash Grid + MLP & SDF  &   &   & & \checkmark & DROID \cite{droidslam} &   & \checkmark &   &   &  DPT \cite{Ranftl2021} \& ZoeDepth \cite{bhat2023zoedepth} & \\
   
   \hyperlink{Q-SLAM}{\textcolor{magenta}{Q-SLAM}}~\cite{peng2024q} & CoRL 2024 & \checkmark & \checkmark &  &  &  &  & Grid Fact. + MLP & Density  &   &   & & \checkmark & DROID \cite{droidslam} &   & &   &   &  Segmentation Network & \\

   \hyperlink{MGS-SLAM}{\textcolor{magenta}{MGS-SLAM}}~\cite{zhu2024mgs} & RAL 2024 & & \checkmark & & & & & 3D Gaussians & Density & & & & \checkmark & DPVO \cite{teed2022deep} & & & & & & \href{https://github.com/Z-Pengcheng/MGS-SLAM}{Code}$\dagger$ \\

   \hline 
   
   \hyperlink{RO-MAP}{\textcolor{magenta}{RO-MAP}}~\cite{RO-MAP} & RAL 2023 &  & \checkmark &   &  &  &  & Hash Grid + MLP & Occupancy &  \checkmark &  & & \checkmark & ORB2 \cite{mur2017orb} & & & & &  YOLOv8 & \href{https://github.com/XiaoHan-Git/RO-MAP}{Code} \\

    \hyperlink{3DIML}{\textcolor{magenta}{3DIML}}~\cite{3diml} & ICRA 2024 & & \checkmark & & & & & Hash Grid + MLP & Density & \checkmark & & & \checkmark & NetVLAD \& LoFTR \cite{sun2021loftr} & & & & & Mask2Former \cite{cheng2021maskformer} &   \\
   
   \hline
   
   \hyperlink{NeRF-SLAM}{\textcolor{magenta}{NeRF-SLAM}}~\cite{rosinol2023nerf} & IROS 2023 &  & \checkmark &  &  &  &  & Hash Grid + MLP  & Density  &  & \checkmark & & \checkmark & DROID \cite{droidslam} &   &   &   &   &  & \href{https://github.com/ToniRV/NeRF-SLAM}{Code} \\
   
   \Xhline{1pt}
   
   \rowcolor{maincategories}\multicolumn{21}{c}{\hyperlink{LiDAR}{\textcolor{magenta}{LiDAR (Sec. \ref{sec:lidar})}}} \\
   
   \Xhline{1pt}
    
   \hyperlink{NeRF-LOAM}{\textcolor{magenta}{NeRF-LOAM}}~\cite{deng2023nerf} & ICCV 2023 &   &   & \checkmark &   &  &   & Octree Grid + MLP & SDF & &  &  \checkmark  &  &  &  &   &   &   &   & \href{https://github.com/JunyuanDeng/NeRF-LOAM}{Code} \\
   
   \hyperlink{LONER}{\textcolor{magenta}{LONER}}~\cite{isaacson2023loner} & RAL 2023 &  &   & \checkmark &  &  &  & Hier. Grid + MLP & Density &  &  & & \checkmark & P2P-ICP \cite{rusinkiewicz2001efficient} &   &  &  &  &  & \href{https://umautobots.github.io/loner}{Code} \\
   
   \hyperlink{PIN-SLAM}{\textcolor{magenta}{PIN-SLAM}}~\cite{pan2024pin} & T-RO 2024 & \checkmark &  & \checkmark &   &  &  & Neural Points + MLP & SDF & \checkmark &  & \checkmark & &   &  & \checkmark &    & \checkmark &   & \href{https://github.com/PRBonn/PIN_SLAM}{Code} \\

   \hyperlink{TNDF-Fusion}{\textcolor{magenta}{TNDF-Fusion}}~\cite{TNDF} & RAL 2024 &  &  & \checkmark &   &  &  & Features Planes + MLP & TNDF & &  & \checkmark & &   &  &  &    & &  & \\
   
   \hline
   
   \hyperlink{LIV-GaussMap}{\textcolor{magenta}{LIV-GaussMap}}~\cite{hong2024liv} & RAL 2024 & \checkmark &  &  &  & \checkmark & &  3D Gaussians & Density & & & \checkmark &  & & & & & & & \href{https://github.com/sheng00125/LIV-GaussMap}{Code}$\dagger$ \\
   \hyperlink{MM-Gaussian}{\textcolor{magenta}{MM-Gaussian}}~\cite{wu2024mm} & IROS 2024 & \checkmark &  &  &  &  & &  3D Gaussians & Density & & &  & \checkmark &  Kiss-ICP \cite{vizzo2023kiss} & & & & & SuperPoint~\cite{detone2018superpoint} \& LightGlue~\cite{lindenberger2306lightglue} & \\
   \Xhline{2pt}
   \end{tabular} \\
   \end{tabular}
}
\label{tab:overview}
\end{table*}

%% file: chapters/taxonomy/rgbd/intro.tex
\section{Simultaneous Localization and Mapping}
\label{sec:methods}

This section introduces SLAM systems that leverage recent progress in radiance field representations. The papers are categorized by their sensor type: RGB-D (\ref{sec:RGB-D}), RGB (\ref{sec:rgb}), and LiDAR (\ref{sec:lidar}), with methods organized chronologically by publication date in major conferences and journals. While the field is rapidly evolving with numerous preprints appearing regularly on arXiv, we focus on peer-reviewed publications to provide a curated overview of established approaches. We encourage readers to explore recent arXiv submissions for the latest developments in this active research area.

For a more advanced understanding, Table \ref{tab:overview} offers a detailed overview of the surveyed methods. This table provides an in-depth summary, highlighting key features of each method, and includes references to project pages or source code whenever available. For further details or method specifics, please refer to the original papers.

\subsection{RGB-D SLAM Approaches}
\label{sec:RGB-D}

\hypertarget{RGB-D}{Here we focus on dense SLAM techniques using RGB-D cameras that capture both color images and per-pixel depth information of the environment. These techniques fall into distinct categories: NeRF-style SLAM solutions (\ref{sec:vanilla-RGB-D}) and alternatives based on the 3D Gaussian Splatting representation (\ref{sec:3DGS}). Specialized solutions derived from both approaches include submap-based SLAM methods for large scenes (\ref{sec:sub-maps}), frameworks that address semantics (\ref{sec:segmentation-priors}), and those tailored for dynamic scenarios (\ref{sec:dynamic-envs}). Within this classification, some techniques assess reliability through uncertainty (\ref{sec:uncertainty}), while others explore the integration of additional sensors like event-based cameras (\ref{sec:event}).}

%% file: chapters/taxonomy/rgbd/vanilla.tex
\subsubsection{NeRF-style RGB-D SLAM}
\label{sec:vanilla-RGB-D}

Advances in implicit neural representations have enabled accurate and dense 3D surface reconstruction. This has led to novel SLAM systems derived from or inspired by NeRF, initially designed for offline use with known camera poses. 
\hypertarget{iMAP}{\textbf{iMAP \cite{sucar2021imap}}, illustrated in Fig.\ref{fig:imap_overview}, marks the first attempt to leverage implicit neural representations for SLAM. This groundbreaking achievement not only pushes the boundaries of SLAM but also establishes a new direction for the field. 
Specifically, this framework  uses an MLP to map 3D coordinates into color and volume density, allowing for rendering depth and color images through network queries. Joint optimization of photometric and geometric losses for a fixed set of keyframes refines network parameters and camera poses. A parallel process ensures close-to-frame-rate camera tracking, with dynamic keyframe selection based on information gain.} 
In contrast to the use of a single MLP by iMAP, \hypertarget{NICE-SLAM}{\textbf{\withcode{NICE-SLAM} \cite{zhu2022nice}}} optimizes a hierarchical representation using three pre-trained MLPs, by encoding geometry into three voxel grids of varying resolutions, each associated with its corresponding pre-trained MLP — coarse, mid, and fine levels. Moreover, a dedicated feature grid and decoder are utilized for capturing scene appearance. 
\hypertarget{Vox-Fusion}{\textbf{\withcode{Vox-Fusion}
\cite{yang2022vox}}}, instead, combines traditional volumetric fusion methods with neural implicit representations, by leveraging a voxel-based neural implicit surface representation and using an octree-based structure to enable dynamic voxel allocation. Local scene geometry is modeled within individual voxels using a continuous SDF, encoded by an MLP along with shared feature embeddings. Furthermore, Vox-Fusion introduces an efficient keyframe selection strategy tailored for sparse voxels, further enhancing its capability for efficient map management. 
\hypertarget{ESLAM}{\textbf{\withcode{ESLAM} \cite{ESLAM}}}, on the other hand, uses multi-scale axis-aligned feature planes, diverging from traditional voxel grids. This approach optimizes memory usage through quadratic scaling, in contrast to the cubic growth exhibited by voxel-based models. Furthermore, ESLAM adopts TSDF as the geometric representation, increasing both convergence speed and reconstruction quality. 
\hypertarget{Co-SLAM}{\textbf{\withcode{Co-SLAM} \cite{Wang_2023_CVPR}} }combines the smoothness of coordinate encodings (using one-blob encoding \cite{muller2019neural}) with the fast convergence and local detail advantages of sparse parametric encodings -- such as hash grids \cite{mueller2022instant}. 
In addition, Co-SLAM introduces global bundle adjustment by sampling few rays from all previous keyframes (around 5\%  of pixels for each keyframe). 
Taking optimization a step further, \hypertarget{GO-SLAM}{\textbf{\withcode{GO-SLAM} \cite{zhang2023go}} is designed for real-time global optimization of camera poses and 3D reconstructions, integrating efficient LC and online full BA that utilizes the full history of input frames. 
Specifically, the architecture operates through three parallel threads: \textit{front-end tracking}, responsible for iterative pose and depth updates along with efficient loop closing; \textit{back-end tracking}, focused on generating globally consistent pose and depth predictions via full BA; and \textit{instant mapping}, which updates the 3D reconstruction based on the latest available poses and depths. 
Notably, GO-SLAM supports monocular, stereo, and RGB-D cameras.}
Unlike grid-based or network-based methods, \hypertarget{Point-SLAM}{\textbf{\withcode{Point-SLAM} \cite{Sandström2023ICCV}}} 
introduces a dynamic neural point cloud representation, adjusting point density based on input data information, ensuring more points in areas with higher detail and fewer points in less informative regions, based on image gradients.
Color and depth are rendered by processing the neural points features through pre-trained MLPs, as in \cite{zhu2022nice}.
The neural point cloud expands incrementally during exploration and stabilizes as all relevant regions are incorporated, thus optimizing memory usage by compressing areas with fewer details and removing the need to model free space. 
In a different direction, \hypertarget{ToF-SLAM}{\textbf{\withcode{ToF-SLAM} \cite{tofslam}}
is the first framework tailored to lightweight ToF sensors providing very few depth measurements. It uses a multi-modal feature grid enabling both zone-level rendering tailored for ToF sparse measurements and pixel-level suited for high-resolution signals (\eg RGB). 
A coarse-to-fine optimization strategy improves the learning of the implicit representation, while temporal information is incorporated to handle noisy ToF sensor signals.}
Addressing the challenge of incomplete data, \hypertarget{ADFP}{\textbf{\withcode{ADFP} \cite{Hu2023LNI-ADFP}} incorporates an attentive depth fusion prior, derived from a TSDF obtained by fusing multiple depth images. This allows neural networks to directly utilize learned geometry and TSDFs during volume rendering to better deal with holes and unawareness of occluded structures. Overall, ADFP uses classical ray tracing, feature interpolation, occupancy prediction priors, and an attention mechanism to balance the contributions of learned geometry and the depth fusion prior.}
\hypertarget{MLM-SLAM}{\textbf{MLM-SLAM \cite{li2023end}} introduces a multi-MLP hierarchical scene representation that utilizes different levels of decoders to extract detailed features. 
Additionally, it implements a refined tracking strategy and keyframe selection approach, enhancing system reliability, especially in challenging dynamic environments.}

\hypertarget{Plenoxel-SLAM}{\textbf{\withcode{Plenoxel-SLAM} \cite{teigen2024rgb}}, instead, builds upon the Plenoxel radiance field model \cite{fridovich2022plenoxels}, devoid of neural networks, by using a voxel grid representation and trilinear interpolation for efficient dense mapping and tracking. 
It is worth mentioning that no explicit 3D mesh is currently reconstructed from the learned representation.}
\hypertarget{Struct-SLAM}{\textbf{\withcode{Structerf-SLAM}\cite{wang2024structerf}} uses two-layer feature grids and pre-trained decoders to decode interpolated features into RGB and depth values. During the tracking phase, three-dimensional planar features based on the Manhattan assumption improve stability and overcome the lack of texture.
In the mapping stage, a planar consistency constraint is applied to rendered depth, resulting in smoother reconstructions.}
\hypertarget{KN-SLAM}{\textbf{\withcode{KN-SLAM}\cite{wu2024kn}} integrates sparse feature-based localization using HF-Net \cite{sarlin2019coarse} with a hierarchical neural implicit representations. It consists of three concurrent threads: 1) tracking, which extracts local and global features for initial pose estimation; 2) mapping, which updates the scene representation and jointly optimizes camera poses and implicit mapping; and 3) optimization, which performs loop detection, pose graph optimization, and global refinement.} 
\hypertarget{SLAIM}{\textbf{SLAIM \cite{cartillier2024slaim}}
implements a Gaussian pyramid filter on top of NeRF to perform coarse-to-fine tracking and mapping, and introduces a KL regularizer on the ray termination distribution to distinguish empty space from opaque surfaces in the scene geometry. Finally, it implements both local and global bundle adjustment.} 
Advancing generalization, \hypertarget{IBD-SLAM}{\textbf{IBD-SLAM \cite{IBDSLAM}} is the first generalizable SLAM solution based on neural implicit representations, which learns an image-based depth fusion model that fuses depth maps of multiple reference views into an xyz-map representation. This model is pretrained and then applied to new RGBD videos of unseen scenes without any retraining, only requiring a light-weight pose optimization procedure. } 
To enhance scene representation, \hypertarget{VPE-SLAM}{\textbf{VPE-SLAM \cite{vpeslam}} proposes voxel-permutohedral encoding, which can incrementally reconstruct maps of unknown scenes. It combines a sparse voxel feature grid created by an octree and multi-resolution permutohedral tetrahedral feature grids to represent the scene effectively. 
Furthermore, it proposes a novel local bundle adjustment module managing adjacent keyframes over a sliding window.} 
\hypertarget{HERO-SLAM}{\textbf{HERO-SLAM \cite{heroslam}} combines the benefits of neural implicit field and feature-metric optimization, offering increased robustness in challenging environments, such as in the presence of sudden viewpoint changes or sparsely collected frames. This, together with a hybrid optimization pipeline tailored to optimize the multi-resolution implicit fields, allows for improved performance in very challenging scenarios.}
Tackling computational efficiency, \hypertarget{LRSLAM}{\textbf{LRSLAM \cite{park2024lrslam}} uses low-rank tensor decomposition methods to reduce the complexity and memory required for storing the features processed by the MLPs, by leveraging Six-axis and CP decompositions.}
Finally, \hypertarget{LCP-Fusion}{\textbf{LCP-Fusion \cite{wang2024lcp}} uses a sparse voxel octree structure containing feature grids and SDF priors as a hybrid scene representation. It also proposes a sliding window selection strategy based on visual overlap to perform loop-closure, a warping loss to constrain relative poses, and SDF priors as coarse initialization for implicit features.}

%% file: chapters/taxonomy/rgbd/3DGS.tex
\subsubsection{3DGS-style RGB-D SLAM}
\label{sec:3DGS}

These approaches typically exploit the advantages of 3D Gaussian Splatting, such as faster and more photorealistic rendering compared to other existing scene representations. They also offer the flexibility to increase map capacity by adding more Gaussian primitives, complete utilization of per-pixel dense photometric losses, and direct parameter gradient flow to facilitate fast optimization. 

\hypertarget{MonoGS}{\textbf{MonoGS\cite{matsuki2023gaussian}} is the first introducing a paradigm shift in the field, by leveraging 3D Gaussians as the representation coupled with splatting rendering techniques using a single moving RGB or RGB-D camera. The framework includes several key components, such as tracking and camera pose optimization, Gaussian shape verification and regularization, mapping and keyframing, and resource allocation and pruning. The tracking phase adopts a direct optimization scheme against the 3D Gaussians. 
For mapping and keyframing, MonoGS integrates techniques for efficient online optimization and keyframe management, which involves selecting and maintaining a small window of keyframes based on inter-frame covisibility. Additionally, resource allocation and pruning methods are used to eliminate unstable Gaussians and avoid artifacts in the model.}

\hypertarget{Photo-SLAM}{Concurrently, \textbf{Photo-SLAM \cite{huang2023photo}} integrates explicit geometric features and implicit texture representations within a hyper primitives map, combining ORB features \cite{rublee2011orb}, rotation, scaling, density, and spherical harmonic coefficients. The framework leverages geometry-based densification and Gaussian-Pyramid-based learning to optimize camera poses while minimizing a photometric loss.} 

On the same track, \hypertarget{SplaTAM}{\textbf{\withcode{SplaTAM} \cite{keetha2023splatam}} represents the scene as a collection of simplified 3D Gaussians, enabling high-quality color and depth image rendering, and includes the following key steps: Camera Tracking, minimizing re-rendering errors for precise camera pose estimation; Gaussian Densification, adding new Gaussians to the scene; Map Updating, which refines gaussian parameters across the scene.}

\hypertarget{GS-SLAM}{\textbf{GS-SLAM \cite{yan2023gs}}, using 3D Gaussians together with opacity and spherical harmonics to encapsulate both scene geometry and appearance, introduces an adaptive expansion strategy that dynamically manages the addition or removal of 3D Gaussians, and a coarse-to-fine tracking technique that iteratively refines estimated camera poses.} 
Focusing more on tracking accuracy, \hypertarget{GS-ICP SLAM}{\textbf{GS-ICP SLAM \cite{ha2024rgbd}} combines two techniques - Generalized Iterative Closest Point (G-ICP) \cite{segal2009generalized} and 3DGS, 
to directly use the 3D Gaussian map representation for tracking, without the need for additional post-processing. 
Covariances matrices are computed during the G-ICP tracking and used to initialize the 3DGS mapping, while the 3D Gaussians in the map are in turn used as 3D points and their covariances for the G-ICP tracking. This bidirectional exchange of information, facilitated by scale alignment techniques, minimizes redundant computations and enables an efficient and high-performance SLAM system.}

To better address mapping challenges, \hypertarget{HF-GS SLAM}{\textbf{HF-GS SLAM \cite{sun2024high}} proposes a Gaussian densification strategy guided by the rendering loss to map unobserved areas and refine reobserved regions, and incorporates regularization parameters during mapping to mitigate the forgetting problem. 
A regularization term is proposed to prevent overfitting and preserve details of previously visited areas.}
\hypertarget{CG-SLAM}{\textbf{CG-SLAM \cite{hu2024cg}} uses a novel uncertainty-aware 3D Gaussian field, by analyzing camera pose derivatives in the EWA (Elliptical Weighted Average) splatting process. 
To reduce overfitting and achieve a consistent Gaussian field, CG-SLAM employs techniques such as scale regularization, depth alignment, and a depth uncertainty model to guide the selection of informative Gaussian primitives.}
Expanding to multi-modal inputs, \hypertarget{MM3DGS-SLAM}{\textbf{MM3DGS-SLAM \cite{sun2024mm3dgs}} takes inputs from a monocular camera or RGB-D camera, along with inertial measurements. The camera pose is optimized using a combined tracking loss incorporating depth measurements and IMU pre-integration. Keyframe selection is based on image covisibility and the Naturalness Image Quality Evaluator (NIQE) metric across a sliding window. New 3D Gaussians are initialized for keyframes with low opacity and high depth error, with positions initialized using depth measurements or estimates.}
Scale ambiguity is tackled by solving for scaling and shift values fitting the depth estimate to the current map.
\begin{figure}[]
    \centering
    \renewcommand{\tabcolsep}{1pt}
    \begin{tabular}{cc}
        \includegraphics[width=0.24\textwidth]{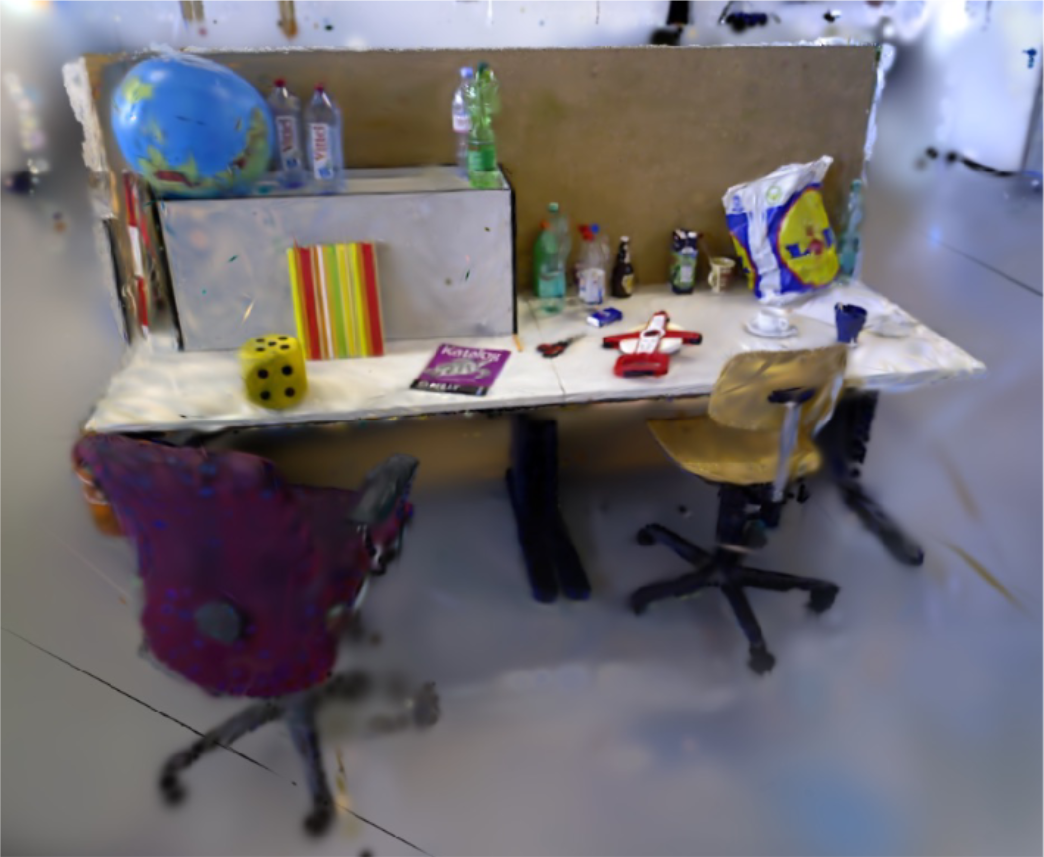} & 
        \includegraphics[width=0.24\textwidth]{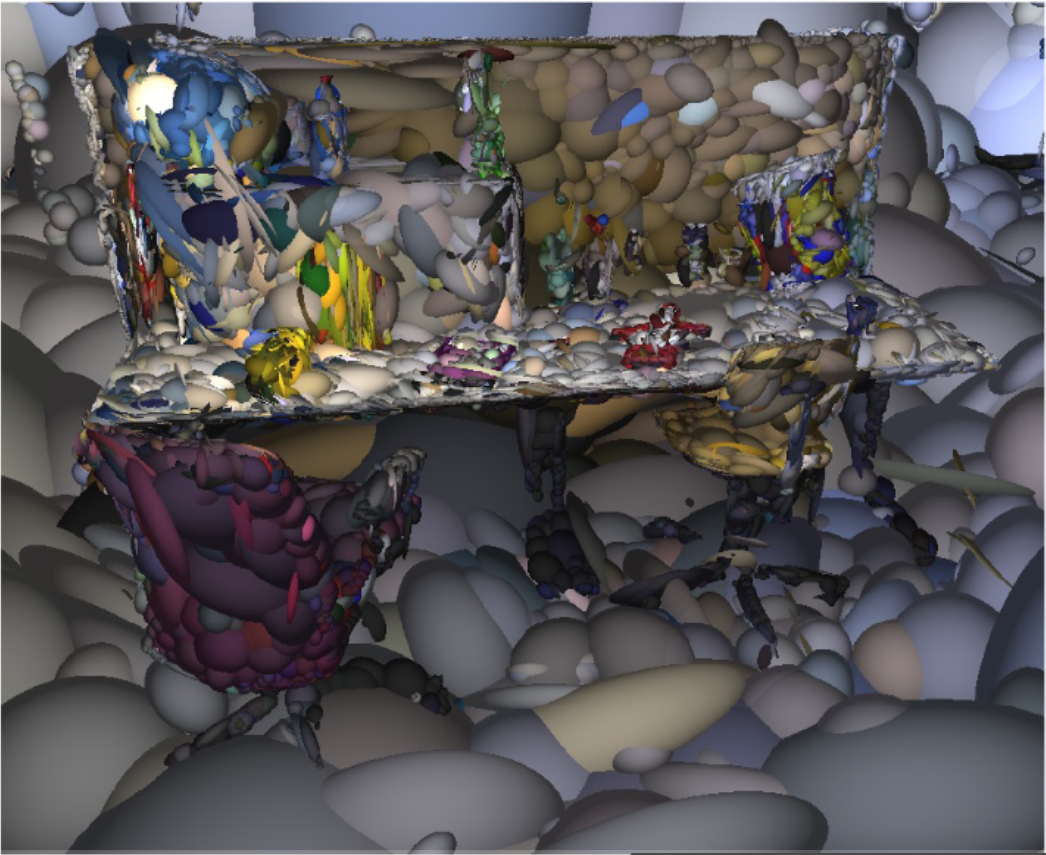} \\
    \end{tabular}
    \caption{\textbf{3D Gaussian Visualization.} (Left) Rasterized Gaussians, (Right) Gaussians shaded to highlight the underlying geometry. Images adapted from \cite{matsuki2023gaussian}.}\vspace{-0.3cm}
    \label{fig:3D_gaussian_visualization}
\end{figure}

Finally, \hypertarget{RTG-SLAM}{\textbf{RTG-SLAM \cite{peng2024rtg}} is a real-time system for large-scale environments, forcing each Gaussian to be either opaque or nearly transparent, with the opaque ones fitting the surface and dominant colors, and transparent ones fitting residual colors. 
Depth and color are rendered in different ways, letting single Gaussians fit a local surface alone. New gaussians are added on the fly for newly observed pixels and pixels with large color or depth errors, and are classified as stable or unstable -- with the latter only being optimized to reduce complexity.}

%% file: chapters/taxonomy/rgbd/submaps.tex
\subsubsection{Submaps-based SLAM}
\label{sec:sub-maps}

In this category, we examine methods addressing catastrophic forgetting and scalability limitations of dense radiance field SLAM systems in large environments. Among them, \hypertarget{MeSLAM}{\textbf{MeSLAM \cite{kruzhkov2022meslam}} introduces a novel SLAM framework for large-scale environment mapping with a minimal memory footprint. 
By using distributed MLP networks, a global mapping module facilitates the segmentation of the environment into distinct regions, mapped by single MLPs, and coordinates the stitching of these regions during the reconstruction process. 
A key aspect relies upon the integration with external odometry \cite{park2017colored}, allowing for robust tracking in regions where maps intersect. Both neural network parameters and poses are optimized simultaneously.}

\begin{figure}[t]
  \centering
  \includegraphics[width=0.45\textwidth]{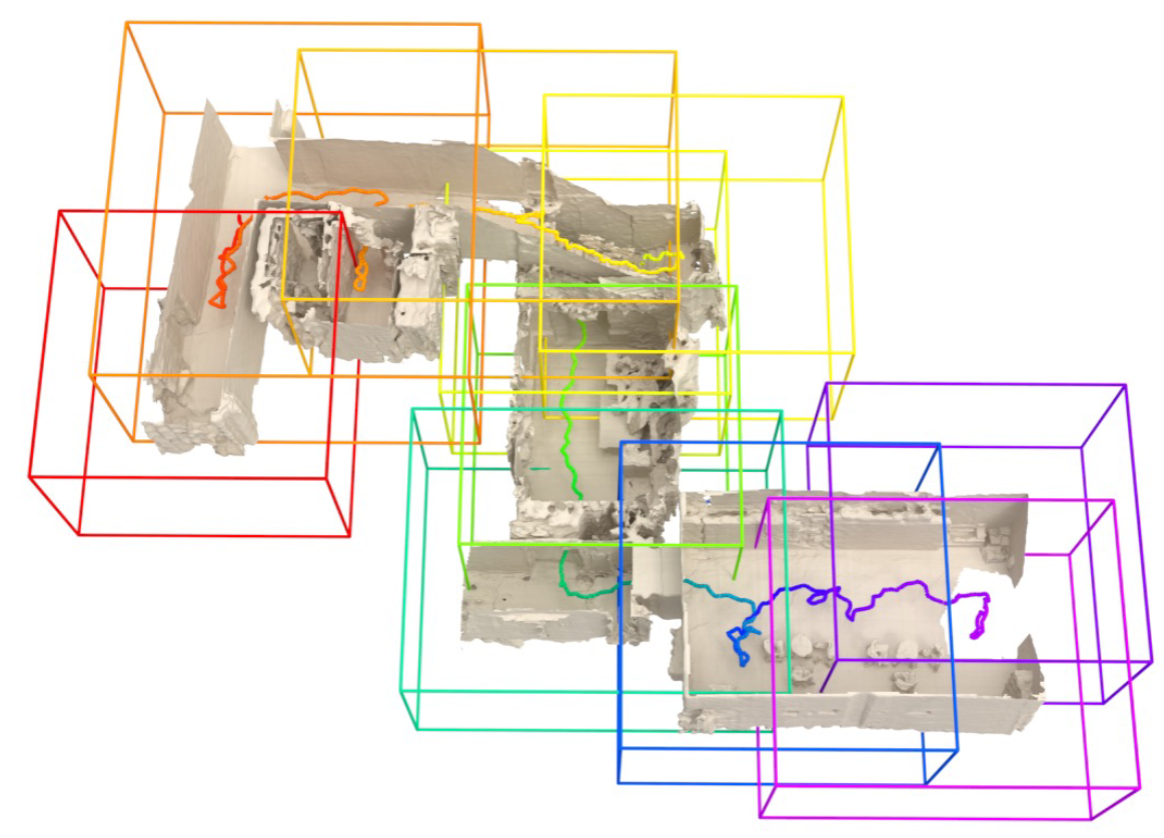}\\
  \vspace{-0.3em}
  \includegraphics[width=0.4\textwidth]{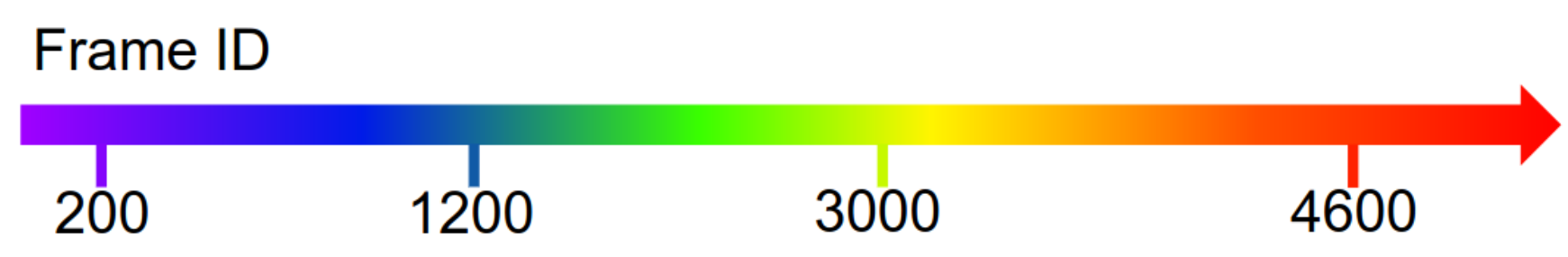}\\
  \caption{\textbf{Submaps Visualization.} Neural submaps, allocated incrementally along the scanning trajectory, encode precise scene geometry and colors in their dedicated local coordinate frames. Figure from \cite{tang2023mips}.}
  \label{fig:submap_overview}
\end{figure}

\hypertarget{CP-SLAM}{\textbf{CP-SLAM \cite{hu2023cp}}, instead, leverages a neural point-based 3D scene representation associated with keyframes and employs a distributed-to-centralized approach to ensure consistency and cooperation among multiple agents, where front-end modules use neural point clouds and differentiable volume rendering to achieve efficient odometry, mapping, and tracking. CP-SLAM also implements loop detection and sub-map alignment techniques to mitigate pose drift and concludes with global optimization techniques such as pose graph optimization and map refinement.}
Focusing on spatial organization, \hypertarget{NISB-Map}{\textbf{NISB-Map \cite{xiang2023nisb}} uses multiple small MLP networks to represent the large-scale environment in compact spatial blocks. 
A distillation procedure for overlapping Neural Implicit Spatial Block (NISBs) is implemented, minimizing density variations and ensuring geometric consistency. In this process, knowledge from the last trained NISB serves as the teacher and is distilled only within overlapping regions with the current NISB. This ensures continuity while reducing computation and training time compared to training a global NISB.}
Similarly, \hypertarget{Multiple-SLAM}{\textbf{Multiple-SLAM \cite{liu2023efficient}} employs multiple SLAM agents to process scenes in blocks. Agents are deployed in the frontend to operate independently, while also facilitating the sharing and fusion of map information through the backend server. 
The pose estimation process efficiently determines relative poses between agents using a two-stage approach: matching keyframes through a NetVLAD-based \cite{arandjelovic2016netvlad} global descriptor extraction model and fine-tuning inter-agent poses through an implicit relocalization process. Conversely, the map fusion stage integrates local maps using a floating-point sparse voxel octree. Overlapping regions are handled by removing redundant voxels based on observation confidence and a reconstruction loss.} 
\hypertarget{MIPS-Fusion}{\textbf{MIPS-Fusion \cite{tang2023mips}} uses a grid-free, purely neural approach with incremental allocation and on-the-fly learning of multiple neural submaps, as depicted in Figure \ref{fig:submap_overview}. It also incorporates efficient on-the-fly learning through local bundle adjustment, distributed refinement with back-end optimization, global optimization through loop closure, and a hybrid tracking scheme, combining gradient-based and randomized optimizations via particle filtering to ensure robust performance, particularly under fast camera motions. Key features include a depth-to-TSDF loss for efficient fitness evaluation, a lightweight network for classification-based TSDF prediction, and support for parallel submap fine-tuning. Loop closure is implemented through covisibility thresholds, not allowing for correcting large drifts.}
\hypertarget{NGEL-SLAM}{\textbf{NGEL-SLAM \cite{mao2023ngel}} deploys two tracking and mapping submodules to integrate the robust tracking capabilities of ORB-SLAM3 \cite{campos2021orb} with the scene representation provided by multiple implicit neural maps. Operating through three concurrent processes—tracking, dynamic local mapping, and loop closing—the system ensures global consistency and low latency.  
Loop closing optimizes poses using global BA, and the use of multiple local maps minimizes re-training time. NGEL-SLAM also incorporates uncertainty-based image rendering for optimal sub-map selection, and its scene representation is based on a sparse octree-based grid with implicit neural maps.} 
Addressing memory efficiency, \hypertarget{PLGSLAM}{\textbf{PLGSLAM \cite{deng2023plgslam}} utilizes axis-aligned triplanes for high-frequency features and an MLP for global low-frequency features. This reduces memory growth from cubic to square with respect to the scene size, enhancing scene representation efficiency. PLGSLAM also integrates traditional SLAM with an end-to-end pose estimation network, introducing a local-to-global BA algorithm to mitigate cumulative errors in large-scale indoor scenes. The efficient management of keyframe databases enables seamless BA across all past observations.}
\hypertarget{Loopy-SLAM}{\textbf{Loopy-SLAM \cite{liso2024loopy}} leverages neural point clouds in the form of submaps for local mapping and tracking. It employs frame-to-model tracking with a data-driven, point-based submap generation approach, dynamically growing submaps based on camera motion during scene exploration. Global place recognition triggers loop closures online, enabling robust pose graph optimization for global alignment of submaps and trajectory, with the point-based representation facilitating efficient map corrections.} 
\hypertarget{NEWTON}{\textbf{NEWTON \cite{matsuki2024newton}} introduces a view-centric neural field-based mapping method designed to overcome the  limitations of a single world-centric map, such as the inability to capture dynamic content, by constructing multiple neural field models based on real-time observations and allowing camera pose updates through loop closures and scene boundary adjustments.
Each neural field is represented as a multi-resolution feature grid \cite{mueller2022instant} in a spherical coordinate system. This is facilitated by the coordination with the camera tracking component of ORB-SLAM2 \cite{mur2017orb}.}
\hypertarget{MAN-SLAM}{\textbf{MAN-SLAM \cite{dengmulti}} is a multi-agent collaborative SLAM
framework with joint scene representation, distributed camera
tracking, intra-to-inter loop closure, and sub-map fusion. Specifically, the intra-to-inter loop closure method is designed to achieve local (single-agent) while ensuring global map consistency. MAN-SLAM is flexible and supports single-agent and multi-agent modes. 
Furthermore, the authors introduce a real-world dataset suited for both settings, providing continuous-time trajectories and high-accuracy 3D meshes as ground truth.}

%% file: chapters/taxonomy/rgbd/segmentation.tex
\subsubsection{Semantic RGB-D SLAM}
\label{sec:segmentation-priors}

Operating as SLAM systems, these methodologies inherently include mapping and tracking processes while also incorporating semantic information to enhance the understanding of the environment. Tailored for tasks such as object recognition or semantic segmentation, these frameworks provide a holistic approach to scene analysis - identifying and classifying objects and/or efficiently categorizing image regions into specific semantic classes (\eg tables, chairs, etc.). Early developments focused on interactive understanding: \hypertarget{iLabel}{\textbf{iLabel \cite{zhi2022ilabel}} introduced a framework mapping 3D coordinates to color, density, and semantic values, built upon iMAP. The system supports both manual user-click annotations and automatic label proposals based on semantic uncertainty, achieving efficient interactive labeling without pre-existing training data.} \hypertarget{FR-Fusion}{\textbf{FR-Fusion \cite{mazur2023feature}}, on the other hand, integrated neural feature fusion into iMAP by incorporating 2D feature extractors (EfficientNet or DINO-based) with latent volumetric rendering, enabling efficient feature map fusion for dynamic open-set segmentation while maintaining low computational requirements.}

\begin{figure}[t]
  \centering
  \begin{tabular}{cc}
    \includegraphics[width=0.48\linewidth]{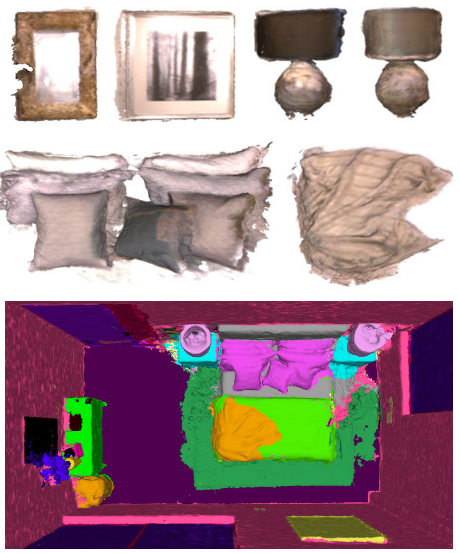} \hspace{-0.5em} &
    \includegraphics[width=0.43\linewidth]{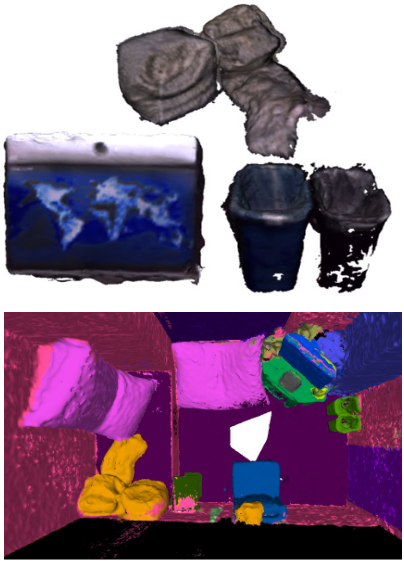} \\
    (a) Room1 & (b) Office1
  \end{tabular}
  \caption{\textbf{Semantic Visualization.} 3D semantic mesh (bottom) and its decomposition with RGB colors (top) for two scenes from the Replica \cite{replica19arxiv} dataset. Images from \cite{li2023dns}.}
  \label{fig:dnsslam}
\end{figure}

Object-centric frameworks introduced a transformative shift in the area. Among them, \hypertarget{vMap}{\textbf{\withcode{vMap} \cite{Kong_2023_CVPR}} introduced a novel approach to object-level dense SLAM, representing each object with dedicated MLPs for watertight and complete object models, even with partial observations. The system efficiently handles object masks for segmentation while leveraging ORB-SLAM3 for tracking.} \hypertarget{SNI-SLAM}{\textbf{SNI-SLAM \cite{zhu2023sni}} employs neural implicit representation with hierarchical semantic encoding, featuring cross-attention mechanisms for integrating appearance, geometry, and semantic features, while its novel decoder ensures unidirectional interaction to prevent mutual interference.} 
\hypertarget{DNS-SLAM}{\textbf{DNS-SLAM \cite{li2023dns}}, instead, leverages 2D semantic priors for stable camera tracking while training class-wise scene representations. By integrating semantic information with multi-view geometry, it achieves comprehensive semantic mesh reconstruction (Figure \ref{fig:dnsslam}), while employing a novel real-time tracking strategy using lightweight coarse scene representation.} The advent of 3D Gaussian Splatting introduced new approaches: \hypertarget{SGS-SLAM}{\textbf{SGS-SLAM \cite{li2024sgsslam}} implements multi-channel optimization during mapping for visual, geometric, and semantic constraints, embedding semantic information in 3D Gaussians through additional color channels.} \hypertarget{NEDS-SLAM}{\textbf{NEDS-SLAM \cite{ji2024neds}} employs Spatially Consistent Feature Fusion combining semantic features with Depth Anything features, using a lightweight encoder-decoder for 3D Gaussian representation and Virtual Camera View Pruning.}
\hypertarget{GS3LAM}{\textbf{GS3LAM \cite{GS3LAM}} introduces Semantic Gaussian Fields with Depth-adaptive Scale Regularization and Random Sampling-based Keyframe Mapping to address scale-invariance and forgetting challenges.} Lastly, \hypertarget{NIS-SLAM}{\textbf{NIS-SLAM \cite{nis_slam}} combines high-frequency tetrahedron-based features with low-frequency positional encoding, implementing semantic probability fusion and confidence-based pixel sampling. 
}

%% file: chapters/taxonomy/rgbd/dynamic.tex
\subsubsection{SLAM in Dynamic Environments}
\label{sec:dynamic-envs}

Most SLAM methods fundamentally assume static environments with rigid, non-moving objects. While effective in static scenes, their performance deteriorates significantly in dynamic environments, limiting real-world applicability. In this section, we provide an overview of the methods that are specifically designed to address the challenges of accurate mapping and localization estimation in dynamic settings.  

Among them, \hypertarget{DN-SLAM}{\textbf{DN-SLAM} \cite{ruan2023dn} addresses this problem through integration of ORB features for object tracking, employing semantic segmentation, optical flow, and the Segment Anything Model (SAM) for precise dynamic object identification and segregation. The system preserves static regions through careful feature extraction and utilizes NeRF for dense map generation.} \hypertarget{DynaMoN}{Building upon DROID-SLAM, \textbf{DynaMoN} \cite{karaoglu2023dynamon} enhances performance through integrated motion and semantic segmentation in dense bundle adjustment. It employs DeepLabV3 for semantic refinement of known object classes while incorporating motion-based filtering for unknown dynamic elements. The framework introduces a 4D scene representation using NeRF, combining implicit and explicit representations with Total Variation loss for regularization.}
\hypertarget{NID-SLAM}{\textbf{NID-SLAM} \cite{xu2024nid}, instead, implements depth-guided semantic mask enhancement for edge region consistency and accurate dynamic object detection. The system performs intelligent background inpainting using static information from previous viewpoints, while its strategic keyframe selection minimizes dynamic object presence for optimized efficiency. The framework employs multi-resolution geometric and color feature grids, jointly optimizing scene representation and camera parameters through geometric and photometric losses.}
\hypertarget{TivNe-SLAM}{\textbf{TivNe-SLAM} \cite{duantivne} advances dynamic scene handling through parallel tracking and mapping processes with time-varying representation. Its two-stage optimization first associates time with 3D positions for deformation field conversion, then links time to canonical field embeddings for color and SDF computation. The system employs motion masks for dynamic object discrimination and implements an overlap-based keyframe selection strategy.}
\hypertarget{RoDyn-SLAM}{\textbf{RoDyn-SLAM} \cite{jiang2024rodynslam} proposes a fusion of optical flow and semantic masks for motion detection, implementing a divide-and-conquer pose optimization that distinguishes between keyframe and non-keyframe frames. Furthermore, the system's edge warp loss strengthens geometric constraints between adjacent frames.} \hypertarget{DG-SLAM}{\textbf{DG-SLAM} \cite{xu2024dgslam} pioneers robust dynamic visual SLAM using 3D Gaussians, achieving precise pose estimation through adaptive Gaussian point management and hybrid camera tracking while maintaining real-time rendering capabilities.} Lastly, \hypertarget{ONeK-SLAM}{\textbf{ONeK-SLAM \cite{onekslam}} integrates feature points with neural radiance fields at the object level, employing joint information for improved localization accuracy and reconstruction detail. The system effectively handles both dynamic objects and illumination variations through joint error analysis.}

%% file: chapters/taxonomy/rgbd/uncertainty.tex
\subsubsection{Uncertainty Estimation}
\label{sec:uncertainty}
Analyzing uncertainties in input data, particularly depth sensor noise, is crucial for robust SLAM processing. This includes both filtering unreliable measurements and incorporating depth uncertainty into optimization processes to prevent inaccuracies that could impact system performance. The field has begun exploring both epistemic and aleatoric uncertainty integration to enhance SLAM reliability, particularly in challenging scenarios. \hypertarget{OpenWorld-SLAM}{\textbf{OpenWorld-SLAM \cite{lisus2023towards}} makes significant progress in this direction by improving upon NICE-SLAM through depth uncertainty integration from RGB-D images, IMU motion information utilization, and a novel division between foreground and background grids for diverse environment handling.} This approach enhances tracking precision while maintaining NeRF-based advantages, though it highlights the need for specialized datasets with outdoor mesh models and well-characterized sensors.
\hypertarget{UncLe-SLAM}{\textbf{\withcode{UncLe-SLAM} \cite{uncleslam2023}} introduces a novel joint learning approach for scene geometry and aleatoric depth uncertainties using the Laplacian error distribution of input depth sensors. Its distinctive feature lies in adaptively weighting different image regions based on confidence levels, achieved without ground truth depth requirements.} This adaptive mechanism not only prioritizes reliable sensor information but also accommodates various sensor configurations with distinct noise characteristics.
\hypertarget{NVINS}{\textbf{NVINS \cite{han2024nvins}} tackles NeRF's computational challenges in real-time robotics applications. By combining NeRF-derived localization with Visual-Inertial Odometry under a Bayesian framework, it effectively addresses positional drift while maintaining system reliability.} \hypertarget{CDA-SLAM}{\textbf{CDA-SLAM \cite{cdaslam}} bridges the gap between explicit and implicit representations through a novel uncertainty modeling approach. Its multi-level feature selection process, combining Bayesian estimation for explicit representation with ray sampling for implicit refinement, demonstrates superior performance in both quantitative and qualitative evaluations while optimizing rendering costs.}

%% file: chapters/taxonomy/rgbd/event.tex
\subsubsection{Event-based SLAM}
\label{sec:event}

\begin{figure}[t]
  \centering
  \begin{tabular}{@{}ccc@{}}
    \includegraphics[width=0.34\linewidth]{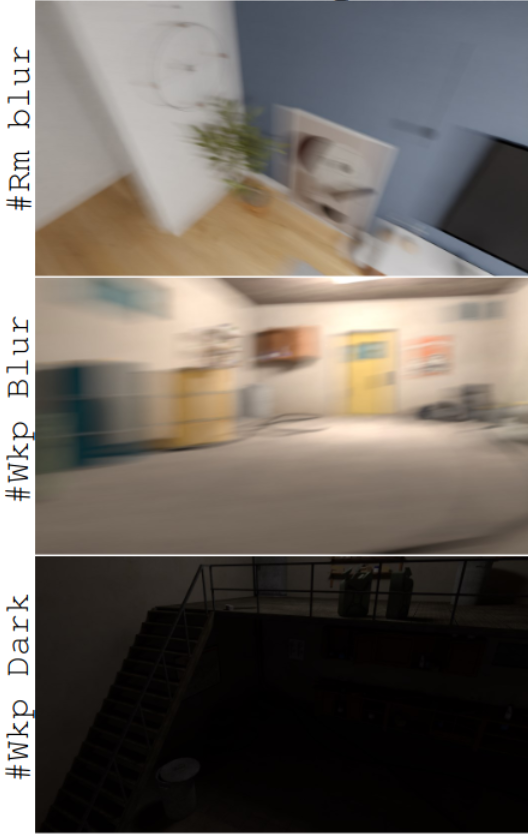}  & \hspace{-1.5em}
    \includegraphics[width=0.32\linewidth]{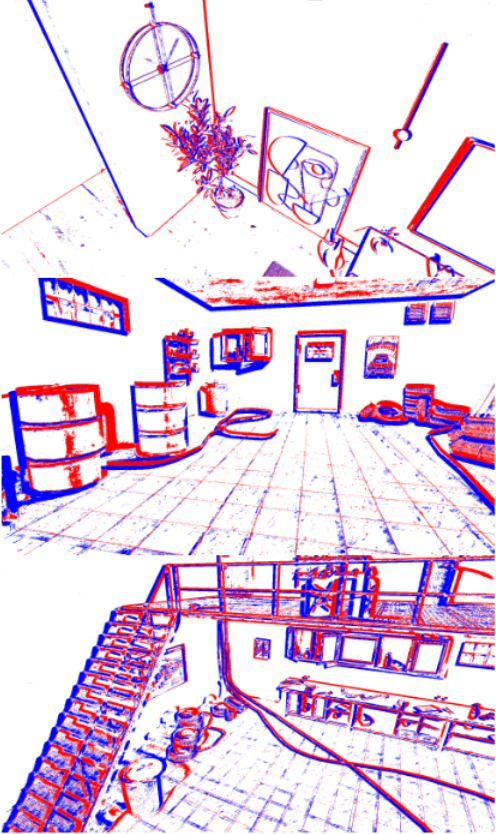}  & \hspace{-1.7em}
    \includegraphics[width=0.32\linewidth]{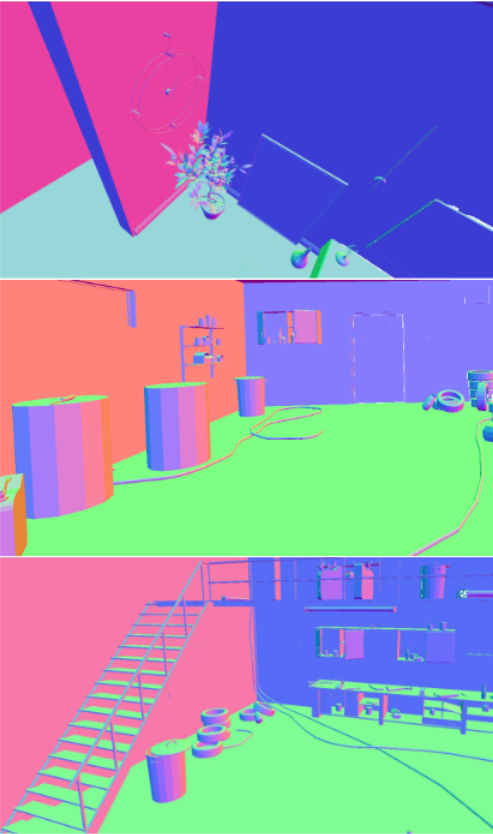}\\
    (a) RGB  & \hspace{-1.5em} (b) Event Input & \hspace{-1.7em} (c) Ground Truth
  \end{tabular}
  \vspace{-0.7em}
  \caption{\textbf{Overview of the DEV-Indoors Dataset \cite{qu2023implicit}.} (a) RGB images depicting normal, motion blur, and dark scenes with corresponding (b) event streams and (c) ground truth meshes. Images from \cite{qu2023implicit}.}
  \label{fig:enslam}
\end{figure}

While radiance field-inspired VSLAM methods offer advantages in accurate dense reconstruction, practical scenarios involving motion blur and lighting variations pose significant challenges that affect the robustness of the mapping and tracking processes. In this section, we explore systems that utilize event cameras, leveraging their dynamic range and temporal resolution. The asynchronous event generation mechanism, triggered by a logarithmic change in luminance at a given pixel, offers advantages in low latency and high temporal resolution, potentially improving neural VSLAM's robustness in extreme environments. While event camera-based SLAM systems are still emerging, they show promise in overcoming traditional RGB-based limitations. \hypertarget{EN-SLAM}{\textbf{EN-SLAM \cite{qu2023implicit}}, as the first work to bridge event cameras with neural implicit representations, exemplifies this potential by integrating event data alongside RGB-D through an implicit neural paradigm. Its novel approach centers on a differentiable Camera Response Function (CRF) rendering technique, unifying event and RGB camera representations. The system decodes scene encoding, establishes a unified geometry and radiance representation, and decomposes shared radiance fields via differentiable CRF Mappers, while implementing optimization strategies for tracking and BA. To validate its effectiveness, EN-SLAM introduces two challenging datasets—DEV-Indoors and DEV-Reals—featuring practical motion blur and lighting variations, as shown in Fig. \ref{fig:enslam}.}

%% file: chapters/taxonomy/rgb/intro.tex
\subsection{RGB-based SLAM Methodologies}
\label{sec:rgb}

This section explores SLAM approaches that rely exclusively on color images, addressing scenarios where depth sensors prove impractical due to their limitations in outdoor environments, sensitivity to lighting conditions, and cost constraints. RGB-only SLAM, using monocular or stereo cameras, offers broader applicability across diverse environments. However, particularly in monocular setups, these methods face unique challenges due to depth ambiguity and the absence of geometric priors, resulting in slower optimization convergence. We organize these approaches into distinct categories: core NeRF-style techniques (\ref{sec:vanilla_rgb}), 3DGS-style techniques (\ref{sec:3dgs_rgb}), systems leveraging external frameworks as supervision during optimization (\ref{sec:priors_driver}), methods incorporating semantic estimation (\ref{sec:segmentation-priors_rgb}), and those addressing system uncertainty (\ref{sec:uncertainty_rgb}).

%% file: chapters/taxonomy/rgb/vanilla.tex
\subsubsection{NeRF-style RGB SLAM}
\label{sec:vanilla_rgb}

\hypertarget{DIM-SLAM}{\textbf{\withcode{DIM-SLAM} \cite{li2023dense}}  presents a SLAM system that employs neural implicit map representation alongside multi-resolution volume encoding and an MLP decoder for depth and color prediction. The framework dynamically learns scene features and decoders on-the-fly, while optimizing occupancies in a single step by fusing features across scales for improved efficiency. The introduced photometric warping loss, inspired by multi-view stereo, enforces alignment between synthesized and observed images while addressing view-dependent intensity variations.} Merging classical and neural approaches, \hypertarget{Orbeez-SLAM}{\textbf{\withcode{Orbeez-SLAM} \cite{chung2023orbeez}} integrates ORB feature-based tracking with neural radiance field modeling. The system achieves real-time performance through parallel tracking and mapping, where ORB-SLAM2-derived visual odometry handles pose estimation while Instant-NGP powers efficient map point generation and bundle adjustment optimization.}
On the other hand, \hypertarget{TT-HO-SLAM}{\textbf{TT-HO-SLAM \cite{lin2023ternary}} introduces an alternative ternary opacity model to overcome the limitations of binary-type opacity priors in rigid 3D scenes. The system's hybrid odometry scheme combines volumetric and warping-based image renderings during tracking for enhanced efficiency, while implementing soft binarization of decoder networks during map initialization. Fine camera odometry adjustments occur during bundle adjustment, jointly optimized with mapping, resulting in superior speed and accuracy through theoretically-grounded opacity optimization.}

\subsubsection{3DGS-style RGB SLAM}
\label{sec:3dgs_rgb}
\hypertarget{MonoGS++}{\textbf{MonoGS++ \cite{monogspp}} extends MonoGS to exploit DPVO \cite{teed2022deep} as an external tracker, used to estimate initial camera poses and 3D points from which 3D Gaussians are bootstrapped. Then, new Gaussians are inserted in new areas guided by planar regularization.}

%% file: chapters/taxonomy/rgb/priors.tex
\subsubsection{Aided Supervision}
\label{sec:priors_driver}

In this section, we explore RGB-based SLAM methods that use external frameworks to integrate regularization information into the optimization process, referred to as aided supervision. These frameworks include various techniques, such as supervision derived from depth estimates obtained from single or multi-view images, surface normal estimation, optical flow, and more. The incorporation of external signals is crucial for disambiguating the optimization process and to significantly improve the performance of SLAM systems using only RGB images as input.

Early efforts focused on multi-threaded architectures and depth supervision. \hypertarget{iMODE}{\textbf{iMODE \cite{matsuki2023imode}} utilizes a multi-threaded architecture with three core processes: a localization process running ORB-SLAM2 on CPU for real-time camera pose estimation and keyframe selection, a semi-dense mapping process enhancing reconstruction through depth-rendered geometry supervision using monocular multi-view stereo methods, and a GPU-based dense reconstruction process optimizing an MLP-based neural field with view dependency and frequency separation features.} The system incorporates view dependency for photometric consistency and frequency separation, using lower frequency embedding for initial input and higher frequency for the color head. Addressing similar challenges, \hypertarget{Hi-SLAM}{\textbf{Hi-SLAM \cite{zhang2023hi}} tackles low texture, rapid movement, and scale ambiguity through DROID-SLAM-based dense correspondence and monocular depth priors. Its joint depth and scale adjustment module resolves scale ambiguity during BA optimization, while Sim(3)-based pose graph bundle adjustment ensures global consistency through online loop closure.}

End-to-end approaches led to innovations in feature representation and loss functions. \hypertarget{NICER-SLAM}{\textbf{\withcode{NICER-SLAM} \cite{Zhu2023NICER}} introduces hierarchical feature grids for SDF and color modeling, combining RGB rendering, warping, and optical flow losses with monocular depth and normal supervision.} Building on this foundation, \hypertarget{NeRF-VO}{\textbf{NeRF-VO \cite{naumann2023nerf}} implements a two-stage approach: first combining DPVO sparse tracking with DPT and Omnidata for comprehensive depth and normal estimation, then employing Nerfacto for dense scene representation with uncertainty-aware optimization.}

Recent advances have focused on sophisticated depth processing and geometric modeling. \hypertarget{MoD-SLAM}{\textbf{MoD-SLAM \cite{zhou2024modslam}} enhances depth estimation through DPT and ZoeDepth architectures with dedicated refinement, while employing multivariate Gaussian encoding for unbounded scenes.} Geometric understanding has been further advanced by \hypertarget{Q-SLAM}{\textbf{Q-SLAM \cite{peng2024q}}, which integrates quadric representations throughout its pipeline, combining DROID-SLAM tracking with quadric-based depth correction and semantic supervision.} The latest developments include \hypertarget{MGS-SLAM}{\textbf{MGS-SLAM \cite{zhu2024mgs}}, which unifies sparse visual odometry with 3D Gaussian Splatting through MVS-derived depth supervision, introducing novel depth smooth loss and adjustment mechanisms to maintain cross-representation consistency}.

%% file: chapters/taxonomy/rgb/segmentation.tex
\subsubsection{Semantic RGB SLAM}
\label{sec:segmentation-priors_rgb}

\hypertarget{RO-MAP}{\textbf{\withcode{RO-MAP}} \cite{RO-MAP}} breaks new ground in multi-object mapping without depth priors, integrating lightweight object-centric SLAM with individual NeRF models for each object. The system achieves real-time performance through efficient loss function design and CUDA implementation, enabling simultaneous localization and object reconstruction from monocular input. Advancing instance-level understanding, \hypertarget{3DIML}{\textbf{3DIML \cite{3diml}} introduces a novel two-phase approach to neural label field learning. The system first employs InstanceMap to associate 2D segmentation masks across images, creating 3D-consistent pseudo-labels, followed by InstanceLift for neural field training that resolves ambiguities and interpolates missing regions. Its InstanceLoc component enables near real-time localization, demonstrating significant speed improvements while maintaining high-quality 3D scene understanding across diverse environments.}

%% file: chapters/taxonomy/rgb/uncertainty.tex
\subsubsection{Uncertainty Estimation}
\label{sec:uncertainty_rgb}

\textbf{\withcode{NeRF-SLAM}\cite{rosinol2023nerf}} \hypertarget{NeRF-SLAM}{ employs real-time implementations of DROID-SLAM \cite{teed2021droid} as the tracking module and Instant-NGP \cite{mueller2022instant} as the hierarchical volumetric neural radiance field map. 
Moreover, it incorporates depth uncertainty estimation to address inherent noise in the input depth maps, used to weight the loss according to depth's marginal covariance.
}

%% file: chapters/taxonomy/lidar/vanilla.tex
\subsection{LiDAR-Based SLAM Strategies}
\label{sec:lidar}

While VSLAM systems discussed so far operate successfully in smaller indoor scenarios where both RGB and dense depth data are available, their limitations become apparent in large outdoor environments where RGB-D cameras are impractical. LiDAR sensors, which provide sparse yet accurate depth information over long distances and in a variety of outdoor conditions, play a critical role in ensuring robust mapping and localization in these settings. However, the sparsity of LiDAR data and the lack of RGB information pose challenges for dense SLAM approaches in outdoor environments. We now explore novel methodologies that exploit the precision of 3D incremental LiDAR data while leveraging radiance field-based scene representations to achieve dense, smooth map reconstruction, even in areas with sparse sensor coverage. Given the limited studies in this domain, we categorize the methodologies into NeRF (\ref{sec:vanilla_lidar}) and 3DGS-style (\ref{sec:multimodal_lidar}) LiDAR-based SLAM approaches.

\subsubsection{NeRF-style LiDAR-based SLAM}
\label{sec:vanilla_lidar}
\hypertarget{NeRF-LOAM}{\textbf{NeRF-LOAM \cite{deng2023nerf}} pioneered this direction by introducing a framework that integrates neural odometry for 6-DoF pose estimation with neural mapping using dynamic voxel embeddings in an octree architecture. Its efficiency derives from a dynamic voxel embedding look-up table and key-scans refinement strategy, effectively addressing catastrophic forgetting during incremental mapping.} Building upon this groundwork, \hypertarget{LONER}{\textbf{\withcode{LONER} \cite{isaacson2023loner}} enhanced the approach through parallel tracking and mapping threads, combining Point-to-Plane ICP for odometry with a hierarchical feature grid-encoded MLP for scene representation. Its novel dynamic margin loss function integrates multiple components to enable adaptive learning while preserving existing geometry.}
Recent developments have further pushed the boundaries of LiDAR-based neural SLAM. In this direction, \hypertarget{PIN-SLAM}{\textbf{\withcode{PIN-SLAM} \cite{pan2024pin}} introduced an elastic point-based implicit neural map representation, alternating between incremental learning of local implicit signed distance fields and correspondence-free registration while incorporating efficient loop closure detection.} Addressing the critical challenge of large-scale applications, \hypertarget{TNDF-Fusion}{\textbf{TNDF-Fusion \cite{TNDF}} developed a compact Tri-Pyramid implicit neural map representation with enhanced supervision through TNDF label rectification. This approach has demonstrated remarkable success in reducing memory consumption while maintaining mapping quality across scales, from room-sized environments to entire city landscapes.}

%% file: chapters/taxonomy/lidar/3DGS.tex
\subsubsection{3DGS-style LiDAR-based SLAM}
\label{sec:multimodal_lidar}

The advent of 3D Gaussian Splatting techniques has opened new possibilities for multimodal sensor fusion in SLAM applications. In this context, \hypertarget{LIV-GaussMap}{\textbf{\withcode{LIV-GaussMap}} \cite{hong2024liv}} develops an integrated LiDAR-Inertial-Visual system using adaptive voxelization for surface representation. This methodology transforms LiDAR data into Gaussian distributions through voxel partitioning, while enhancing reconstruction quality by optimizing spherical harmonics and geometric structures using photometric cues. Extending these advances, \hypertarget{MM-Gaussian}{\textbf{\withcode{MM-Gaussian} \cite{wu2024mm}} introduces a refined fusion architecture that seamlessly integrates LiDAR and visual data streams. The framework orchestrates point cloud registration with image-based refinement, while incorporating a relocalization mechanism to maintain robust tracking. By continuously refining Gaussian attributes via keyframe processing, the system achieves consistent and accurate mapping.}

%% file: chapters/experiments.tex
\section{Experiments and Analysis}
\label{sec:experiments}
In this section, we compare methods across datasets, focusing on tracking (visual in \ref{sec:tracking_visual}, LiDAR in \ref{sec:tracking_lidar}) and 3D reconstruction (visual in \ref{sec:mapping_visual}, LiDAR in \ref{sec:mapping_lidar}). Additionally, we explore novel view synthesis (\ref{sec:rendering}), semantic segmentation (\ref{sec:semantic}), and analyze performance in terms of runtime and memory usage (\ref{sec:runtime}). In each subsequent table, we emphasize the best results within a subcategory using \textbf{bold} and highlight the absolute best in \textcolor{purple}{\textbf{purple}}.
In our analysis, we organized quantitative data from papers with a common evaluation protocol and cross-verified the results. Our priority was to include papers with consistent benchmarks, ensuring a reliable basis for comparison across multiple sources. Although not exhaustive, this approach guarantees the inclusion of methods with verifiable results and a shared evaluation framework in our tables. For performance analysis, we utilized methods with available code to report runtime and memory requirements on a common hardware platform, a single NVIDIA 3090 GPU. For specific implementation details of each method, readers are encouraged to refer to the original papers.

\input{chapters/tables/tum}
\subsection{Visual SLAM Evaluation}

In line with existing protocols, this section compares SLAM systems using RGB-D or RGB data. We evaluate tracking, 3D reconstruction, rendering, and consider runtime and memory usage. Additionally, for methods that estimate semantic segmentation, we assess the quality of the semantic segmentation using the mIoU metric. Specifically, results are presented on the TUM-RGB-D \cite{tum}, Replica \cite{replica19arxiv}, and ScanNet \cite{dai2017scannet} datasets. For semantic segmentation evaluation, we focus on the Replica dataset, as it provides ground truth semantic labels, allowing for a comprehensive comparison of the semantic segmentation performance.

\subsubsection{Tracking}
\label{sec:tracking_visual}
\input{chapters/tables/scannet}
\textbf{TUM-RGB-D.} Table \ref{tab:tum_RGB-D} provides a thorough analysis of camera tracking results on three scenes of the TUM RGB-D dataset, marked by challenging conditions such as sparse depth sensor information and high motion blur in RGB images. Key benchmarks include established methods like Kintinuous, BAD-SLAM, and ORB-SLAM2, representing traditional hand-crafted baselines.

In the RGB-D setting, it is evident that most methods based on radiance field representations generally exhibit lower performance compared to reference methods like BAD-SLAM and ORB-SLAM2. One notable observation, however, is that decoupled methods using external trackers such as ORB3 and DROID, along with advanced strategies such as Global BA and LC, emerge as top performers. Specifically, NGEL-SLAM, 
Q-SLAM, and GO-SLAM demonstrate superior accuracy. 

When shifting the focus to the RGB scenario, ORB-SLAM2  and DROID-SLAM serve as baselines, with ORB-SLAM2 exhibiting superior tracking accuracy.  
Orbeez-SLAM, MoD-SLAM, and MonoGS++, jointly with external tracking components, such as ORB-SLAM2, DROID-SLAM, or DPVO, leads with an ATE RMSE comparable to the one achieved by the best RGB-D methods.

These results emphasize the varied performance of SLAM frameworks, with approaches based on the latest radiance field representations exhibiting effective results in RGB-D scenarios by separating mapping and tracking processes through external tracking approaches and additional optimization strategies. However, when these latter are not applied, most methods still struggle with trajectory drift and sensitivity to noise. 

\begin{figure*}[t]
  \centering
  \begin{tabular}{ccccc}
    \centering
    \begin{overpic}[width=0.2\linewidth]{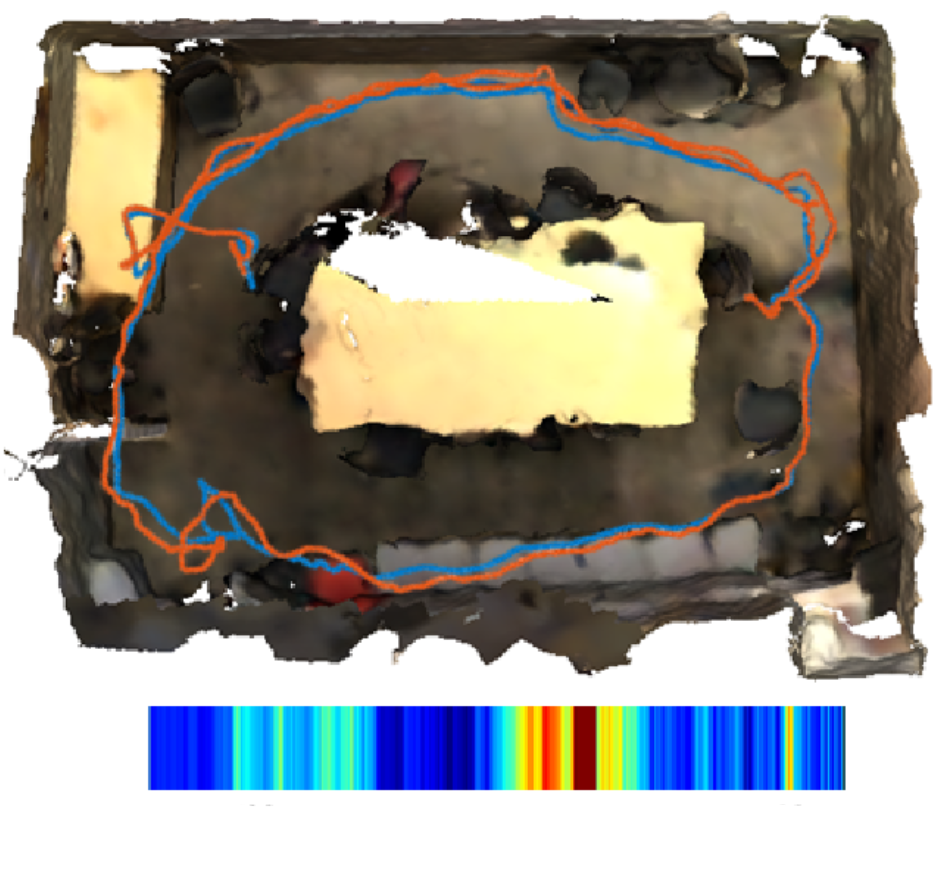}
        \put (23.2,6) {\tiny{500}} 
        \put (40.2,6) {\tiny{1000}} 
        \put (60.,6) {\tiny{1500}} 
        \put (78.6,6) {\tiny{2000}} 
        \put (48,0.5) {\tiny{Frames}} 
    \end{overpic} &
    \begin{overpic}[width=0.2\linewidth]{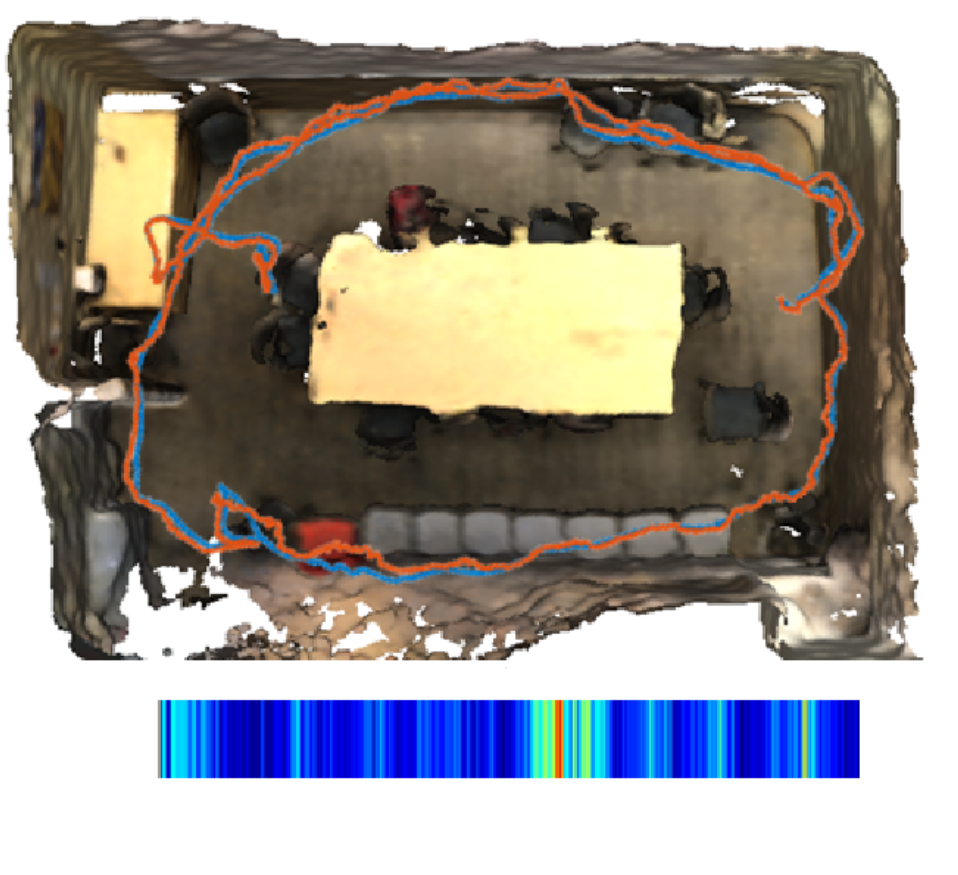}
        \put (23.2,6) {\tiny{500}} 
        \put (40.2,6) {\tiny{1000}} 
        \put (60.,6) {\tiny{1500}} 
        \put (78.6,6) {\tiny{2000}} 
        \put (48,0.5) {\tiny{Frames}} 
    \end{overpic} &
    \begin{overpic}[width=0.2\linewidth]{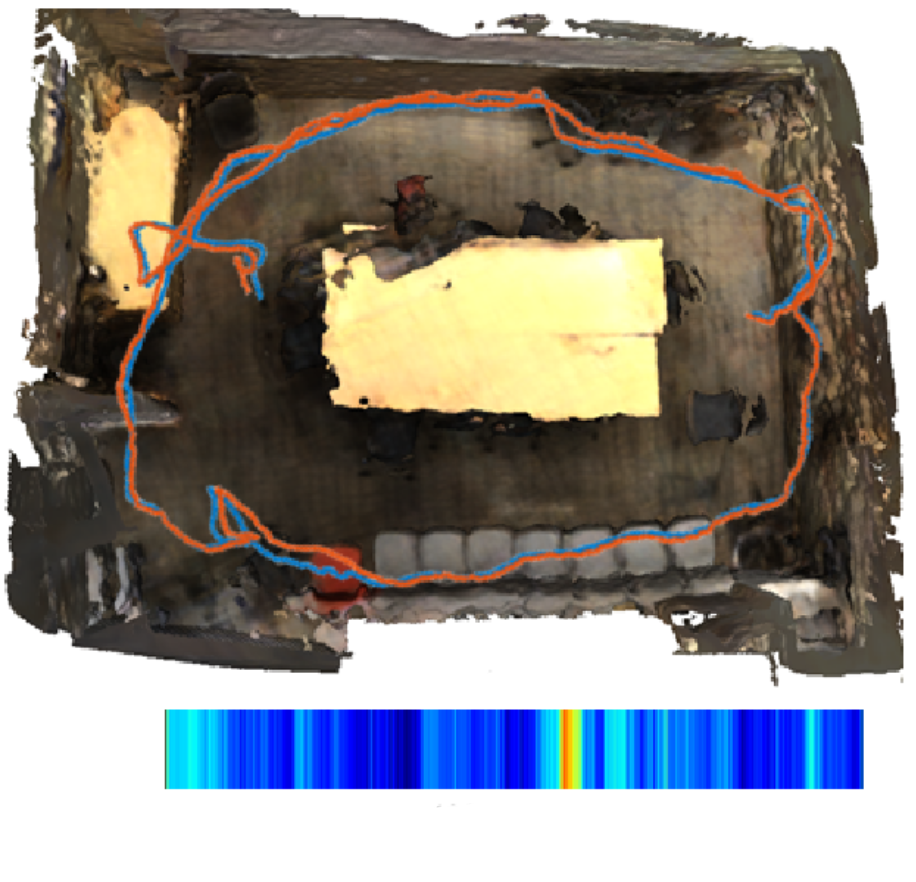}
        \put (23.2,6) {\tiny{500}} 
        \put (40.2,6) {\tiny{1000}} 
        \put (60.,6) {\tiny{1500}} 
        \put (78.6,6) {\tiny{2000}} 
        \put (48,0.5) {\tiny{Frames}} 
    \end{overpic} &
    \begin{overpic}[width=0.2\linewidth]{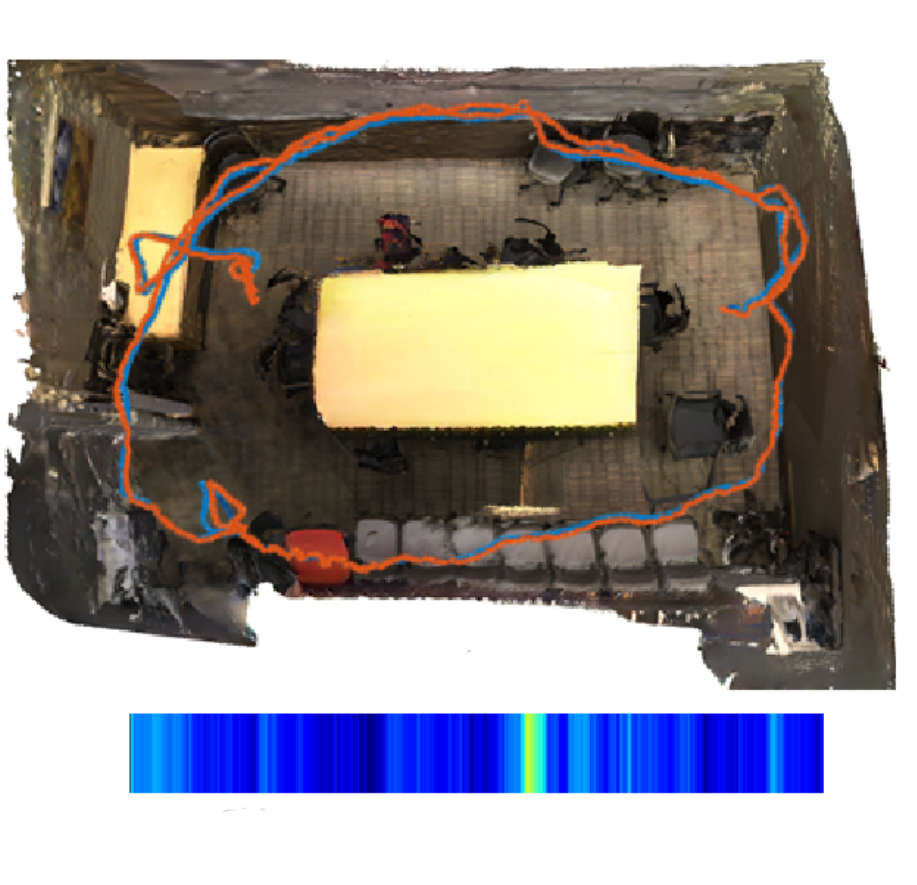}
        \put (23.2,6) {\tiny{500}} 
        \put (40.2,6) {\tiny{1000}} 
        \put (60.,6) {\tiny{1500}} 
        \put (78.6,6) {\tiny{2000}} 
        \put (48,0.5) {\tiny{Frames}} 
    \end{overpic} &
    \begin{overpic}[width=0.05\linewidth]{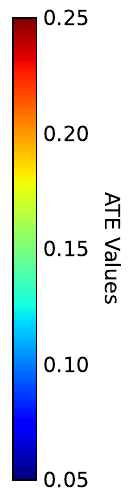}
    \end{overpic} \\
    NICE-SLAM \cite{zhu2022nice} & Co-SLAM \cite{Wang_2023_CVPR} & ESLAM \cite{ESLAM} & PLGSLAM \cite{deng2023plgslam}
 \end{tabular}
 \caption{\textbf{SLAM Methods Comparison on the ScanNet \cite{dai2017scannet} Dataset -- Surface Reconstruction and Localization Accuracy.} Ground truth trajectory in blue, estimated trajectory in orange. ATE visualized with a color bar.}\label{fig:qualitative_scannet}
\end{figure*}

\textbf{ScanNet.} Table \ref{tab:scannet} presents the evaluation of camera tracking methods on six scenes of the ScanNet dataset. In the RGB-D domain, standout performers are the frame-to-frame models MoD-SLAM and GO-SLAM. Both leverage well-crafted visual odometries (such as DROID-SLAM) and LC strategies, with GO-SLAM incorporating also Global BA. Significantly, MoD-SLAM achieves the best average ATE RMSE result of 6.23. It's worth noting that the frame-to-model system SLAIM achieves a competitive 6.32 without requiring additional trackers and by leveraging Global BA. A similar trend can be observed in the RGB case, where once again, the best results are achieved by methods employing external trackers. Nevertheless, it is worth noting that these solutions manage to be comparable or even superior to many other SLAM methods that leverage depth information from RGB-D sensors.
In Figure \ref{fig:qualitative_scannet}, we report some qualitative results from selected RGB-D SLAM systems on ScanNet, highlighting recent improvements in trajectory error compared to the seminal systems.
\input{chapters/tables/replica_tracking}
\input{chapters/tables/replica_mapping}

\textbf{Replica.} Table \ref{tab:replica_tracking} evaluates camera tracking across eight scenes from Replica, using higher-quality images compared to challenging counterparts like ScanNet and TUM RGB-D. The evaluation includes the reporting of ATE RMSE results for each individual scene, alongside the averaged outcomes. 

On top, we report the evaluation concerning RGB-D methods. In line with observations from TUM RGB-D and ScanNet datasets, the highest accuracy is achieved by leveraging external tracking and methodologies involving Global BA and/or LC. In particular, GO-SLAM, Loopy-SLAM, and MoD-SLAM (in its RGB-D version) once again stand out on Replica, confirming their effectiveness in optimizing camera tracking accuracy. Additionally, promising results are evident for methods utilizing 3D Gaussian Splatting, with the best results among all achieved with CG-SLAM, HF-GS SLAM and GS-ICP SLAM. This suggests that these approaches struggle with noise and work best in simpler situations, showing less reliability in complex conditions -- similarly to what was observed in the TUM RGB-D and ScanNet datasets. 

At the bottom, we collect results achieved by RGB-only frameworks. Again, 
we observe a substantial superiority of frame-to-frame models that exploit external trackers such as DROID-SLAM or DPVO, with MonoGS++ standing among the others in terms of average accuracy.

\begin{figure*}[t]
    \centering
    \renewcommand{\tabcolsep}{1pt}
    \begin{tabular}{ccccc}
        \rotatebox[origin=l]{90}{\quad Room 0} \includegraphics[clip, trim=0cm 0cm 0cm 2cm, width=0.19\textwidth]{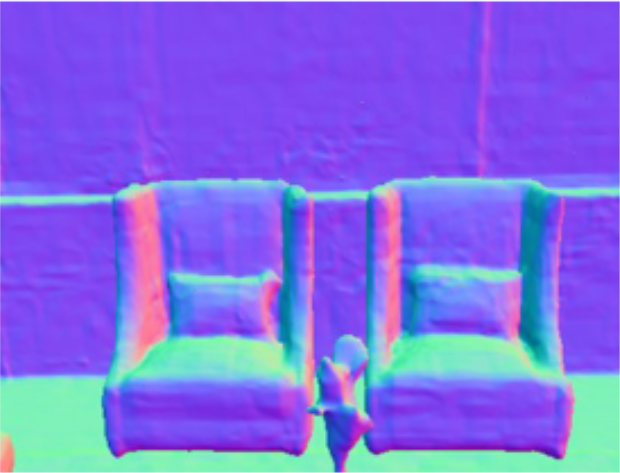} & 
        \includegraphics[clip, trim=0cm 0cm 0cm 2cm, width=0.19\textwidth]{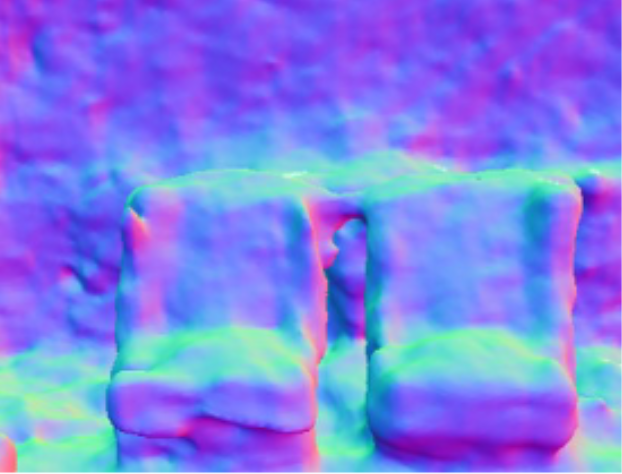} &
        \includegraphics[clip, trim=0cm 0cm 0cm 2cm, width=0.19\textwidth]{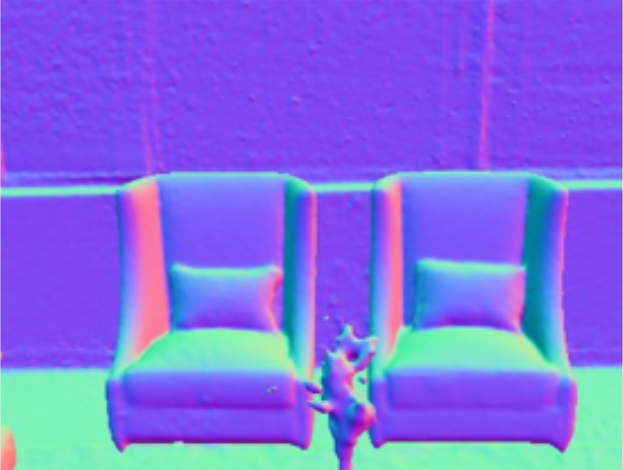} &
        \includegraphics[clip, trim=0cm 0cm 0cm 2cm, width=0.19\textwidth]{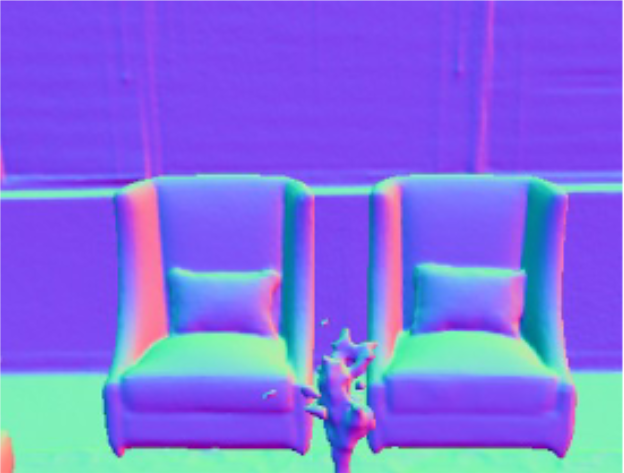} &
        \includegraphics[clip, trim=0cm 0cm 0cm 2cm, width=0.19\textwidth]{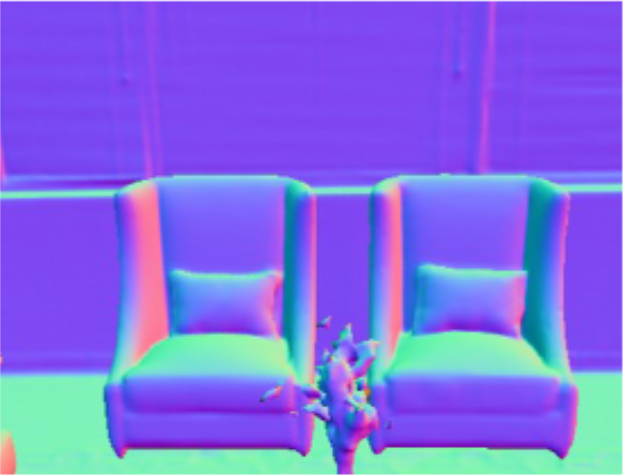}
        \\
        ESLAM \cite{ESLAM} & GO-SLAM \cite{zhang2023go} & Point-SLAM \cite{Sandström2023ICCV} & Loopy-SLAM \cite{liso2024loopy} & Ground-Truth
    \end{tabular}
    \vspace{-0.2cm}
    \caption{\textbf{SLAM Methods Comparison on the Replica \cite{replica19arxiv} Dataset -- Mapping.}
    Images sourced from \cite{liso2024loopy}.}
    \label{fig:qualitative_mapping_replica}
\end{figure*}

\subsubsection{Mapping}
\label{sec:mapping_visual}

\textbf{Replica.}  In Table \ref{tab:replica_mapping}, we provide mapping results according to the evaluation protocol proposed in \cite{zhu2022nice}, highlighting the performance in terms of both 3D reconstruction and 2D depth estimation on the Replica dataset.
Examining the table, a noticeable progression in both 3D reconstruction and 2D depth estimation metrics is observed, showing an improvement from iMap to more recent methods such as VPE-SLAM and HERO-SLAM. Notably, Loopy-SLAM leads in the L1-Depth metric, closely followed by Point-SLAM.  This suggests that the neural point representation holds significant promise for generating highly accurate scene reconstructions. In terms of 3D error metrics, Vox-Fusion++, NIS-SLAM, HERO-SLAM and VPE-SLAM outperform other methods, even surpassing hand-crafted approaches like COLMAP and TSDF. Point-SLAM performs comparably, excelling in the Accuracy metric with a value of 1.41, while CG-SLAM achieves the best overall performance with a score of 1.01. 
Notably, despite GO-SLAM's notable achievements in tracking, it holds a relatively low position in this ranking, indicating challenges for the mapping process. In Figure \ref{fig:qualitative_mapping_replica}, qualitatives from a subset of reviewed systems on Replica are presented, emphasizing specific improvements achieved by recent methods in the mapping process.

Shifting the focus to RGB methods, NICER-SLAM and Hi-SLAM show a balanced performance with competitive scores in both Accuracy and Completion metrics. However, among these,
MoD-SLAM stands out as the most accurate, while MGS-SLAM achieves the highest completion rate. Notably, Q-SLAM achieves the best L1-Depth metric with a score of 2.76. Nonetheless, the distinction among different methods in the RGB context is less pronounced compared to RGB-D scenarios. As expected, methods relying solely on RGB perform less favorably than those leveraging depth sensor information, with iMAP being the only exception to this trend. This emphasizes the crucial role of depth sensors in SLAM and points towards the potential for advancements in RGB-only methodologies. 

\input{chapters/tables/replica_rendering}

\subsubsection{Image Rendering}
\label{sec:rendering}

\textbf{Replica.} Table \ref{tab:replica_rendering} presents the rendering quality evaluation on Replica's training input views, following the standard evaluation protocol established by Point-SLAM and NICE-SLAM.

On top, we focus on RGB-D frameworks: recent solutions, particularly those based on Gaussian Splatting or neural points such as Point-SLAM, achieve significantly better average metrics in PSNR, SSIM, and LPIPS compared to earlier neural SLAM methods (showing an improvement of over 10dB in PSNR). These earlier approaches relied on multi-resolution feature grids like NICE-SLAM or voxel-based neural implicit surface representations like Vox-Fusion. This demonstrates that paradigms based on explicit Gaussian primitives or neural points lead to substantial improvements in image rendering. Among these, GS-ICP SLAM achieves the highest performance with an average PSNR of 38.83, highlighting the effectiveness of Gaussian-based approaches for high-quality image rendering.
Regarding RGB-only methods, the adoption of the 3DGS framework enables Photo-SLAM and MonoGS++ to produce novel view renderings with superior quality compared to other NeRF-style SLAM systems.

\begin{figure*}[t]
    \centering
    \renewcommand{\tabcolsep}{1pt}
    \begin{tabular}{ccccc}
        \rotatebox[origin=l]{90}{\quad Room0} \includegraphics[width=0.19\textwidth]{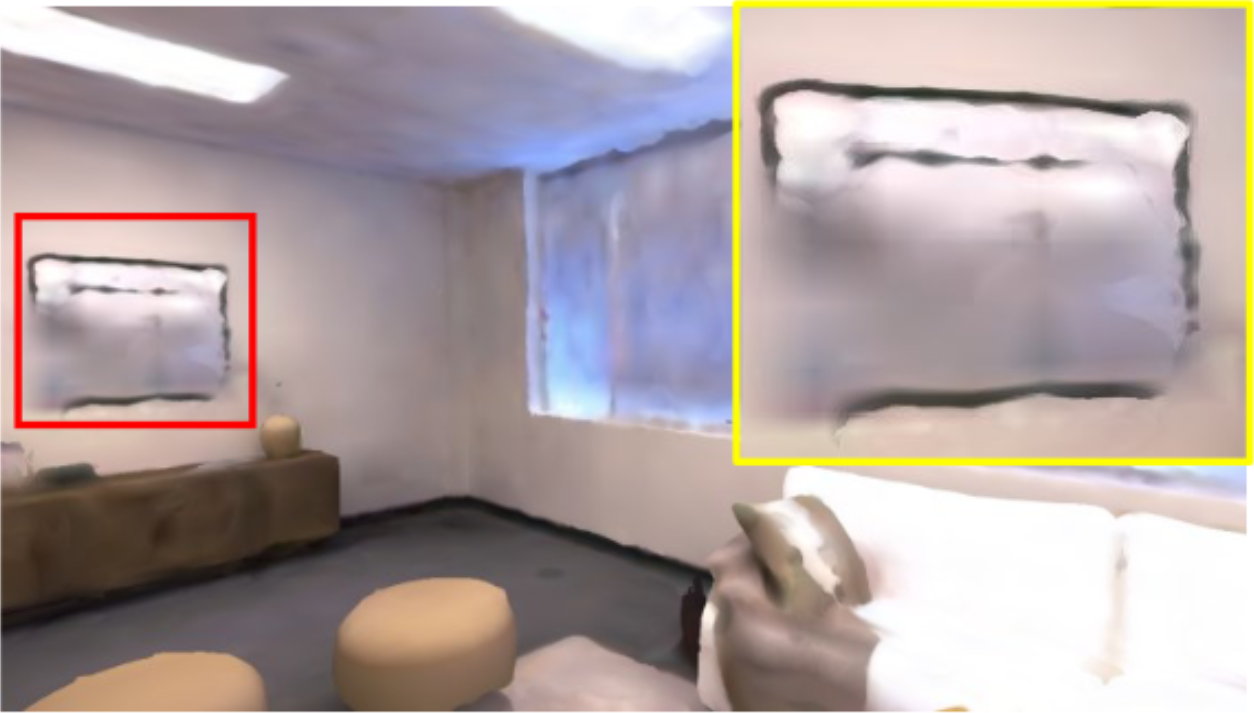} & 
        \includegraphics[width=0.19\textwidth]{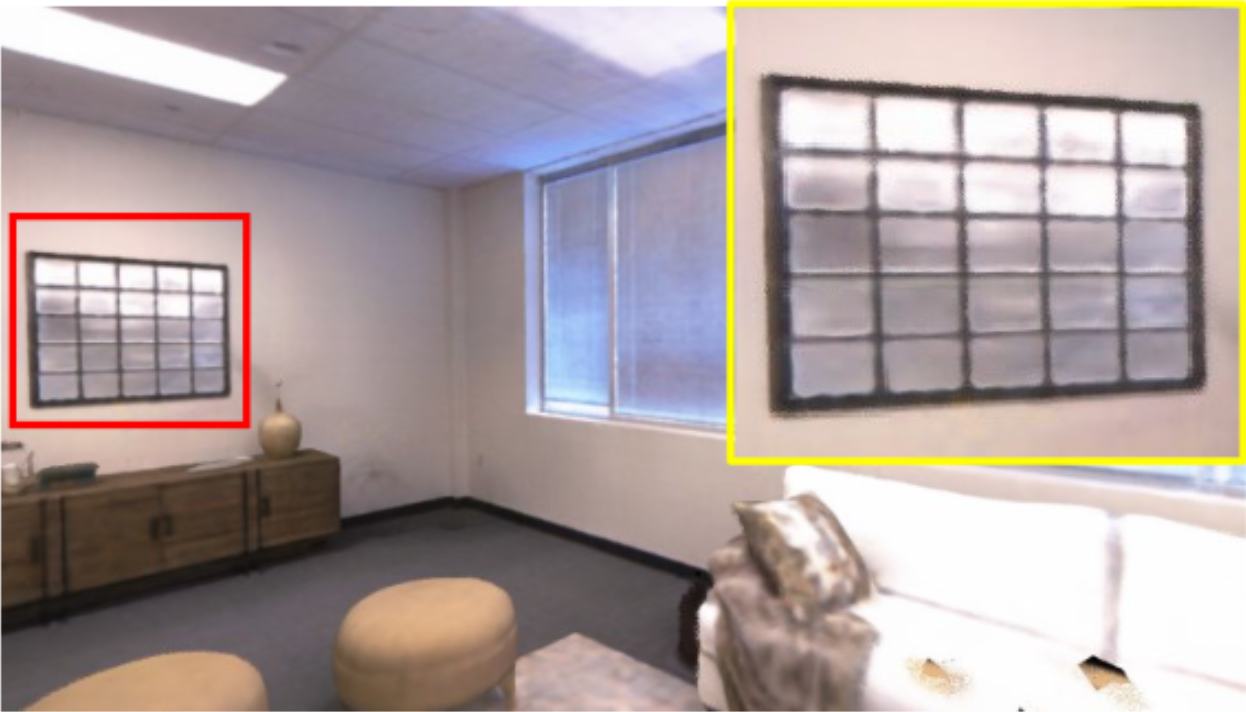} &
        \includegraphics[width=0.19\textwidth]{images/room0_coslam.png} &
        \includegraphics[width=0.19\textwidth]{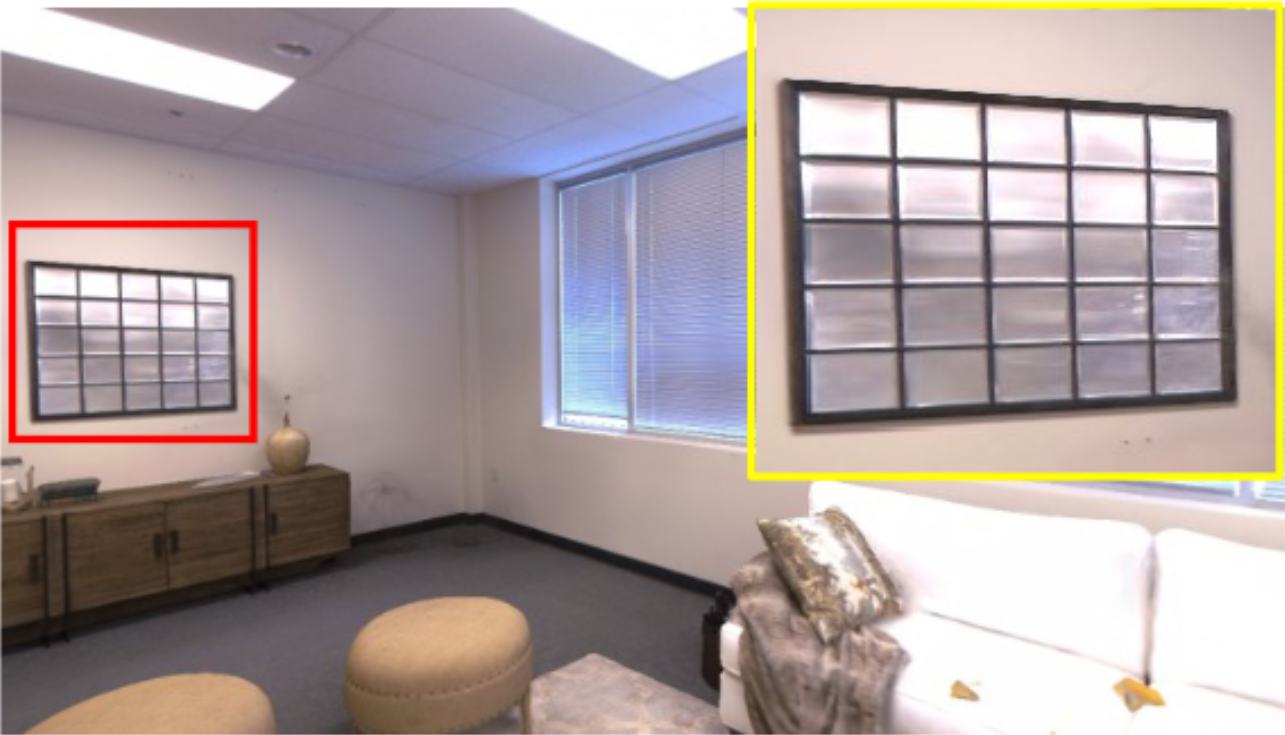} &
        \includegraphics[width=0.19\textwidth]{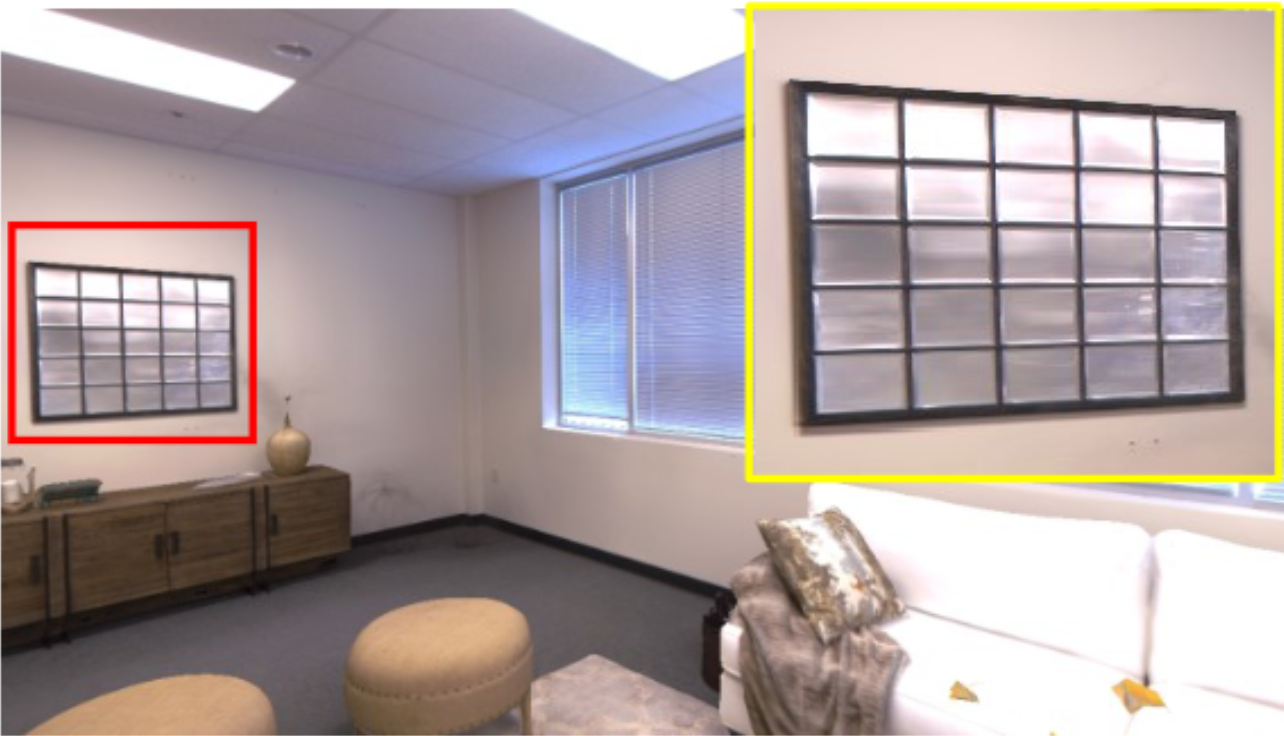}
        \\
        NICE-SLAM \cite{zhu2022nice} & Co-SLAM \cite{Wang_2023_CVPR} & ESLAM \cite{ESLAM} & GS-SLAM \cite{yan2023gs} & Ground-Truth
    \end{tabular}
    \vspace{-0.2cm}
    \caption{\textbf{SLAM Methods Comparison on the Replica \cite{replica19arxiv} Dataset-- Image Rendering.}
    Images sourced from \cite{yan2023gs}.}
    \label{fig:qualitative_replica}
\end{figure*}

In Figure \ref{fig:qualitative_replica}, we present qualitative results for image rendering from selected RGB-D SLAM systems on Replica. The latest frameworks demonstrate improved rendering of fine details, with GS-SLAM showing superior rendering quality due to its 3DGS representation.

In our analysis, we concur with the concerns raised in the SplaTAM paper regarding the evaluation of rendering results on the Replica dataset. Assessing the same training views used as input may introduce biases due to high model capacity and potential overfitting. We support exploring alternative methods for evaluating novel view rendering in this context, acknowledging the limitations of current SLAM benchmarks.

\subsubsection{Semantic Segmentation Results}
\label{sec:semantic}

\textbf{Replica.}
Table \ref{tab:replica_semantic} presents a comparative analysis of state-of-the-art RGB-D semantic SLAM methods on the Replica dataset \cite{replica19arxiv}, using the mIoU metric for evaluating the semantic segmentation performance of input views, following the evaluation protocol from SemGauss-SLAM \cite{zhu2024semgauss}. 
The table highlights the use of external priors, such 
Dinov2 \cite{oquab2023dinov2} or hand-made semantic segmentation masks \cite{GS3LAM}, by some of these methods to improve their semantic understanding capabilities. Among the compared methods, SemGauss-SLAM achieves the highest mIoU scores across all eight scenes of Replica, demonstrating its superior performance in semantic segmentation.

\subsection{LiDAR SLAM/Odometry Evaluation}

\subsubsection{Tracking}
\label{sec:tracking_lidar}

\textbf{KITTI}. Table \ref{tab:lidar_kitti} presents the evaluation of LiDAR SLAM strategies on the KITTI dataset, detailing odometry accuracy at the top and SLAM performance metrics at the bottom. The odometry section reports the average relative translational drift error (\%) and highlights the performance of PIN-LO, a variant of PIN-SLAM that disables the loop closure detection correction and pose graph optimization modules. PIN-LO outperforms several LiDAR odometry systems using different map representations (feature points \cite{pan2021mulls}, denser voxel downsampling points \cite{dellenbach2022ct}, normal distribution transformation \cite{yokozuka2021litamin2}, surfels \cite{behley2018efficient} and triangle meshes \cite{ruan2023slamesh}) achieving an impressive translation error of 0.5\%, competing with 
CT-ICP and outperforming the neural implicit approach Nerf-LOAM due to improved SDF training and robust point-to-SDF registration.

In the LiDAR SLAM evaluation at the bottom of the table \ref{tab:lidar_kitti}, the ATE RMSE [m] is used as the evaluation metric. As a representative of implicit LiDAR-based SLAM strategies, PIN-SLAM consistently outperforms state-of-the-art LiDAR SLAM systems. Specifically, PIN-SLAM achieves an average RMSE of 1.1 m on sequences with loop closure and 1.0 m over all eleven sequences. The results of PIN-LO underscore the significant improvement of PIN-SLAM in ensuring global trajectory consistency.

\input{chapters/tables/replica_semantic}
\input{chapters/tables/kitti_lidar}
\input{chapters/tables/college_tracking}
\input{chapters/tables/college_mapping}

\textbf{Newer College}. Table \ref{tab:college_tracking} reports the tracking accuracy on the Newer College dataset, measured in terms of ATE RMSE [cm]. Again, we can observe how PIN-SLAM consistently outperforms PIN-LO, with an average RMSE of 0.19 cm over the whole set of sequences, which is 5$\times$ lower compared to PIN-LO. This further confirms the superiority of PIN-SLAM at global trajectory tracking.

\subsubsection{Mapping}
\label{sec:mapping_lidar}

\textbf{Newer College}. Table \ref{tab:recon_experiments_on_ncd} collects the results concerning the quality of 3D reconstruction on the New College dataset -- specifically, on \textit{Quad} and \textit{Math Institute} sequences. Accuracy and Completeness scores are used to assess the effectiveness of Nerf-LOAM and PIN-SLAM, with the latter confirming again as the best LiDAR-based SLAM system among those evaluating on this dataset. In particular, on \textit{Quad} we can appreciate a large margin in terms of completeness between PIN-SLAM and Nerf-LOAM -- \ie, about 7 cm.

\input{chapters/tables/performance}

\subsection{Performance Analysis}
\label{sec:runtime}

We conclude the experimental studies by considering the efficiency of the SLAM systems reviewed so far. For this purpose, we run methods with source code publicly available and measure 1) the GPU memory requirements (as the peak memory use in GB) and 2) the average FPS (computed as the total time required to process a single sequence, divided by the total amount of frames in it) achieved on a single NVIDIA RTX 3090 board.
Table \ref{tab:performance_analysis} collects the outcome of our benchmark for RGB-D and RGB systems running on Replica, sorted in increasing order of average FPS. 
On top, we consider RGB-D frameworks: we can notice how SplaTAM, despite its high efficiency at rendering images, is however much slower at processing both tracking and mapping simultaneously. This is also the case for hybrid methods using hierarchical feature grids,  on the other hand require much less GPU memory -- 4 to 5$\times$ lower compared to SplaTAM. Finally, the use of more advanced representations such as hash grids or point features allows for much faster processing. This is confirmed also by the studies on the RGB-only methods, in the middle, with NeRF-SLAM resulting 6$\times$ faster than DIM-SLAM.
Finally, concerning LiDAR SLAM systems, we can observe how PIN-SLAM is much more efficient than Nerf-LOAM, requiring as few as 7 GB of GPU memory while running at nearly 7 FPS, compared to the nearly 12 GB and 4 seconds per frame required by Nerf-LOAM.

This analysis highlights how, despite the great promise brought by this new generation of SLAM systems, most of them are still unsatisfactory in terms of hardware and runtime requirements, making them not yet ready for real-time applications.

%% file: chapters/tables/tum.tex
\begin{table}[t]
    \centering
    \caption{\textbf{TUM RGB-D \cite{tum} Camera Tracking Results}. ATE RMSE [cm] ($\downarrow$) is used as the evaluation metric.}
    \resizebox{0.5\textwidth}{!}{
      \setlength\tabcolsep{5pt}{}
      \renewcommand\arraystretch{1.05}
      \rowcolors{2}{salmon}{white}
      \begin{tabular}{l|ccc|ccc|c}
         \Xhline{2pt}
          \multirow{2}{*}{\rowcolors{white}{}{}Method} & Tracker & Global & Loop & \cellcolor{lower} & \cellcolor{lower} & \cellcolor{lower} & \cellcolor{lower} \\
         & Based on & BA & Closure & \multirow{-2}{*}{\cellcolor{lower} fr1/desk} & \multirow{-2}{*}{\cellcolor{lower} fr2/xyz} & \multirow{-2}{*}{\cellcolor{lower} fr3/office} & \multirow{-2}{*}{\cellcolor{lower}  Avg ($\downarrow$)} \\
         \Xhline{1pt}
         \rowcolor{maincategories}\multicolumn{8}{c}{RGB-D} \\
         \Xhline{1pt}
         {Kintinuous}\cite{kinectfusion} &   &  &  & 3.7 &  2.9  & 3.0 & 3.2 \\  
         {BAD-SLAM}\cite{schops2019bad} &   & \checkmark & \checkmark & 1.7  & 1.1 & 1.7 & 1.5 \\  
         {ORB-SLAM2}\cite{mur2017orb} &   & \checkmark & \checkmark & \bf 1.6  & \bf 0.4  & \textcolor{purple}{\bf 1.0} & \textcolor{purple}{\bf 1.0} \\ 
         \hline
         \hyperlink{Vox-Fusion}{\textcolor{magenta}{Vox-Fusion}} ~\cite{yang2022vox} &  &  &  & 3.5 & 1.5 & 26.0 & 10.3 \\     
         \hyperlink{MeSLAM}{\textcolor{magenta}{MeSLAM}} ~\cite{kruzhkov2022meslam}  &  &  &  & 6.0 & 6.5 & 7.8 & 6.8 \\
         \hyperlink{iMAP}{\textcolor{magenta}{iMAP}}~\cite{sucar2021imap} &  &  &  & 4.9 & 2.0 & 5.8 & 4.2  \\  
         \hyperlink{GS-SLAM}{\textcolor{magenta}{GS-SLAM}}~\cite{GS-SLAM} &  &   &  & 3.3 & 1.3 & 6.6 & 3.7 \\  
         \hyperlink{SplaTAM}{\textcolor{magenta}{SplaTAM}}~\cite{keetha2023splatam} &  &  &  & 3.4 & 1.2 & 5.2 & 3.3 \\ 
         \hyperlink{HF-GS SLAM}{\textcolor{magenta}{HF-GS SLAM}}~\cite{sun2024high} &  &  &  & 3.4 & - & 5.1 & - \\ 
         
         \hyperlink{MIPS-Fusion }{\textcolor{magenta}{MIPS-Fusion }}~\cite{ tang2023mips} &  &  & \checkmark & 3.0 & 1.4 & 4.6 & 3.0 \\
         \hyperlink{Point-SLAM}{\textcolor{magenta}{Point-SLAM}}~\cite{Sandström2023ICCV} &  &  &  & 4.3 & 1.3 & 3.5 & 3.0 \\
        
        \hyperlink{Loopy-SLAM}{\textcolor{magenta}{Loopy-SLAM}}~\cite{liso2024loopy} &  &  & \checkmark &  3.8 & 1.6 & 3.4 &  2.9  \\   
         \hyperlink{NICE-SLAM}{\textcolor{magenta}{NICE-SLAM}}~\cite{zhu2022nice}  &   &  &  & 2.7 & 1.8 & 3.0 & 2.5 \\

         \hyperlink{GS-ICP SLAM}{\textcolor{magenta}{GS-ICP SLAM}}~\cite{ha2024rgbd}  & G-ICP \cite{segal2009generalized} &   &  & 2.7 & 1.8 & 2.7 & 2.4 \\
         \hyperlink{vMAP}{\textcolor{magenta}{vMAP}}~\cite{Kong_2023_CVPR}  & ORB3 \cite{campos2021orb} &   &  & 2.6 & 1.6 & 3.0 & 2.4 \\
         \hyperlink{Co-SLAM}{\textcolor{magenta}{Co-SLAM}}~\cite{Wang_2023_CVPR} &  & \checkmark &  & 2.4 & 1.7 & 2.4 & 2.2 \\ 
         \hyperlink{NIS-SLAM}{\textcolor{magenta}{NIS-SLAM}}~\cite{nis_slam} & & \checkmark & & 2.3 & 1.6 & 2.4 & 2.1 \\
         \hyperlink{LRSLAM}{\textcolor{magenta}{LRSLAM}}~\cite{park2024lrslam} & & & & 2.5 & 1.0 & 2.8 & 2.1 \\

         \hyperlink{ESLAM}{\textcolor{magenta}{ESLAM}}~\cite{ESLAM} &  &  &  & 2.5 & 1.1 & 2.4 & 2.0 \\  
         \hyperlink{CG-SLAM}{\textcolor{magenta}{CG-SLAM}}~\cite{hu2024cg} &  &  &  & 2.4 & 1.2 & 2.5 & 2.0 \\ 
         \hyperlink{IBD-SLAM}{\textcolor{magenta}{IBD-SLAM}}~\cite{IBDSLAM} & SuperPoint \cite{detone2018superpoint} \& SuperGlue \cite{sarlin20superglue} & & & 1.7 & 1.6 & 2.6 & 2.0 \\
        \hyperlink{SLAIM}{\textcolor{magenta}{SLAIM}}~\cite{cartillier2024slaim} & & \checkmark & & 2.1 & 1.5 & 2.3 & 2.0 \\ 
        \hyperlink{MonoGS}{\textcolor{magenta}{MonoGS}}~\cite{matsuki2023gaussian} &   &  &  &  1.5 &  1.6 &  1.7 &  1.6 \\ 
         \hyperlink{GO-SLAM}{\textcolor{magenta}{GO-SLAM}}~\cite{zhang2023go} & DROID \cite{droidslam} & \checkmark & \checkmark &  1.5 &  0.6 & 1.3 &  1.1 \\  
        \hyperlink{RTG-SLAM}{\textcolor{magenta}{RTG-SLAM}}~\cite{peng2024rtg} & & & & 1.7 & 0.4 & 1.1 & 1.1 \\
         \hyperlink{Q-SLAM}{\textcolor{magenta}{Q-SLAM}}~\cite{peng2024q} & DROID \cite{droidslam} &  &  & \textcolor{purple}{\textbf{1.4}} &  0.5 & 1.1 &  \textcolor{purple}{\bf 1.0} \\  
        \hyperlink{NGEL-SLAM}{\textcolor{magenta}{NGEL-SLAM}}~\cite{mao2023ngel} & ORB3 \cite{campos2021orb} & \checkmark & \checkmark &  1.5 & 0.5 & \textcolor{purple}{\bf 1.0} &  \textcolor{purple}{\bf 1.0} \\  

         \hyperlink{ONeK-SLAM}{\textcolor{magenta}{ONeK-SLAM}}~\cite{onekslam} & SIFT matching \cite{lowe2004distinctive} & & & 1.5 & \textcolor{purple}{\bf 0.3} & 1.1 & \textcolor{purple}{\bf 1.0} \\

        \Xhline{1pt}
        \rowcolor{maincategories}\multicolumn{8}{c}{RGB} \\
        \Xhline{1pt}
        {DROID-SLAM}\cite{droidslam}  & - & \checkmark &  & \bf 1.8 & \bf 0.5 & 2.8 & 1.7 \\  
        {ORB-SLAM2}\cite{mur2017orb} & - & \checkmark & \checkmark & 1.9  & 0.6  & \bf 2.4 & \bf 1.6 \\ 
        \hline
        \hyperlink{MonoGS}{\textcolor{magenta}{MonoGS}}~\cite{matsuki2023gaussian} &   &  &  & 4.2 &  4.8 &  4.4 &  4.4 \\ 
        \hyperlink{DDN-SLAM}{\textcolor{magenta}{DDN-SLAM}}~\cite{li2024ddn} & ORB3 \cite{campos2021orb} & \checkmark & \checkmark & 1.9 & 2.4 & 2.9 & 2.4  \\ 
        \hyperlink{MGS-SLAM}{\textcolor{magenta}{MGS-SLAM}}~\cite{zhu2024mgs} & DPVO \cite{teed2022deep} & \checkmark & & 2.3 & \textcolor{purple}{\textbf{0.4}} & 3.0 & 1.9 \\
        \hyperlink{DIM-SLAM}{\textcolor{magenta}{DIM-SLAM}}~\cite{li2023dense} &  &  &   & 2.0 &  0.6 &  2.3 &  1.6 \\ 
        \hyperlink{I$^2$-SLAM}{\textcolor{magenta}{I$^2$-SLAM}}~\cite{bae2024i2slam} & &  & & 1.6 & 0.3 & 2.0 & 1.3 \\
         \hyperlink{GO-SLAM}{\textcolor{magenta}{GO-SLAM}}~\cite{zhang2023go} & DROID \cite{droidslam} & \checkmark & \checkmark &  1.6 & 0.6 & 1.5 &  1.2 \\ 
        \hyperlink{Orbeez-SLAM}{\textcolor{magenta}{Orbeez-SLAM}}~\cite{chung2023orbeez} & ORB2 \cite{mur2017orb} &   &  & 1.9 &  0.3 & 1.0 & 1.1 \\ 
         \hyperlink{MoD-SLAM}{\textcolor{magenta}{MoD-SLAM}}~\cite{zhou2024modslam} & DROID \cite{droidslam} &  & \checkmark & \textcolor{purple}{\textbf{1.5}} & 0.7 &  1.1 & 1.1 \\ 

        \hyperlink{MonoGS++}{\textcolor{magenta}{MonoGS++}}~\cite{monogspp} & DPVO \cite{teed2022deep} &  & & 1.8 & \textcolor{purple}{\textbf{0.4}} & \textcolor{purple}{\textbf{0.4}} & \textcolor{purple}{\textbf{0.9}} \\

         \Xhline{2pt}
       \end{tabular}
    }
    \label{tab:tum_RGB-D}
\end{table}

%% file: chapters/tables/scannet.tex
\begin{table}[t]
    \centering
    \caption{\textbf{ScanNet\cite{dai2017scannet} Camera Tracking Results}. ATE RMSE [cm] ($\downarrow$) is used as the evaluation metric.}
    \resizebox{0.48\textwidth}{!}{
      \setlength\tabcolsep{5pt}{}
      \renewcommand\arraystretch{1.05}
      \rowcolors{2}{salmon}{white}
      \begin{tabular}{l|ccc|cccccc|c}
         \Xhline{2pt}
         \multirow{2}{*}{\rowcolors{white}{}{}Method} &  Tracker & Global & Loop & \cellcolor{lower} & \cellcolor{lower} & \cellcolor{lower} & \cellcolor{lower} & \cellcolor{lower} & \cellcolor{lower} & \cellcolor{lower} \\
         & Based on & BA & Closure & \multirow{-2}{*}{\cellcolor{lower} 0000} & \multirow{-2}{*}{\cellcolor{lower} 0059} & \multirow{-2}{*}{\cellcolor{lower} 0106} & \multirow{-2}{*}{\cellcolor{lower} 0169} & \multirow{-2}{*}{\cellcolor{lower} 0181} & \multirow{-2}{*}{\cellcolor{lower} 0207} & \multirow{-2}{*}{\cellcolor{lower}  Avg ($\downarrow$)} \\

        \Xhline{1pt}
        \rowcolor{maincategories}\multicolumn{11}{c}{RGB-D} \\
        \Xhline{1pt}

         DROID-SLAM (VO)~\cite{droidslam} &   &  &  & 8.00   & 11.30   & 9.97  & 8.64  & 7.38  & - & - \\ 
         DROID-SLAM~\cite{droidslam} &   & \checkmark &  & \textbf{5.36}  & \textbf{7.72}  & \textbf{7.06} & \textbf{8.01} & \textbf{6.97}  & - & - \\ 
         
         \Xhline{1pt} 
         \hyperlink{iMAP}{\textcolor{magenta}{iMAP}}~\cite{sucar2021imap}  &  &  &  & 55.95 & 32.06 & 17.50 & 70.51 & 32.10 & 11.91 & 36.67 \\  
         \hyperlink{ADFP}{\textcolor{magenta}{ADFP}}~\cite{Hu2023LNI-ADFP} &  &  &  & -  & 10.50 & 7.48  & 9.31  & - & 5.67 & -  \\ 
         \hyperlink{Point-SLAM}{\textcolor{magenta}{Point-SLAM}}~\cite{Sandström2023ICCV} &  &  &  & 10.24 & 7.81 & 8.65  & 22.16  & 14.77 & 9.54 & 12.19  \\ 
         \hyperlink{SplaTAM}{\textcolor{magenta}{SplaTAM}}~\cite{keetha2023splatam} &  &  &  & 12.83 & 10.10 & 17.72 & 12.08 & 11.10 & 7.46 & 11.88   \\ 
         \hyperlink{MIPS-Fusion }{\textcolor{magenta}{MIPS-Fusion }}~\cite{tang2023mips}  &  &  & \checkmark &  7.9- & 10.7- & 9.7- & 9.7- & 14.2- & 7.8- & 10.0-  \\
         \hyperlink{Vox-Fusion}{\textcolor{magenta}{Vox-Fusion}}~\cite{yang2022vox} &  &  &  &  8.39 & - & 7.44  & 6.53  & 12.20 & 5.57 & -  \\
         \hyperlink{NEDS-SLAM}{\textcolor{magenta}{NEDS-SLAM}}~\cite{ji2024neds} &  &  &  & 12.34 & - & - & 11.21 & 10.35 & 6.56 & - \\ 
         \hyperlink{NeuV-SLAM}{\textcolor{magenta}{NeuV-SLAM}}~\cite{guo2024neuvslam} &  &  &  & 12.71 & 9.70 & 8.50 & 8.92 & 12.72 & 5.61 & 9.68 \\
         \hyperlink{NICE-SLAM}{\textcolor{magenta}{NICE-SLAM}}~\cite{zhu2022nice}  &   &  &  & 8.64 & 12.25 & 8.09 & 10.28 & 12.93 & 5.59 & 9.63 \\  
         \hyperlink{Co-SLAM}{\textcolor{magenta}{Co-SLAM}}~\cite{Wang_2023_CVPR} &  & \checkmark &  &  7.18 & 12.29 & 9.57 & 6.62 & 13.43 & 7.13 & 9.37 \\ 
        \hyperlink{DG-SLAM}{\textcolor{magenta}{DG-SLAM}} & DROID \cite{droidslam} &  \checkmark & & 7.9- & 11.5- & 8.0- & 8.3- & 7.3- & 8.2- & 8.6- \\
        \hyperlink{VPE-SLAM}{\textcolor{magenta}{VPE-SLAM}}~\cite{vpeslam} & & & & 9.24 & 9.22 & 7.37 & 6.06 & 14.51 & 4.91 & 8.55 \\
          \hyperlink{CG-SLAM}{\textcolor{magenta}{CG-SLAM}}~\cite{hu2024cg} &  &  &  & 7.09 & 7.46 & 8.88 & 8.16 & 11.60 & 5.34 & 8.08 \\ 
        \hyperlink{NIS-SLAM}{\textcolor{magenta}{NIS-SLAM}}~\cite{nis_slam} & & \checkmark & & - & 8.70 & 9.62 & 8.35 & 5.64 & 7.10 & 7.88 \\
          \hyperlink{Loopy-SLAM}{\textcolor{magenta}{Loopy-SLAM}}~\cite{liso2024loopy} &  &  & \checkmark & \textcolor{purple}{\textbf{4.2-}}  &  7.5- &  8.3- & 7.5- & 10.6-  & 7.9- & 7.7- \\ 
         \hyperlink{ESLAM}{\textcolor{magenta}{ESLAM}}~\cite{ESLAM} &  &  &  & 7.3- & 8.5- & 7.5- & 6.5- & 9.0- & 5.7- & 7.4- \\ 
        \hyperlink{IBD-SLAM}{\textcolor{magenta}{IBD-SLAM}}~\cite{cartillier2024slaim} & SuperPoint \cite{detone2018superpoint} \& SuperGlue \cite{sarlin20superglue}  &  & & 6.69 & 9.07 & 7.17 & 6.34  & - & - & - \\

        \hyperlink{ONeK-SLAM}{\textcolor{magenta}{ONeK-SLAM}}~\cite{onekslam} & SIFT matching \cite{lowe2004distinctive} & & & 5.36 & 5.86 & 8.82 & 8.08 & - & 6.76 & - \\

        \hyperlink{LCP-Fusion}{\textcolor{magenta}{LCP-Fusion}}~\cite{wang2024lcp} &  & & & - & 7.56 & - & 5.91 & 10.18 & 6.29 & - \\
         \hyperlink{Structerf-SLAM}{\textcolor{magenta}{Structerf-SLAM}}~\cite{wang2024structerf} & ORB2 \cite{mur2017orb} &  &  & 7.28 & 6.07  & 8.50 & 7.35 & - & 7.28 &  \\  
         \hyperlink{Vox-Fusion++}{\textcolor{magenta}{Vox-Fusion++}}~\cite{zhai2024vox} &  &  & \checkmark &  6.38 & 7.28 & 6.75  & 5.86  & 13.68 & \textcolor{purple}{\textbf{4.73}} & 7.44  \\
         \hyperlink{NGEL-SLAM}{\textcolor{magenta}{NGEL-SLAM}}~\cite{mao2023ngel} & ORB3 \cite{campos2021orb} & \checkmark & \checkmark & 7.23 & 6.98 & 7.95 & 6.12 & 10.14 & 6.27 & 7.44 \\  
         \hyperlink{DNS SLAM}{\textcolor{magenta}{DNS SLAM}}~\cite{li2023dns} &  & \checkmark &  & 5.42 & \textcolor{purple}{\textbf{5.20}} & 9.11 & 7.70 & 10.12 & 4.91 & 7.07  \\ 
         \hyperlink{LRSLAM}{\textcolor{magenta}{LRSLAM}}~\cite{park2024lrslam} & & & & 5.8- & 8.2- & 7.6- & 6.5- & 8.4- & 5.6- & 7.0-\\
         \hyperlink{SNI-SLAM}{\textcolor{magenta}{SNI-SLAM}}~\cite{zhu2023sni}  &  &  &  & 6.90 & 7.38  & 7.19 & \textcolor{purple}{\textbf{4.70}} & - & - & -   \\  
         \hyperlink{GO-SLAM}{\textcolor{magenta}{GO-SLAM}}~\cite{zhang2023go} & DROID \cite{droidslam} & \checkmark & \checkmark & 5.35 & 7.52 & 7.03 & 7.74 & 6.84 & 4.78 & 6.54  \\ 
         \hyperlink{Q-SLAM}{\textcolor{magenta}{Q-SLAM}}~\cite{peng2024q}  & DROID \cite{droidslam} &  &  & 5.23  & 7.63 & 7.02 & 7.66 & 6.52 & - & - \\  

        \hyperlink{SLAIM}{\textcolor{magenta}{SLAIM}}~\cite{cartillier2024slaim} &  &  \checkmark & & 4.56 & 6.12 & 6.9 & 5.82 & 8.88 & 5.69 & 6.32 \\
        
         \hyperlink{MoD-SLAM}{\textcolor{magenta}{MoD-SLAM}}~\cite{zhou2024modslam}  & DROID \cite{droidslam} &  & \checkmark & 5.27  & 7.44 &  6.73 & 6.48 & \textcolor{purple}{\textbf{6.14}} & 5.31 & \textcolor{purple}{\textbf{6.23}}\\   
 
        \Xhline{1pt}
        \rowcolor{maincategories}\multicolumn{11}{c}{RGB} \\
        \Xhline{1pt}

         DROID-SLAM (VO)~\cite{droidslam} & - &  &  & 11.05 & 67.26 & 11.20 & 16.21  & 9.94  & -   & -  \\  
         DROID-SLAM~\cite{droidslam}  & - & \checkmark &  & \textbf{5.48}  & \textbf{9.00} & \textbf{6.76} & \textbf{7.86} & \textbf{7.41}  & -  & - \\ 
         \Xhline{1pt}
         
         \hyperlink{Orbeez-SLAM}{\textcolor{magenta}{Orbeez-SLAM}}~\cite{chung2023orbeez}  & ORB2 \cite{mur2017orb} &   &  & 7.22 & \textcolor{purple}{\textbf{7.15}} & 8.05 & \textcolor{purple}{\textbf{6.58}} & 15.77 & 7.16 & 8.66 \\  
         \hyperlink{Hi-SLAM}{\textcolor{magenta}{Hi-SLAM}}~\cite{zhang2023hi} & DROID \cite{droidslam} & \checkmark & \checkmark & 6.40 & 7.20 &  \textcolor{purple}{\textbf{6.50}} & 8.50 & 7.60 & 8.40 & 7.40  \\  
         \hyperlink{GO-SLAM}{\textcolor{magenta}{GO-SLAM}}~\cite{zhang2023go} & DROID \cite{droidslam} & \checkmark & \checkmark & 5.94 & 8.27  &  8.07  & 8.42 & 8.29  & \textcolor{purple}{\textbf{5.31}} & 7.38  \\ 
         \hyperlink{Q-SLAM}{\textcolor{magenta}{Q-SLAM}}~\cite{peng2024q}  & DROID \cite{droidslam} &  &  & 5.77  & 8.46 & 8.38 & 8.74 & 8.76 & - & - \\  
         \hyperlink{MoD-SLAM}{\textcolor{magenta}{MoD-SLAM}}~\cite{zhou2024modslam}  & DROID \cite{droidslam} &  & \checkmark & \textcolor{purple}{\textbf{5.39}}  & 7.78 & 7.64 & 6.79 & \textcolor{purple}{\textbf{6.58}} & 5.63 & \textcolor{purple}{\textbf{6.64}}  \\  
         \Xhline{2pt}
       \end{tabular}
    }
    \label{tab:scannet}
 \vspace{-1.em}
\end{table}

%% file: chapters/tables/replica_tracking.tex
\begin{table}[t]
    \centering
    \caption{\textbf{Replica \cite{replica19arxiv} Camera Tracking Results}. ATE RMSE [cm] ($\downarrow$) is used as the evaluation metric. }
    \resizebox{0.49\textwidth}{!}{
      \setlength\tabcolsep{5pt}{}
      \renewcommand\arraystretch{1.05}
      \rowcolors{2}{salmon}{white}
      \begin{tabular}{l|ccc|cccccccc|c}
         \Xhline{2pt}

         \multirow{2}{*}{\rowcolors{white}{}{}Method} &  Tracker & Global & Loop & \cellcolor{lower} & \cellcolor{lower} & \cellcolor{lower} & \cellcolor{lower} & \cellcolor{lower} & \cellcolor{lower} & \cellcolor{lower} & \cellcolor{lower} & \cellcolor{lower} \\
         & Based on & BA & Closure & \multirow{-2}{*}{\cellcolor{lower} R0} & \multirow{-2}{*}{\cellcolor{lower} R1} & \multirow{-2}{*}{\cellcolor{lower} R2} & \multirow{-2}{*}{\cellcolor{lower} O0} & \multirow{-2}{*}{\cellcolor{lower} O1} & \multirow{-2}{*}{\cellcolor{lower} O2} & \multirow{-2}{*}{\cellcolor{lower} O3} & \multirow{-2}{*}{\cellcolor{lower} O4} & \multirow{-2}{*}{\cellcolor{lower}  Avg ($\downarrow$)} \\
         
        \Xhline{1pt}
        \rowcolor{maincategories}\multicolumn{13}{c}{RGB-D} \\
        \Xhline{1pt}
         
         \hyperlink{iMAP}{\textcolor{magenta}{iMAP}}~\cite{sucar2021imap} &  &  &  & 3.12 & 2.54 & 2.31 & 1.69 & 1.03 & 3.99 & 4.05 & 1.93 & 2.58 \\
         \hyperlink{NICE-SLAM}{\textcolor{magenta}{NICE-SLAM}}~\cite{zhu2022nice} &   &  &  & 1.69 & 2.04 & 1.55 & 0.99 & 0.90 & 1.39 & 3.97 & 3.08 & 1.95 \\ 
         \hyperlink{ADFP}{\textcolor{magenta}{ADFP}}~\cite{Hu2023LNI-ADFP} &  &  &  & 1.39 & 1.55 & 2.60 & 1.09 & 1.23 & 1.61 & 3.61 & 1.42 & 1.81  \\ 
          \hyperlink{MIPS-Fusion }{\textcolor{magenta}{MIPS-Fusion }}~\cite{tang2023mips}  &  &  & \checkmark & 1.10 & 1.20 & 1.10 & 0.70 & 0.80 & 1.30 &  2.20 & 1.10 & 1.19\\  
          \hyperlink{LCP-Fusion}{\textcolor{magenta}{LCP-Fusion}}~\cite{wang2024lcp} & & & & 0.54 & 1.02 & 0.78 & -&  1.08 & 0.92 & 0.66 & 0.85 & - \\
         \hyperlink{Co-SLAM}{\textcolor{magenta}{Co-SLAM}}~\cite{Wang_2023_CVPR} &  & \checkmark &  & 0.65 & 1.13 & 1.43 & 0.55 & 0.50 & 0.46 & 1.40 & 0.77 &  0.86 \\ 
         \hyperlink{ESLAM}{\textcolor{magenta}{ESLAM}}~\cite{ESLAM} &  &  &  & 0.71 & 0.70  & 0.52 & 0.57 & 0.55 & 0.58 & 0.72 & 0.63 & 0.63 \\ 
         \hyperlink{MonoGS}{\textcolor{magenta}{MonoGS}}~\cite{matsuki2023gaussian} &  &  &  & 0.76 & 0.37 & 0.23 & 0.66 & 0.72 & 0.30 & 0.19 & 1.46 & 0.58 \\
        
         \hyperlink{Vox-Fusion}{\textcolor{magenta}{Vox-Fusion}} ~\cite{yang2022vox} &  &  &  & 0.40 & 0.54 & 0.54 & 0.50 & 0.46 & 0.75 & 0.50 & 0.60 & 0.54 \\ 
         \hyperlink{Point-SLAM}{\textcolor{magenta}{Point-SLAM}}~\cite{Sandström2023ICCV} &  &  &  & 0.61 & 0.41 & 0.37 & 0.38 & 0.48 & 0.54 & 0.72 & 0.63 & 0.52 \\
         \hyperlink{GS-SLAM}{\textcolor{magenta}{GS-SLAM}}~\cite{yan2023gs} &  &   &  & 0.48 & 0.53 & 0.33 & 0.52 & 0.41 & 0.59 & 0.46 & 0.70 & 0.50 \\ 
          \hyperlink{Vox-Fusion++}{\textcolor{magenta}{Vox-Fusion++}}~\cite{zhai2024vox} &  &  & \checkmark & 0.38 & 0.47 & 0.49 & 0.44 & 0.42 & 0.62 & 0.41 & 0.59 & 0.48 \\ 
         \hyperlink{SNI-SLAM}{\textcolor{magenta}{SNI-SLAM}}~\cite{zhu2023sni} &  &  &  & 0.50 & 0.55 & 0.45 & 0.35 & 0.41 & 0.33 & 0.62 & 0.50 & 0.46   \\ 
         \hyperlink{ONeK-SLAM}{\textcolor{magenta}{ONeK-SLAM}}~\cite{onekslam} & SIFT matching \cite{lowe2004distinctive} & & & - & - & - & - & - & - & - & - & 0.46 \\
         \hyperlink{NIS-SLAM}{\textcolor{magenta}{NIS-SLAM}}~\cite{nis_slam} & & \checkmark & & 0.30 & 0.40 & 0.36 & 0.29 & 0.31 & 0.92 & 0.67 & 0.44 & 0.46 \\
         \hyperlink{DNS SLAM}{\textcolor{magenta}{DNS SLAM}}~\cite{li2023dns} &  & \checkmark &  & 0.49 & 0.46 & 0.38 & 0.34 & 0.35 & 0.39 & 0.62 & 0.60 & 0.45   \\ 
         \hyperlink{VPE-SLAM}{\textcolor{magenta}{VPE-SLAM}}~\cite{vpeslam} & & & & 0.32 & 0.26 & 0.42 & 0.42 & 0.41 & 0.39 & 0.47 & 0.33 & 0.38 \\
         \hyperlink{GS3LAM}{\textcolor{magenta}{GS3LAM} \cite{GS3LAM}} & & & & 0.27 & 0.25 & 0.28 & 0.67 & 0.21 & 0.33 & 0.30 & 0.65 & 0.37 \\
         \hyperlink{SplaTAM}{\textcolor{magenta}{SplaTAM}}~\cite{keetha2023splatam} &  &  &  & 0.31 & 0.40 & 0.29 & 0.47 & 0.27 & 0.29 & 0.32 & 0.55 & 0.36 \\ 
         \hyperlink{NEDS-SLAM}{\textcolor{magenta}{NEDS-SLAM}}~\cite{ji2024neds} &  &  &  & - & - & - & - & - & - & - & - & 0.35 \\ 
         \hyperlink{GO-SLAM}{\textcolor{magenta}{GO-SLAM}}~\cite{zhang2023go} & DROID \cite{droidslam} & \checkmark & \checkmark & 0.32 & 0.30 & 0.39 & 0.39 & 0.46 & 0.34 & 0.29 & 0.29 & 0.34 \\ 
         \hyperlink{MoD-SLAM}{\textcolor{magenta}{MoD-SLAM}}~\cite{zhou2024modslam}  & DROID \cite{droidslam} &  & \checkmark & - & - & - & - & - & - & - & - & 0.33\\ 
          
          \hyperlink{Loopy-SLAM}{\textcolor{magenta}{Loopy-SLAM}}~\cite{liso2024loopy}  &  &  & \checkmark & 0.24 & 0.24 & 0.28 & 0.26 & 0.40 & 0.29 & 0.22 & 0.35 & 0.29\\  
          \hyperlink{CG-SLAM}{\textcolor{magenta}{CG-SLAM}}~\cite{hu2024cg}  &  &  &  & 0.29 & 0.27 & 0.25 & 0.33 & 0.14 & 0.28 & 0.31 & 0.29 & 0.27\\  
          \hyperlink{HF-GS SLAM}{\textcolor{magenta}{HF-GS SLAM}}~\cite{sun2024high}  &  &  &  & 0.19 & 0.34 & 0.16 & 0.21 & 0.26 & 0.23 & 0.21 & 0.38 & 0.25\\  
          \hyperlink{GS-ICP SLAM}{\textcolor{magenta}{GS-ICP SLAM}}~\cite{ha2024rgbd}  & G-ICP \cite{segal2009generalized} &  &  & \textcolor{purple}{\textbf{0.15}} & \textcolor{purple}{\textbf{0.16}} & \textcolor{purple}{\textbf{0.11}} & \textcolor{purple}{\textbf{0.18}} & \textcolor{purple}{\textbf{0.12}} & \textcolor{purple}{\textbf{0.17}} & \textcolor{purple}{\textbf{0.16}} & \textcolor{purple}{\textbf{0.21}} & \textcolor{purple}{\textbf{0.16}}\\ 
 
        \Xhline{1pt}
        \rowcolor{maincategories}\multicolumn{13}{c}{RGB} \\
        \Xhline{1pt}

         DROID-SLAM~\cite{droidslam}  & - & \checkmark &  & - & - & - & - & - & - & - & - & 0.42\\ 
         \Xhline{1pt}
         
         \hyperlink{TT-HO-SLAM}{\textcolor{magenta}{TT-HO-SLAM}}~\cite{lin2023ternary} &  &  &  & 4.51 & 0.91 & 7.49 & 0.59 & 1.74 & 1.70 & 0.81 & 3.47 & 2.65 \\  
         \hyperlink{NICER-SLAM}{\textcolor{magenta}{NICER-SLAM}}~\cite{Zhu2023NICER} &  &  &  & 1.36 & 1.60 & 1.14 & 2.12 & 3.23 & 2.12 & 1.42 & 2.01 & 1.88 \\ 
         \hyperlink{DIM-SLAM}{\textcolor{magenta}{DIM-SLAM}}~\cite{li2023dense} &  &  &  & 0.48 & 0.78 & 0.35 & 0.67 & 0.37 & 0.36 & 0.33 & 0.36 & 0.46 \\ 
         
         \hyperlink{GO-SLAM}{\textcolor{magenta}{GO-SLAM}}~\cite{zhang2023go}  & DROID \cite{droidslam} & \checkmark & \checkmark & - & - & - & - & - & - & - & - & 0.39\\  
         
         \hyperlink{MoD-SLAM}{\textcolor{magenta}{MoD-SLAM}}~\cite{zhou2024modslam}  & DROID \cite{droidslam} &  & \checkmark & 0.28 & 0.29 & 0.30 & 0.40 & 0.45 & 0.50 & 0.31 & \textcolor{purple}{\textbf{0.27}} & 0.35\\  
          \hyperlink{MGS-SLAM}{\textcolor{magenta}{MGS-SLAM}}~\cite{zhu2024mgs} & DPVO \cite{teed2022deep} & \checkmark & & 0.36 & 0.35 & 0.32 & 0.35 & 0.28 &  \textcolor{purple}{\textbf{0.26}} & 0.32 & 0.34 & 0.32 \\

         \hyperlink{MonoGS++}{\textcolor{magenta}{MonoGS++}}~\cite{monogspp} & DPVO \cite{teed2022deep} & & &  \textcolor{purple}{\textbf{0.20}} &  \textcolor{purple}{\textbf{0.17}} &  \textcolor{purple}{\textbf{0.22}} &  \textcolor{purple}{\textbf{0.29}} &  \textcolor{purple}{\textbf{0.13}} & 0.42 &  \textcolor{purple}{\textbf{0.20}} & 0.42 &  \textcolor{purple}{\textbf{0.26}} \\
         
         \Xhline{2pt}   

       \end{tabular}
    }   
    \label{tab:replica_tracking}

\end{table}

%% file: chapters/tables/replica_mapping.tex
\begin{table}[t]
    \centering
    \caption{\textbf{Replica \cite{replica19arxiv} Mapping Results.} L1-Depth ($\downarrow$), Acc. [cm] ($\downarrow$), Comp. [cm] ($\downarrow$) and Comp. Ratio [\%] ($\uparrow$) with 5 cm threshold are used as the evaluation metrics. * evaluates on ground truth poses.}
    \resizebox{0.48\textwidth}{!}{
      \setlength\tabcolsep{8pt}{}
      \renewcommand\arraystretch{1.05}
      \rowcolors{2}{salmon}{white}
      \begin{tabular}{l|cccc}
         \Xhline{2pt}
         Method & \cellcolor{lower} L1-Depth $\downarrow$ & \cellcolor{lower} Acc. [cm] $\downarrow$ & \cellcolor{lower} Comp. [cm]$\downarrow$ & \cellcolor{higher}  Comp. Ratio [$\%$]$\uparrow$ \\

        \Xhline{1pt}
        \rowcolor{maincategories}\multicolumn{5}{c}{RGB-D} \\
        \Xhline{1pt}
        
         {COLMAP }\cite{schoenberger2016mvs} & - &  8.69 & 12.12 & 67.62 \\  
         {TSDF}\cite{zeng20173dmatch}  & 7.57 & \textbf{1.60} & \textbf{3.49} & \textbf{86.08} \\  
         \Xhline{1pt} 
         \hyperlink{iMAP}{\textcolor{magenta}{iMAP}}~\cite{sucar2021imap}  & 7.64 &  6.95  & 5.33  & 66.60 \\   
         \hyperlink{GO-SLAM}{\textcolor{magenta}{GO-SLAM}}~\cite{zhang2023go} & 4.68 & 2.50  & 3.74  & 88.09 \\ 
         \hyperlink{NICE-SLAM}{\textcolor{magenta}{NICE-SLAM}}~\cite{zhu2022nice} & 3.53  &  2.85  & 3.00  & 89.33  \\  
         \hyperlink{GO-SLAM}{\textcolor{magenta}{GO-SLAM}}*~\cite{zhang2023go} & 3.38 & 2.50  & 3.74  & 88.09 \\  
          \hyperlink{DNS SLAM}{\textcolor{magenta}{DNS SLAM}}~\cite{li2023dns}  & 3.16 & 2.76 & 2.74 & 91.73 \\ 
         \hyperlink{MoD-SLAM}{\textcolor{magenta}{MoD-SLAM}}~\cite{zhou2024modslam}  & 3.11 & 2.13  & -  & - \\  
         \hyperlink{ADFP}{\textcolor{magenta}{ADFP}}~\cite{Hu2023LNI-ADFP}  & 3.01 & 2.77  & 2.45  & 92.79 \\  
         \hyperlink{NID-SLAM}{\textcolor{magenta}{NID-SLAM}}~\cite{xu2024nid} & 2.87 & 2.72  & 2.56 & 91.16 \\  
         \hyperlink{Vox-Fusion}{\textcolor{magenta}{Vox-Fusion}}~\cite{yang2022vox} & -  & 2.37  & 2.28  & 92.86 \\ 

         \hyperlink{Vox-Fusion++}{\textcolor{magenta}{Vox-Fusion++}}~\cite{zhai2024vox} & -  & 1.44  & 2.43  & 92.37 \\  
          
         \hyperlink{Q-SLAM}{\textcolor{magenta}{Q-SLAM}}~\cite{liso2024loopy} & 1.87 & -  & -  & - \\

         \hyperlink{CG-SLAM}{\textcolor{magenta}{CG-SLAM}}~\cite{hu2024cg} & - & \textcolor{purple}{\textbf{1.01}}  & 2.84  & 88.51 \\  

         \hyperlink{VPE-SLAM}{\textcolor{magenta}{VPE-SLAM}}~\cite{vpeslam} & 1.52 & 2.14 & - & \textcolor{purple}{\textbf{93.61}} \\
          \hyperlink{Co-SLAM }{\textcolor{magenta}{Co-SLAM }}~\cite{Wang_2023_CVPR}  & 1.51  & -  & -  & - \\ 
         \hyperlink{HERO-SLAM}{\textcolor{magenta}{HERO-SLAM}}~\cite{heroslam} & 1.41 & 2.62 & \textcolor{purple}{\textbf{2.15}} & 93.22 \\ 
         \hyperlink{NGEL-SLAM}{\textcolor{magenta}{NGEL-SLAM}}~\cite{mao2023ngel}  & 1.28  & -  & -  & - \\ 
         \hyperlink{ESLAM}{\textcolor{magenta}{ESLAM}}~\cite{ESLAM} & 1.18  & -  & -  & - \\ 
         \hyperlink{SplaTAM}{\textcolor{magenta}{SplaTAM}}~\cite{keetha2023splatam} & 0.72  & -  & -  & - \\ 
        \hyperlink{HF-GS SLAM}{\textcolor{magenta}{HF-GS SLAM}}~\cite{sun2024high} & 0.52  & -  & -  & - \\
        \hyperlink{NEDS-SLAM}{\textcolor{magenta}{NEDS-SLAM}}~\cite{ji2024neds} & 0.47  & -  & -  & - \\ 
         \hyperlink{Point-SLAM}{\textcolor{magenta}{Point-SLAM}}~\cite{Sandström2023ICCV} & 0.44 & 1.41  & 3.10  & 88.89 \\  
        \hyperlink{NIS-SLAM}{\textcolor{magenta}{NIS-SLAM}}~\cite{nis_slam} & - & 1.48 & 2.44 & 92.49 \\
        \hyperlink{Loopy-SLAM}{\textcolor{magenta}{Loopy-SLAM}}~\cite{liso2024loopy} & \textcolor{purple}{\textbf{0.35}} & -  & -  & - \\  
        
        \Xhline{1pt}
        \rowcolor{maincategories}\multicolumn{5}{c}{RGB} \\
        \Xhline{1pt}

         \hyperlink{MGS-SLAM}{\textcolor{magenta}{MGS-SLAM}}~\cite{zhu2024mgs} & 7.77 & 7.51 & 3.64 & \textcolor{purple}{\textbf{82.71}} \\
         \hyperlink{NeRF-SLAM}{\textcolor{magenta}{NeRF-SLAM}}~\cite{rosinol2023nerf}  & 4.49 & - & - & - \\ 
         
         \hyperlink{DIM-SLAM}{\textcolor{magenta}{DIM-SLAM}}~\cite{li2023dense}  & - & 4.03  & 4.20  & 79.60 \\  
         \hyperlink{GO-SLAM}{\textcolor{magenta}{GO-SLAM}}~\cite{zhang2023go}  & 4.39 & 3.81  & 4.79  & 78.00 \\ 
         \hyperlink{NICER-SLAM}{\textcolor{magenta}{NICER-SLAM}}~\cite{Zhu2023NICER}  & - & 3.65  & \textcolor{purple}{\textbf{4.16}}  & 79.37 \\ 
         \hyperlink{Hi-SLAM}{\textcolor{magenta}{Hi-SLAM}}~\cite{zhang2023hi}  & 3.63 & 3.62  & 4.59  & 80.60 \\  
         \hyperlink{MoD-SLAM}{\textcolor{magenta}{MoD-SLAM}}~\cite{zhou2024modslam}  & 3.23 & \textcolor{purple}{\textbf{2.48}}  & -  & - \\  
         
         \hyperlink{Q-SLAM}{\textcolor{magenta}{Q-SLAM}}~\cite{liso2024loopy} & \textcolor{purple}{\textbf{2.76}} & -  & -  & - \\ 
         \Xhline{2pt}
       \end{tabular}
    }
    \label{tab:replica_mapping}
\end{table}

%% file: chapters/tables/replica_rendering.tex
\begin{table}[t]
    \centering
    \caption{\textbf{Replica \cite{replica19arxiv} Train View Rendering Results}. We report the PSNR $\uparrow$ as main error metric. }
    \resizebox{0.48\textwidth}{!}{
      \setlength\tabcolsep{3pt}{}
      \renewcommand\arraystretch{1.05}
      \begin{tabular}{l|cccccccc|c}
         \Xhline{2pt}
         Method  & R0 & R1 & R2 & O0 & O1 & O2 & O3 & O4 & Avg \\
        \Xhline{1pt}
        \rowcolor{maincategories}\multicolumn{10}{c}{RGB-D} \\
        \Xhline{1pt}

         \rowcolor{salmon} {\hyperlink{Vox-Fusion}{\textcolor{magenta}{Vox-Fusion}}~\cite{yang2022vox}\!\!\!}  & 22.39 & 22.36 & 23.92 & 27.79 & 29.83 & 20.33 & 23.47 & 25.21 & 24.41 \\

         \hyperlink{NICE-SLAM}{\textcolor{magenta}{NICE-SLAM}}~\cite{zhu2022nice}\!\!\! & 22.12 & 22.47 & 24.52 & 29.07 & 30.34 & 19.66 & 22.23 & 24.94 & 24.42 \\
         
         \rowcolor{salmon} {\hyperlink{GO-SLAM}{\textcolor{magenta}{GO-SLAM}}~\cite{zhang2023go}\!\!\!}  & - & -  & - & - & - & - & - & - & 27.38 \\

         {\hyperlink{ESLAM}{\textcolor{magenta}{ESLAM}}~\cite{ESLAM}\!\!\!} & - & -  & - & - & - & - & - & - & 27.80 \\

         \rowcolor{salmon}\hyperlink{MoD-SLAM}{\textcolor{magenta}{MoD-SLAM}}~\cite{zhou2024modslam}  & - & -  & - & - & - & - & - & - & 29.95 \\

         {\hyperlink{SplaTAM}{\textcolor{magenta}{SplaTAM}}~\cite{keetha2023splatam}\!\!\!}  & 32.86 & 33.89 & 35.25 & 38.26 & 39.17 & 31.97 & 29.70 & 31.81 & 34.11 \\

         \rowcolor{salmon}\hyperlink{GS-SLAM}{\textcolor{magenta}{GS-SLAM}}~\cite{yan2023gs}\!\!\!   & 31.56 & 32.86 & 32.59 & 38.70 & 41.17 & 32.36 & 32.03 & 32.92 & 34.27 \\

         \rowcolor{salmon}\hyperlink{NEDS-SLAM}{\textcolor{magenta}{NEDS-SLAM}}~\cite{ji2024neds}\!\!\! & 35.23 & 34.86 & 35.16 & 37.53 & 39.71 & 32.68 & 31.07 & 31.82 & 34.76 \\

         \rowcolor{salmon}\hyperlink{Point-SLAM}{\textcolor{magenta}{Point-SLAM}}~\cite{Sandström2023ICCV}\!\!\!  & 32.40 & 34.08 & 35.50 & 38.26 & 39.16 & 33.99 & 33.48 & 33.49 & 35.17 \\

        \hyperlink{Q-SLAM}{\textcolor{magenta}{Q-SLAM}}~\cite{peng2024q}\!\!\! & 33.24 & 34.81 & 34.16 & 39.32 & 39.51 & 34.08 & 32.65 & 34.93 & 35.34 \\

         \rowcolor{salmon} \hyperlink{Loopy-SLAM}{\textcolor{magenta}{Loopy-SLAM}}~\cite{liso2024loopy}\!\!\!  & - & -  & - & - & - & - & - & - & 35.47 \\

         \hyperlink{NIDS-SLAM}{\textcolor{magenta}{NIDS-SLAM }~\cite{haghighi2023neural}\!\!\!}  & 33.16 & 35.18 & 36.49 & 40.22 & 38.90 & 34.22 & 34.74 & 33.24 & 35.76 \\

         \rowcolor{salmon} {\hyperlink{HF-GS SLAM}{\textcolor{magenta}{HF-GS SLAM}}~\cite{sun2024high}\!\!\!}  & 33.06 & 35.74 & 37.21 & 41.12 & 41.11 & 33.56 & 33.21 & 34.48 & 36.19 \\

        \rowcolor{salmon} \hyperlink{GS3LAM}{{\textcolor{magenta}{GS3LAM}} \!\!\!}~\cite{GS3LAM} & 33.67 & 35.80 & 35.96 & 40.28 & 41.21 & 34.30 & 34.27 & 34.59 & 36.26 \\

         {\hyperlink{MonoGS}{\textcolor{magenta}{MonoGS}}~\cite{matsuki2023gaussian}\!\!\!}  & 34.83 & 36.43 & 37.49 & 39.95 & 42.09 & 36.24 & 36.70 & 36.07 & 37.50 \\

         \rowcolor{salmon} \hyperlink{GS-ICP SLAM}{{\textcolor{magenta}{GS-ICP SLAM}} \!\!\!}~\cite{ha2024rgbd} & \textcolor{purple}{\textbf{35.37}} & \textcolor{purple}{\textbf{37.80}} & \textcolor{purple}{\textbf{38.50}} & \textcolor{purple}{\textbf{43.13}} & \textcolor{purple}{\textbf{43.26}} & \textcolor{purple}{\textbf{36.93}} & \textcolor{purple}{\textbf{36.90}} & \textcolor{purple}{\textbf{38.75}} & \textcolor{purple}{\textbf{38.83}} \\

        \Xhline{1pt}
        \rowcolor{maincategories}\multicolumn{10}{c}{RGB} \\
        \Xhline{1pt}

         \rowcolor{salmon} {\hyperlink{GO-SLAM}{\textcolor{magenta}{GO-SLAM}}~\cite{zhang2023go}\!\!\!}  & - & -  & - & - & - & - & - & - & 22.13 \\
         
         \hyperlink{NICER-SLAM}{\textcolor{magenta}{NICER-SLAM}}~\cite{Zhu2023NICER}\!\!\! & 25.33 & 23.92  & 26.12 & 28.54 & 25.86 & 21.95 & 26.13 & 25.47 & 25.41 \\

         \rowcolor{salmon} {\hyperlink{MoD-SLAM}{\textcolor{magenta}{MoD-SLAM}}~\cite{zhou2024modslam}} & - & -  & - & - & - & - & - & - & 27.31 \\

         \rowcolor{salmon}{\hyperlink{MGS-SLAM}{\textcolor{magenta}{MGS-SLAM}} \cite{zhu2024mgs}\!\!\!} & 29.91 & 31.06 & 31.49 & 35.51 & 34.25 & 30.83 & 31.86 & 34.38 & 32.41 \\
         \rowcolor{salmon} {\hyperlink{Q-SLAM}{\textcolor{magenta}{Q-SLAM}}~\cite{peng2024q}\!\!\!} & 29.58 & 32.74 & 31.25 & 36.31 & 37.22 & 30.68 & 30.21 & 31.96 & 32.49 \\

         {\hyperlink{Photo-SLAM}{\textcolor{magenta}{Photo-SLAM}}~\cite{huang2023photo}\!\!\!} & - & -  & - & - & - & - & - & - & 33.30 \\
          
         
        {\hyperlink{MonoGS++}{\textcolor{magenta}{MonoGS++}} \cite{monogspp}\!\!\!}  & \textcolor{purple}{\textbf{33.75}} & \textcolor{purple}{\textbf{36.47}} & \textcolor{purple}{\textbf{37.01}} & \textcolor{purple}{\textbf{42.31}} & \textcolor{purple}{\textbf{43.05}} & \textcolor{purple}{\textbf{36.11}} & \textcolor{purple}{\textbf{36.34}} & \textcolor{purple}{\textbf{37.28}} & \textcolor{purple}{\textbf{37.79}} \\

        \Xhline{2pt}  
         
       \end{tabular}
    }
        \label{tab:replica_rendering}
\end{table}

%% file: chapters/tables/replica_semantic.tex
\begin{table}[t]
    \centering
    \caption{\textbf{Replica \cite{replica19arxiv} Semantic Results}.  Quantitative comparison of input views semantic segmentation performance on the Replica dataset \cite{replica19arxiv} using the mIoU metric.}
    \resizebox{0.49\textwidth}{!}{
      \setlength\tabcolsep{5.3pt}{}
      \renewcommand\arraystretch{1.05}
      \rowcolors{2}{salmon}{white}
      \begin{tabular}{l|c|cccccccc}
         \Xhline{2pt}

         \multirow{2}{*}{\rowcolors{white}{}{}Method} &  External & \cellcolor{lower} & \cellcolor{lower} & \cellcolor{lower} & \cellcolor{lower} & \cellcolor{lower} & \cellcolor{lower} & \cellcolor{lower} & \cellcolor{lower}\\
         & Priors & \multirow{-2}{*}{\cellcolor{lower} R0} & \multirow{-2}{*}{\cellcolor{lower} R1} & \multirow{-2}{*}{\cellcolor{lower} R2} & \multirow{-2}{*}{\cellcolor{lower} O0} & \multirow{-2}{*}{\cellcolor{lower} O1} & \multirow{-2}{*}{\cellcolor{lower} O2} & \multirow{-2}{*}{\cellcolor{lower} O3} & \multirow{-2}{*}{\cellcolor{lower} O4}\\
         
        \Xhline{1pt}
        \rowcolor{maincategories}\multicolumn{10}{c}{RGB-D} \\
        \Xhline{1pt}

         \hyperlink{DNS-SLAM}{\textcolor{magenta}{DNS-SLAM}}~\cite{li2023dns} & GT semantics & 88.32 & 84.90 & 81.20  & 84.66 & – & – & – & – \\ 
         \hyperlink{SNI-SLAM}{\textcolor{magenta}{SNI-SLAM}}~\cite{zhu2023sni} & Dinov2 \cite{oquab2023dinov2} & 88.42 & 87.43 & 86.16 & 87.63 & 78.63 & 86.49 & 74.01 & 80.22\\ 
         \hyperlink{NEDS-SLAM}{\textcolor{magenta}{NEDS-SLAM}}~\cite{ji2024neds} & DINO \cite{oquab2023dinov2} & 90.73 & 91.20 & - & 90.42 & - & - & - & - \\ 
         \hyperlink{SemGauss-SLAM}{\textcolor{magenta}{SemGauss-SLAM}}~\cite{zhu2024semgauss} & Dinov2 \cite{oquab2023dinov2}& 92.81 & 94.10 & 94.72 & 95.23 & \textcolor{purple}{\textbf{90.11}} & \textcolor{purple}{\textbf{94.93}} & \textcolor{purple}{\textbf{92.93}} & \textcolor{purple}{\textbf{94.82}} \\ 

         \hyperlink{GS3LAM}{\textcolor{magenta}{GS3LAM}}~\cite{GS3LAM} & GT semantics & \textcolor{purple}{\textbf{96.83}} & \textcolor{purple}{\textbf{96.68}} & \textcolor{purple}{\textbf{96.40}} & \textcolor{purple}{\textbf{96.61}} & - & - & - & - \\

         \Xhline{2pt}  
       \end{tabular}
    }   
    \label{tab:replica_semantic}

\end{table}

%% file: chapters/tables/kitti_lidar.tex
\begin{table}[t]
    \centering
    \caption{\textbf{KITTI \cite{Geiger2012CVPR} LiDAR Odometry/SLAM Results.} $\dagger$ indicates sequences with loops and Avg.$\dagger$ denotes the average metric for such sequences.}
    \setlength\tabcolsep{2pt}
    \resizebox{0.48\textwidth}{!}{
        \rowcolors{2}{salmon}{white}
        \begin{tabular}{l|ccccccccccc|cc}
            \Xhline{2pt}
            Method & \cellcolor{lower} 00 & \cellcolor{lower} 01 & \cellcolor{lower} 02 & \cellcolor{lower} 03 & \cellcolor{lower} 04 & \cellcolor{lower} 05 & \cellcolor{lower} 06 & \cellcolor{lower} 07 & \cellcolor{lower} 08 & \cellcolor{lower} 09 & \cellcolor{lower} 10 & \cellcolor{lower} \textbf{Avg.} & \cellcolor{lower} 11-21 \\

            \Xhline{1pt}
            \rowcolor{maincategories} \multicolumn{14}{c}{LiDAR Odometry Evaluation} \\
            \Xhline{1pt}
            MULLS \cite{pan2021mulls} & \textbf{0.56} & \textbf{0.64} & 0.55 & \textcolor{purple}{\textbf{0.71}} & 0.41 & 0.30 & 0.30 & 0.38 & \textcolor{purple}{\textbf{0.78}} & \textcolor{purple}{\textbf{0.48}} & 0.59 & 0.52 & 0.65\\
            CT-ICP \cite{dellenbach2022ct} & \textcolor{purple}{\textbf{0.49}} & 0.76 & \textcolor{purple}{\textbf{0.52}} & 0.72 & 0.39 & \textcolor{purple}{\textbf{0.25}} & \textcolor{purple}{\textbf{0.27}} & \textcolor{purple}{\textbf{0.31}}  & 0.81 & 0.49 & \textbf{0.48} & \textcolor{purple}{\textbf{0.50}} & \textcolor{purple}{\textbf{0.59}} \\

            SuMa-LO \cite{behley2018efficient} & 0.72 & 1.71 & 1.06 & 0.66 & \textbf{0.38} & 0.50 & 0.41 & 0.55 & 1.02 & \textcolor{purple}{\textbf{0.48}} & 0.71 & 0.75 & 1.39 \\

            Litamin-LO \cite{yokozuka2021litamin2} & 0.78 & 2.10 & 0.95 & 0.96 & 1.05 & 0.55 & 0.55 & 0.48 & 1.01 & 0.69 & 0.80 & 0.88 & - \\
            \Xhline{1pt}
            \hyperlink{Nerf-LOAM}{\textcolor{magenta}{Nerf-LOAM}}~\cite{deng2023nerf} & 1.34 & 2.07 & - & 2.22 & 1.74 & 1.40 & - & 1.00 & - & 1.63 & 2.08 & 1.69 & - \\
            \hyperlink{PIN-SLAM}{\textcolor{magenta}{PIN-LO}}~\cite{pan2024pin} & \textbf{0.55} & \textcolor{purple}{\textbf{0.54}} & \textcolor{purple}{\textbf{0.52}} & \textbf{0.74} & \textcolor{purple}{\textbf{0.28}} & \textbf{0.29} & \textbf{0.32} & \textbf{0.36} & \textbf{0.83} & \textbf{0.56} & \textcolor{purple}{\textbf{0.47}} & \textcolor{purple}{\textbf{0.50}}  \\
            \Xhline{1pt}
             \cellcolor{white} & \cellcolor{lower} 00$\dagger$ & \cellcolor{lower} 01 & \cellcolor{lower} 02$\dagger$ & \cellcolor{lower} 03 & \cellcolor{lower} 04 & \cellcolor{lower} 05$\dagger$ & \cellcolor{lower} 06$\dagger$ & \cellcolor{lower} 07$\dagger$ & \cellcolor{lower} 08$\dagger$ & \cellcolor{lower} 09$\dagger$ & \cellcolor{lower} 10 & \cellcolor{lower} \textbf{Avg.}$\dagger$ & \cellcolor{lower} Avg. \\
            \Xhline{1pt}
            \rowcolor{maincategories} \multicolumn{14}{c}{LiDAR SLAM Evaluation} \\
            \Xhline{1pt}
            MULLS \cite{pan2021mulls} & 1.1 & \textcolor{purple}{\textbf{1.9}} & 5.4 & 0.7 & 0.9 & 1.0 & 0.3 & 0.4 & 2.9 & 2.1 & 1.1 & 1.9 & 1.6 \\
            SuMa \cite{behley2018efficient} & 1.0 & 13.8 & 7.1 & 0.9 & \textbf{0.4} & 0.6 & 0.6 & 1.0 & 3.4 & \textbf{1.1} & 1.3 & 2.1  & 3.2\\
            Litamin2 \cite{yokozuka2021litamin2} & 1.3 & 15.9 & \textcolor{purple}{\textbf{3.2}} & 0.8 & 0.7 & 0.6 & 0.8 & 0.5 & \textbf{2.1} & 2.1 & \textbf{1.0} & \textbf{1.5} & 2.4\\
            HLBA \cite{liu2023large} & \textcolor{purple}{\textbf{0.8}} & \textcolor{purple}{\textbf{1.9}} & 5.1 & \textcolor{purple}{\textbf{0.6}} & 0.8 & \textbf{0.4} & \textcolor{purple}{\textbf{0.2}} & \textcolor{purple}{\textbf{0.3}} & 2.7 & 1.3 & 1.1 & \textbf{1.5} & \textbf{1.4}\\
            \Xhline{1pt}
            \hyperlink{PIN-LO}{\textcolor{magenta}{PIN-LO}}~\cite{pan2024pin} & 4.3 & \textbf{2.0} & 7.3 & \textbf{0.7} & \textcolor{purple}{\textbf{0.1}} & 2.1 & 0.7 & 0.4 & 3.5 & 1.8 & \textcolor{purple}{\textbf{0.6}} & 2.9 & 2.1\\
            \hyperlink{PIN-SLAM}{\textcolor{magenta}{PIN-SLAM}}~\cite{pan2024pin} & \textcolor{purple}{\textbf{0.8}} & \textbf{2.0} & \textbf{3.3} & \textbf{0.7} & \textcolor{purple}{\textbf{0.1}} & \textcolor{purple}{\textbf{0.2}} & \textbf{0.4} & \textcolor{purple}{\textbf{0.3}} & \textcolor{purple}{\textbf{1.7}} & \textcolor{purple}{\textbf{1.0}} & \textcolor{purple}{\textbf{0.6}} & \textcolor{purple}{\textbf{1.1}} & \textcolor{purple}{\textbf{1.0}}\\
            \Xhline{2pt}
        \end{tabular}
    }
    \label{tab:lidar_kitti}
\end{table}

%% file: chapters/tables/college_tracking.tex
\begin{table}[t]
  \centering
    \caption{\textbf{Newer College \cite{ramezani2020newer} Camera Tracking Results}. ATE RMSE [cm] ($\downarrow$) is used as the evaluation metric.}
  \setlength\tabcolsep{2pt}
  \rowcolors{2}{salmon}{white}
  \resizebox{0.48\textwidth}{!}{
    \begin{tabular}{l|ccccccc|c}
      \Xhline{2pt}
      Method & \cellcolor{lower} 01 & \cellcolor{lower} 02 & \cellcolor{lower} quad & \cellcolor{lower} math\textunderscore{}e & \cellcolor{lower} ug\textunderscore{}e & \cellcolor{lower} cloister\textunderscore{}e & \cellcolor{lower} stairs & \cellcolor{lower} Avg. \\
      \Xhline{1pt}
      MULLS \cite{pan2021mulls}  & 2.51 & 8.39 & \textbf{0.12} & 0.35 & 0.86 & 0.41 & - & 2.11 \\
      SuMa \cite{behley2018efficient}  & \textbf{2.03} & \textbf{3.65} & 0.28 & \textbf{0.16} & \textbf{0.09} & \textbf{0.20} & \textbf{1.85} & \textbf{1.18} \\

      \Xhline{1pt}
      \hyperlink{PIN-SLAM}{\textcolor{magenta}{PIN-LO}}~\cite{pan2024pin} & 2.21 & 4.93 & \textcolor{purple}{\textbf{0.09}} & 0.10 & \textcolor{purple}{\textbf{0.07}} & 0.41 & \textcolor{purple}{\textbf{0.06}} & 1.12 \\
      \hyperlink{PIN-SLAM}{\textcolor{magenta}{PIN-SLAM}}~\cite{pan2024pin}  & \textcolor{purple}{\textbf{0.43}} & \textcolor{purple}{\textbf{0.30}} & \textcolor{purple}{\textbf{0.09}} & \textcolor{purple}{\textbf{0.09}} & \textcolor{purple}{\textbf{0.07}} & \textcolor{purple}{\textbf{0.18}} & \textcolor{purple}{\textbf{0.06}} & \textcolor{purple}{\textbf{0.19}} \\
      \Xhline{2pt}
    \end{tabular}
  }
  \label{tab:college_tracking}
\end{table}

%% file: chapters/tables/college_mapping.tex
\begin{table}[!t]
  \centering
  \caption{\textbf{Newer College \cite{ramezani2020newer} Mapping Results.} Acc. [cm] ($\downarrow$) and Comp. [cm] ($\downarrow$) are used as the evaluation metrics.}
   \setlength\tabcolsep{2pt}
     \resizebox{0.48\textwidth}{!}{
     \rowcolors{2}{salmon}{white}
     \begin{tabular}{l|cc|cc}
        \Xhline{2pt}
        \multirow{2}{*}{\rowcolors{white}{}{}Method} & \multicolumn{2}{c|}{\cellcolor{lower} \text{Quad}}& \multicolumn{2}{c}{\cellcolor{lower} \text{Math Institute}}\\
        & \cellcolor{lower} Acc. [cm] $\downarrow$& \cellcolor{lower} Comp. [cm] $\downarrow$ & \cellcolor{lower}  Acc. [cm] $\downarrow$ &  \cellcolor{lower} Comp. [cm] $\downarrow$\\
        \Xhline{1pt}
        SLAMesh~\cite{ruan2023slamesh} & 19.21 & 48.83 & \textcolor{purple}{\textbf{12.80}} & 23.50 \\
        \Xhline{1pt}
        \hyperlink{Nerf-LOAM}{\textcolor{magenta}{Nerf-LOAM}}~\cite{deng2023nerf} & 12.89 & 22.21 & - & - \\
        \hyperlink{PIN-SLAM}{\textcolor{magenta}{PIN-SLAM}}~\cite{pan2024pin} & \textcolor{purple}{\textbf{11.55}} & \textcolor{purple}{\textbf{15.25}} & \textbf{13.70} & \textcolor{purple}{\textbf{21.91}}\\
        \Xhline{2pt}
     \end{tabular}
     }
  \label{tab:recon_experiments_on_ncd}
\end{table}

%% file: chapters/tables/performance.tex
\begin{table}[t]
\centering
\caption{\textbf{Performance Evaluation:} GPU memory requirements (GB) and average FPS efficiency on Replica room0 (RGB/RGB-D) and KITTI 00 sequence (LiDAR).} 
\setlength\tabcolsep{4pt}
\resizebox{0.5\textwidth}{!}{
    \rowcolors{2}{salmon}{white}
    \begin{tabular}{l|c|cc}
    \Xhline{2pt}
    Method  & Scene Encoding & \cellcolor{lower} GPU Mem. [G] $\downarrow$ & \cellcolor{higher} Avg. FPS $\uparrow$ \\
    \Xhline{1pt}
    \rowcolor{maincategories}\multicolumn{4}{c}{RGB-D} \\
    \Xhline{1pt}
    \hyperlink{iMAP}{\textcolor{magenta}{iMAP}}~\cite{sucar2021imap}  & MLP & 6.44 & 0.13 \\
    \hyperlink{SplaTAM}{\textcolor{magenta}{SplaTAM}}~\cite{keetha2023splatam}  & 3D Gaussians & 18.54 & 0.14 \\
    
    \hyperlink{Point-SLAM}{\textcolor{magenta}{Point-SLAM}}~\cite{Sandström2023ICCV}  & Neural Points + MLP & 7.11 & 0.23 \\
    
    \hyperlink{UncLe-SLAM}{\textcolor{magenta}{UncLe-SLAM}}~\cite{uncleslam2023}  & Hier. Grid + MLP & 8.24 & 0.24 \\
    \hyperlink{NICE-SLAM}{\textcolor{magenta}{NICE-SLAM}}~\cite{zhu2022nice} & Hier. Grid + MLP & 4.70 & 0.61 \\
    \hyperlink{ADFP}{\textcolor{magenta}{ADFP}}~\cite{Hu2023LNI-ADFP}  & Hier. Grid + MLP & 3.76 & 0.74 \\
    \hyperlink{Vox-Fusion}{\textcolor{magenta}{Vox-Fusion}} ~\cite{yang2022vox} & Sparse Voxels + MLP & 21.22 & 0.74 \\
    \hyperlink{Plenoxel-SLAM}{\textcolor{magenta}{Plenoxel-SLAM}}~\cite{teigen2024rgb}  & Plenoxels & 13.04 & 1.25 \\
    \hyperlink{ESLAM}{\textcolor{magenta}{ESLAM}}~\cite{ESLAM}  & Feature Planes + MLP & 13.04 & 4.62 \\
    \hyperlink{Co-SLAM}{\textcolor{magenta}{Co-SLAM}}~\cite{Wang_2023_CVPR}  & Hash Grid + MLP & \textcolor{purple}{\textbf{3.56}} & 7.97 \\
    \hyperlink{GO-SLAM}{\textcolor{magenta}{GO-SLAM}}~\cite{zhang2023go}  & Hash Grid + MLP & 18.50 & \textcolor{purple}{\textbf{8.36}} \\

    \Xhline{1pt}
    \rowcolor{maincategories}\multicolumn{4}{c}{RGB} \\
    \Xhline{1pt}
    
    \hyperlink{DIM-SLAM}{\textcolor{magenta}{DIM-SLAM}}~\cite{li2023dense} & Hier. Grid + MLP & \textcolor{purple}{\textbf{4.78}} & 3.14 \\
    \hyperlink{Orbeez-SLAM}{\textcolor{magenta}{Orbeez-SLAM}}~\cite{chung2023orbeez}  & Voxels + MLP & 7.55 & 17.70 \\
    \hyperlink{NeRF-SLAM}{\textcolor{magenta}{NeRF-SLAM}}~\cite{rosinol2023nerf}  & Hash Grid + MLP & 9.38 & \textcolor{purple}{\textbf{20.00}} \\

    \Xhline{1pt}
    \rowcolor{maincategories}\multicolumn{4}{c}{LiDAR} \\
    \Xhline{1pt}

    \hyperlink{Nerf-LOAM}{\textcolor{magenta}{Nerf-LOAM}}~\cite{deng2023nerf} & Sparse Voxel + MLP & 11.58 & 0.24 \\
    \hyperlink{PIN-SLAM}{\textcolor{magenta}{PIN-SLAM}}~\cite{pan2024pin} & Neural Points + MLP & \textcolor{purple}{\textbf{6.93}} & \textcolor{purple}{\textbf{6.67}} \\
    
    \Xhline{2pt}
    \end{tabular}
}
\label{tab:performance_analysis}
\end{table}

%% file: chapters/discussion.tex
\section{Discussion}
\label{sec:discussion}
In this section we focus on highlighting the key findings of the survey. We will outline the main advances achieved through the most recent methodologies examined, while identifying ongoing challenges and potential avenues for future research in this area.

\textbf{Scene Representation.} The choice of scene representation is critical in current SLAM solutions, significantly affecting mapping/tracking accuracy, rendering quality, and computation. Early approaches, such as iMAP \cite{sucar2021imap}, used network-based methods, implicitly modeling scenes with coordinate-based MLP(s). While these provide compact, continuous modeling of the scene, they struggle with real-time reconstruction due to challenges in updating local regions and scaling for large scenes. In addition, they tend to produce over-smoothed scene reconstructions. Subsequent research has explored grid-based representations, such as multi-resolution hierarchical \cite{zhu2022nice,Wang_2023_CVPR} and sparse octree grids \cite{yang2022vox, liu2023efficient}, which have gained popularity. Grids allow for fast neighbor lookups, but require a pre-specified grid resolution, resulting in inefficient memory use in empty space and a limited ability to capture fine details constrained by the resolution. Recent advances, such as Point-SLAM \cite{Sandström2023ICCV} and Loopy-SLAM \cite{liso2024loopy}, favor hybrid neural point-based representations. Unlike grids, point densities vary naturally and need not be pre-specified. Points concentrate efficiently around surfaces while assigning higher density to details, facilitating scalability and local updates compared to network-based methods. At present, point-based methods have demonstrated superior performance in 3D reconstruction, yielding highly accurate 3D surfaces, as evidenced by experiments conducted on the Replica dataset. However, similar to other NeRF-style approaches, volumetric ray sampling significantly restricts its efficiency. 

Promising techniques include explicit representations based on the 3D Gaussian Splatting paradigm. Explicit representations based on 3DGS have been shown to achieve state-of-the-art rendering accuracy compared to other representations while also exhibiting faster rendering. However, these methods have several limitations, including a heavy reliance on initialization and a lack of control over primitive growth in unobserved regions. Furthermore, the original 3DGS-based scene representation requires a large number of 3D Gaussian primitives to achieve high-fidelity reconstruction, resulting in substantial memory consumption.

Despite significant progress over the past three years, ongoing research is still actively engaged in overcoming existing scene representation limitations and finding ever more effective alternatives to improve accuracy and real-time performance in SLAM.

\textbf{Catastrophic Forgetting.} Existing methods often exhibit a tendency to forget previously learned information, particularly in large scenarios or extended video sequences. In the case of network-based methods, this is attributed to their reliance on single neural networks or global models with fixed capacity, which are affected by global changes during optimization. One common approach to alleviate this problem is to train the network using sparse ray sampling with current observations while replaying keyframes from historical data. However, in large-scale incremental mapping, such a strategy results in a cumulative increase in data, requiring complex resampling procedures for memory efficiency.
The forgetting problem extends to grid-based approaches. Despite efforts to address this issue, obstacles arise due to quadratic or cubic spatial complexity, which poses scalability challenges.
Similarly, while explicit representations, such as 3DGS-style solutions, offer a practical workaround for catastrophic forgetting, they face challenges due to increased memory requirements and slow processing, especially in large scenes. Some methods attempt to mitigate these limitations by employing sparse frame sampling, but this leads to inefficient information sampling across 3D space, resulting in slower and less uniform model updates compared to approaches that integrate sparse ray sampling.

Eventually, some strategies recommend dividing the environment into submaps and assigning local SLAM tasks to different agents.  However, this introduces additional challenges in handling multiple distributed models and devising efficient strategies to manage overlapping regions while preventing the occurrence of map fusion artifacts.

\textbf{Real-Time Constraints}. Many of the techniques reviewed face challenges in achieving real-time processing, often failing to match the sensor frame rate. This limitation is mainly due to the chosen map data structure or the computationally intensive ray-wise rendering-based optimization, which is especially noticeable in NeRF-style SLAM methods. In particular, hybrid approaches using hierarchical grids require less GPU memory but exhibit slower runtime performance. On the other hand, advanced representations such as hash grids or sparse voxels allow for faster computation, but with higher memory requirements. Finally, despite their advantages in fast image rendering, current 3DGS-style methods still struggle to efficiently handle simultaneous tracking and mapping processing, preventing their effective use in real-time applications.

\textbf{Global Optimization.} Implementing LC and global BA requires significant computational resources, risking performance bottlenecks, especially in real-time applications. Many reviewed frame-to-model methods (\eg, iMap~\cite{sucar2021imap}, NICE-SLAM~\cite{zhu2022nice}, etc.) face challenges with loop closure and global bundle adjustment due to the prohibitive computational complexity of updating the entire 3D model. In contrast, frame-to-frame techniques (\eg, GO-SLAM~\cite{zhang2023go}, etc.) facilitate global correction by executing the global BA in a background thread, which significantly improves tracking accuracy, as demonstrated in the reported experiments, although at a slower computational speed compared to real-time rates. For both approaches, the computational cost is largely due to the lack of flexibility of latent feature grids to accommodate pose corrections from loop closures. Indeed, this requires re-allocating feature grids and retraining the entire map once a loop is corrected and poses are updated. However, this challenge becomes more pronounced as the number of frames processed increases, leading to the accumulation of camera drift errors and eventually either an inconsistent 3D reconstruction or a rapid collapse of the reconstruction process.

Overall, decoupled methods, which separate the mapping and tracking processes, tend to achieve better tracking performance compared to coupled approaches. By allowing the tracking module to focus solely on camera pose estimation without the added complexity of simultaneously updating the map representation, decoupled methods can achieve more accurate and robust tracking. However, this improved accuracy and robustness come at the cost of increased computational overhead, as the independent mapping and tracking stages require separate processing pipelines and memory allocation, which may impact the overall efficiency of the SLAM system.

\textbf{NeRF vs. 3DGS in SLAM.} NeRF-style SLAM, which relies mostly on MLP(s), is well suited for novel view synthesis, mapping and tracking but faces challenges such as oversmoothing, susceptibility to catastrophic forgetting, and computational inefficiency due to its reliance on per-pixel ray marching.  3DGS bypasses per-pixel ray marching and exploits sparsity through differentiable rasterization over primitives. This benefits SLAM with an explicit volumetric representation, fast rendering, rich optimization, direct gradient flow, increased map capacity, and explicit spatial extent control. Thus, while NeRF shows a remarkable ability to synthesize novel views, its slow training speed and difficulty in adapting to SLAM are significant drawbacks. 3DGS, with its efficient rendering, explicit representation, and rich optimization capabilities, emerges as a powerful alternative. Despite its advantages, current 3DGS-style SLAM approaches have limitations. These include scalability issues for large scenes, the lack of a direct mesh extraction algorithm (although recent methods such as \cite{guedon2023sugar} have been proposed), the inability to accurately encode precise geometry and, among others, the potential for uncontrollable Gaussian growth into unobserved areas, causing artifacts in rendered views and the underlying 3D structure. Moreover, the computational complexity of 3DGS-based SLAM systems is significantly higher than NeRF-based methods, which can hinder real-time performance and practical deployment, especially on resource-constrained devices. In order to mitigate these issues, recent research efforts, such as Compact-GSSLAM \cite{deng2024compact}, have focused on developing compact 3D Gaussian scene representations that optimize storage efficiency while maintaining high-quality reconstruction, rapid training convergence, and real-time rendering capabilities.

\textbf{Evaluation Inconsistencies.} The lack of standardized benchmarks or online servers with well-defined evaluation protocols results in inconsistent evaluation methods, making it difficult to conduct fair comparisons between approaches and introducing inconsistencies within the methodologies presented in different research papers. This is exemplified by challenges in datasets such as ScanNet \cite{dai2017scannet}, where ground truth poses are derived from BundleFusion \cite{bundlefusion}, raising concerns about the reliability and generalizability of evaluation results. 
Xu et al. \cite{xu2024customizable} and Hua et al. \cite{hua2024benchmarking} both acknowledge these inconsistencies and propose solutions to address them. Xu et al. \cite{xu2024customizable} introduce a comprehensive taxonomy of perturbations for SLAM in dynamic and unstructured environments, along with the Robust-SLAM dataset\footnote{\url{https://github.com/Xiaohao-Xu/SLAM-under-Perturbation/}}, created using 3D scene models sourced from Replica, which includes diverse perturbation types and offers a consistent evaluation protocol. Similarly, Hua et al. \cite{hua2024benchmarking} establish an open-source benchmark framework\footnote{\url{https://vlis2022.github.io/nerf-slam-benchmark/}} to evaluate the performance of a wide spectrum of commonly used implicit neural representations and geometric rendering functions for examining their effectiveness in mapping and localization. They propose a novel RGB-D SLAM benchmark framework, featuring a unified evaluation protocol to assess different NeRF components effectively.
These works highlight the importance of standardized benchmarks and evaluation protocols in mitigating inconsistencies and enabling more reliable and generalizable research outcomes in SLAM. However, to further address these issues, we believe it is crucial to establish online evaluation platforms with well-defined protocols, error metrics, and leaderboards for tracking, mapping, similar in spirit to the ETH3D benchmark \cite{schops2019bad}. These online benchmarks should provide high-quality ground truth data for both mapping and tracking, ensuring that the proposed methods are evaluated against reliable and accurate reference data.  Moreover, we consider that they should include a dedicated evaluation protocol for novel view rendering to address overfitting risks and promote more generalizable rendering methods. In summary, by adopting standardized benchmarks, well-defined protocols, and high-quality ground truth data, we believe that the research community can make more informed and fair comparisons between different approaches.

\textbf{Additional Challenges.} SLAM approaches, whether traditional, deep learning based, or influenced by radiance field representations, face common challenges. One notable obstacle is the handling of \textit{dynamic scenes}, which proves difficult due to the underlying assumption of a static environment, leading to artifacts in the reconstructed scene and errors in the tracking process. While some approaches attempt to address this issue, there is still significant room for improvement, especially in highly dynamic environments.

Another challenge is sensitivity to \textit{sensor noise}, which includes motion blur, depth noise, and aggressive rotation, all of which affect tracking and mapping accuracy. This is further compounded by the presence of \textit{non-Lambertian objects} in the scene, such as glass or metal surfaces, which introduce additional complexity due to their varying reflective properties. In the context of these challenges, it is noteworthy that many approaches often overlook explicit \textit{uncertainty estimation} across input modalities, hindering a comprehensive understanding of system reliability.

Additionally, the absence of external sensors, especially depth information, poses a fundamental problem to \textit{RGB-only} SLAM, leading to depth ambiguity and 3D reconstruction optimization convergence issues.

A less critical but specific issue is the quality of rendered images of the scene. Reviewed techniques often struggle with \textit{view-dependent appearance} elements, such as specular reflections, due to the lack of modeling of view directions in the model, which affects rendering quality.

%% file: chapters/conclusion.tex
\section{Conclusion}
\label{sec:conclusion}

In summary, this overview pioneers the exploration of SLAM methods influenced by recent advances in radiance field representations. Ranging from seminal works such as iMap \cite{sucar2021imap} to the latest advances, the review reveals a substantial body of literature that has emerged in just three years.  Through structured classification and analysis, it highlights key limitations and innovations, providing valuable insights with comparative results across tracking, mapping, and rendering. It also identifies current open challenges, providing interesting avenues for future exploration.